%% file: main.tex
\RequirePackage[colaction]{multicol}
\documentclass[10pt,twocolumn,letterpaper]{article}
\usepackage{cvpr}
\usepackage{iftex}
\ifPDFTeX
  \usepackage{times}
\else
  \usepackage{fontspec}
\fi
\input{math_commands.tex}

\input{format_compat.tex}

\usepackage{graphicx}
\usepackage{booktabs}
\usepackage{amsmath,amssymb}
\usepackage{algorithm}
\usepackage{algpseudocode}
\usepackage{array}
\usepackage{tabularx}
\usepackage{longtable}
\usepackage{microtype}
\usepackage{float}
\usepackage{marvosym}
\usepackage{xcolor}
\usepackage{colortbl}
\usepackage{xspace}
\definecolor{cvprblue}{rgb}{0.21,0.49,0.74}
\usepackage[pagebackref,breaklinks,colorlinks,allcolors=cvprblue]{hyperref}
\def\confName{CVPR}
\def\confYear{2027}
\title{RoboDemo: Reconciling Demonstration Intent and Execution Progress \\ for Long-Horizon Manipulation}
\input{authors}

\makeatletter
\g@addto@macro\@maketitle{\input{segments/teaser}}
\makeatother



\begin{document}
\raggedbottom
\camerareadymaketitle

\input{segments/00_abstract}

\input{segments/paper_body}

\input{segments/statements}

{\small
\bibliographystyle{ieeenat_fullname}
\bibliography{references}
}

\clearpage
\onecolumn
\begin{center}
{\Large\bfseries Supplementary Material\par}
\end{center}
\appendix
\raggedbottom
\input{segments/appendix_layout}
\input{segments/06_more_related_work}
\clearpage
\input{segments/appendix}

\begin{multicols}{2}
\raggedcolumns
\input{segments/appendix_experiment_reporting}
\end{multicols}
\begin{multicols}{2}
\raggedcolumns
\input{segments/appendix_refreward}

\end{multicols}
\end{document}

%% file: math_commands.tex
\usepackage{amsmath,amsfonts,bm}

\def\eqref#1{equation~\ref{#1}}

\def\1{\bm{1}}

\DeclareMathAlphabet{\mathsfit}{\encodingdefault}{\sfdefault}{m}{sl}
\SetMathAlphabet{\mathsfit}{bold}{\encodingdefault}{\sfdefault}{bx}{n}



%% file: format_compat.tex
\ifPDFTeX
  \DeclareUnicodeCharacter{2003}{\quad}
\else
  \catcode"2003=\active
  \begingroup
  \lccode`\~="2003
  \lowercase{\endgroup\def~}{\quad}
  
\fi

%% file: authors.tex
\author{%
Xiangkai Ma\textsuperscript{1,}\thanks{\raggedright Xiangkai Ma
(\mbox{\href{mailto:xiangkai.ma@smail.nju.edu.cn}{xiangkai.ma@smail.nju.edu.cn}})
and Zhihao Yuan
(\mbox{\href{mailto:zhihaoyuan@link.cuhk.edu.cn}{zhihaoyuan@link.cuhk.edu.cn}})
contributed equally to this work and are co-first authors.}\quad
Zhihao Yuan\textsuperscript{2,*}\quad
Linbo Zhai\textsuperscript{3}\quad
Xinyi Wang\textsuperscript{1}\quad
Junjie Wang\textsuperscript{4}\quad
Sheng Xu\textsuperscript{4}\\
Guiliang Liu\textsuperscript{4}\quad
Han Zhang\textsuperscript{1}\quad
Chuyi Chen\textsuperscript{1}\quad
Jifeng Gao\textsuperscript{1}\quad
Wenzhong Li\textsuperscript{1,}\thanks{Corresponding author: Wenzhong Li
(\href{mailto:lwz@nju.edu.cn}{lwz@nju.edu.cn}).}\\[3pt]
{\normalsize\textsuperscript{1}Nanjing University\quad
\textsuperscript{2}Joy Future Academy, JD}\\
{\normalsize\textsuperscript{3}South China University of Technology\quad
\textsuperscript{4}The Chinese University of Hong Kong, Shenzhen}\\[3pt]
{\normalsize\url{https://robodemo-2027.github.io/}}%
}

\hypersetup{
  pdftitle={RoboDemo: Reconciling Demonstration Intent and Execution Progress for Long-Horizon Manipulation},
  pdfauthor={Xiangkai Ma, Zhihao Yuan, Linbo Zhai, Xinyi Wang, Junjie Wang, Sheng Xu, Guiliang Liu, Han Zhang, Chuyi Chen, Jifeng Gao, Wenzhong Li}
}

\makeatletter
\newcommand{\camerareadymaketitle}{%
  \begingroup
  \renewcommand{\@fnsymbol}[1]{%
    \ifcase##1\or *\or\mbox{\Letter}\else\@ctrerr\fi}%
  \maketitle
  \endgroup
}
\makeatother

%% file: segments/teaser.tex
\noindent
\begin{minipage}{\textwidth}
\centering
\captionsetup{type=figure}
\vspace{-25pt}
\includegraphics[width=0.98\linewidth]{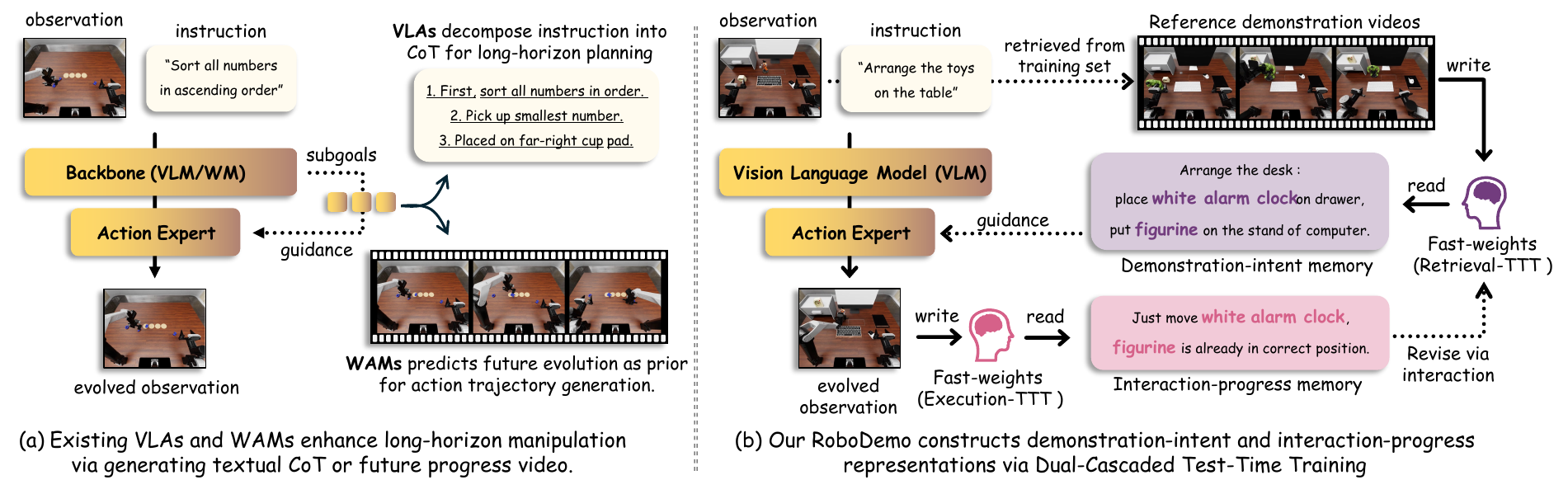}
\vspace{-10pt}
\caption{From (a) generated guidance to (b) retrieved video as guidance. 
RoboDemo preserves reference guidance while tracking actual execution, enabling the policy to adapt to deviations without losing the intended procedure.
}
\label{fig:teaser}
\end{minipage}
\par
\vspace{10pt}

%% file: segments/00_abstract.tex
\begin{abstract}
Long-horizon robot manipulation requires more than predicting the next action from the current observation. 
The policy must maintain a global procedural prior that specifies the expected progress of each subtask stage, while also tracking the realized progress produced by its interactions. 
To this end, we propose RoboDemo, a demonstration-augmented manipulation framework built on Dual-Cascaded Test-Time Training. 
RoboDemo accepts either a task-relevant human demonstration video or a semantically compatible robot video retrieved from simulation corpus as reference, allowing us to evaluate its applicability with and without paired human demonstrations. 
Task semantics and physical motion information from the reference video are then transferred into persistent demonstration-intent memory.
During deployment, we causally compress online observations, robot states, and executed actions into interaction-progress memory. 
To keep reference intent stable while interaction progress continually evolves, reference and execution video tokens independently update their respective fast-weight states. 
The action expert performs cascaded memory reads, first obtaining reference guidance and then refining it using realized progress, allowing the policy to adapt to execution deviations without overwriting the stored intent. 
RoboDemo achieves 90.7\% success on Imitator Game and 73.3\% success across three real-world household tasks.
\textbf{Real-world videos, code and hf-weights are available in our project website.}
\end{abstract}

%% file: segments/paper_body.tex
\input{segments/01_introduction}

\input{segments/02_related_work}

\input{segments/03_method}
\input{segments/04_experiment_design}

\input{segments/05_discussion}

%% file: segments/01_introduction.tex
\section{Introduction}
\label{sec:introduction}

Vision Language Action models (VLAs) \citep{black2024pi0,brohan2023rt2,kim2024openvla,bjorck2025gr00t,octo2024,black2025pi05} have achieved remarkable success by mapping observations and instructions to executable robot actions. 
World Action Models (WAMs) \citep{cheang2024gr2,jang2025dreamgen,li2025unifiedvideoaction,yu2026maskwam} further enrich this paradigm by injecting predictive visual dynamics into trajectory generation. 
Nevertheless, existing policies solve each control step from the current observation and proprioceptive state, and focus on short action window. 
This locally reactive interaction becomes insufficient as long-horizon manipulation increasingly emerges as a central challenge in embodied intelligence \citep{jiang2026robotttt,sun2026memobench,yin2026simplememvla}.
Successful long-horizon manipulation requires two complementary forms of knowledge beyond the current state.
First, the policy needs a global procedural prior that guides the entire manipulation process: it should specify how each sub-task ought to be executed and what robot states should be reached when that stage terminates. 
Second, it needs an estimate of realized progress that records which sub-goals have actually been completed, and determines whether the current execution still follows the intended procedure \citep{gao2026agm,huang2026chainvla,nie2026pace,zhao2026mapwam}. 
Long-horizon manipulation is not merely a next-action-prediction, but a persistent context construction over intended and realized progress.

Early VLAs \citep{jiang2023vima,zawalski2024ecot,zhang2025cotvla} generate chains of thought (CoT) to precisely plan future motion trajectories, which serve as context to guide action generation. 
However, text-only CoT does not fully specify the manipulation procedure. 
Such as ``place the object on the plate'' does not determine the initial spatial layout, approach direction and relative gripper-object pose at contact \citep{jain2024vid2robot,papagiannis2025rx,wang2023mimicplay}. 
Subsequently, WAMs \citep{cheang2024gr2,jang2025dreamgen,li2025unifiedvideoaction,yu2026maskwam} alleviate this ambiguity by predicting future observations as intermediate goals and recovering actions through inverse dynamics. 
Yet explicit prediction introduces substantial training cost and test-time latency \citep{li2026lawa,yuan2026fastwam}.

\begin{figure*}[t]
\begin{center}
\vspace{-20pt}
\centerline{\includegraphics[width=\linewidth]{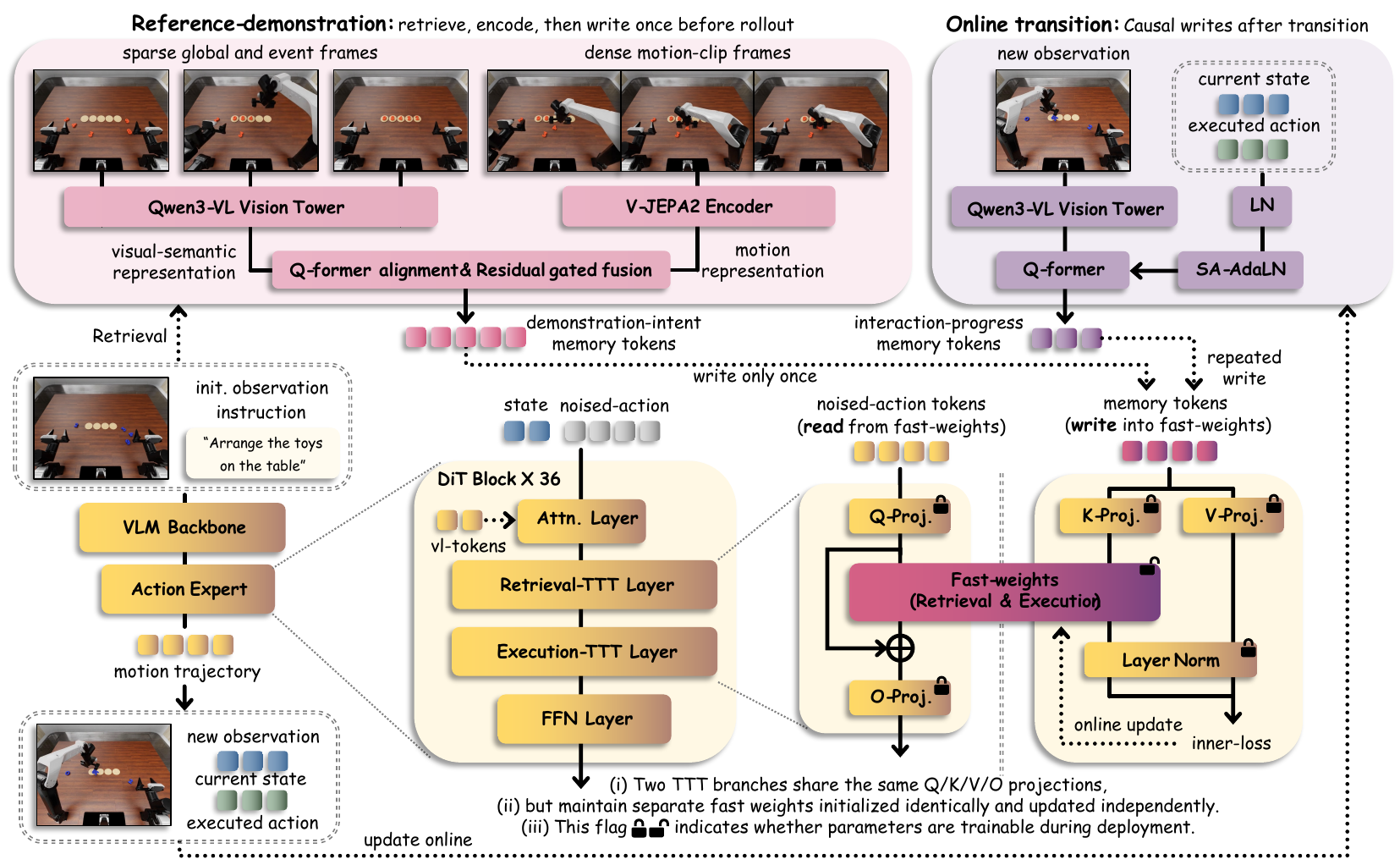}}
\vspace{-10pt}
\caption{\textbf{Overview of RoboDemo.} 
A reference demonstration is encoded into demonstration-intent memory, while online observations, robot states, and executed actions update a separate interaction-progress memory. 
The action expert sequentially reads these memories through Retrieval-TTT and Execution-TTT, combining reference guidance with realized progress to generate robot actions.}
\label{fig:method}
\end{center}
\vspace{-30pt}
\end{figure*}

More recently, video-conditioned policies \citep{feng2026wamttt,jain2024vid2robot,papagiannis2025rx,wang2023mimicplay,zeroWAM2026} have used demonstrations as prompts or test-time adaptation signals. 
WAM-TTT \citep{feng2026wamttt} and RoboTTT \citep{jiang2026robotttt} compress human demonstrations into fast weights at deployment, steering frozen WAM and VLA to rapidly adapt to new task scenarios. 
Zero-WAM \citep{zeroWAM2026} constructs paired human-robot videos and utilizes a generated human demonstration as prefix context for zero-shot task. 
Long-WAM \citep{huang2026longwam} builds long context by extending the causal visual prefix with the robot's own observation history. 
These approaches reveal the value of video as procedural guidance, but recording or synthesizing a new demonstration for every task and scene remains costly, and generated videos may violate physics laws \citep{yang2026genvid2robot,li2026badwam}.
To address these challenges, we consider two settings: conditioning action generation on task-relevant human demonstrations, or on semantically and visually compatible robot videos retrieved from the simulation corpus.

However, a reference demonstration describes only the intended evolution of the task.
It can show where the robot should move, but it fails to determine what the deployed robot has actually completed after contact errors, delays or perturbations.
Fortunately, the accumulating online observations provide this complementary information.
A long-horizon policy should therefore preserve two semantic contexts with different roles and life-cycles: a stable demonstration-intent memory and a causal interaction-progress memory.
Together, they allow the policy to compare intention with realization: the reference indicates where the robot should go, while online interaction reveals where it currently is.
Crucially, reference information guides the entire deployment process, whereas interaction information depends on instantaneous policy feedback and may contain noise from irrelevant observations.
If both are compressed into the same contextual reasoning space, later execution information can overwrite or obscure the original procedure.
The two contexts should therefore be maintained in separate fast-weight states, preserving reference intent while allowing interaction progress to accumulate independently.

To realize this principle, we propose RoboDemo, a retrieval-augmented long-context manipulation policy built on Dual-Cascaded Test-Time Training. 
We use a task-relevant human demonstration as reference when available, or retrieve a compatible robot video from the simulation corpus.
We design a hierarchical sampling strategy and a joint visual encoder to extract both task-aligned sparse semantics and dense fine-grained motion dynamics.
As shown in Fig.~\ref{fig:method}, Retrieval-TTT writes the reference memory tokens before rollout and retains the fast-weight state to provide persistent procedural guidance. 
During rollout, we compress each new observation into interaction-progress memory tokens, then inject the current robot state and the preceding executed action as conditioning information. 
These writes accumulate interaction-progress memory, enabling Execution-TTT to track the robot's progress.
The Action-DiT block first reads the Retrieval-TTT memory and then lets Execution-TTT revise the reference-conditioned representation according to realized progress. 
Unlike conventional TTT with Self-Memorization, where the same token sequence both writes to and reads from fast weights, we introduce Cross-Memorization to decouple writing from reading: demonstration-intent and interaction-progress memory tokens write to fast weights, while action tokens read.

Our contributions are summarized as follows:
\begin{itemize}
\item We propose RoboDemo, a novel VLA framework that constructs demonstration-intent and interaction-progress representations, preserving a global procedural prior as persistent guidance while correcting reference-execution deviations after each interaction.
\item We introduce a cross-memorization regimen that decouples fast-weight writes from reads, thereby avoiding the conflation of demonstration-intent and interaction-progress memory lifecycles.
\item RoboDemo achieves 90.7\% mean Seen-task success across C15/C30/C45 on Imitator Game and 73.3\% averaged across three real-world household tasks. A separate C15 comparison obtains 90.0\% with retrieved robot references versus 89.5\% with human references. Real-world visual-transfer tests further evaluate robustness to lighting and background changes without target-condition fine-tuning.
\end{itemize}

%% file: segments/02_related_work.tex
\section{Related Work}
\label{sec:related}

\subsection{In-Context Learning}
VLAs and WAMs have achieved remarkable success by using textual CoT and predicted trajectories as context for action generation \citep{zawalski2024ecot,zhang2025cotvla,lyu2026lda}. 
However, producing these intermediate results introduces additional latency \citep{zawalski2024ecot,yang2026genvid2robot}. 
Fast-WAM~\citep{yuan2026fastwam}, VLA-JEPA~\citep{sun2026vlajepa} and PILOT~\citep{ma2026pilot} alleviate this limitation by retaining future prediction as training supervision while generating actions without explicitly synthesizing future videos at test time. 
Demonstration-conditioned in-context learning offers a more direct alternative: obtaining procedural guidance from demonstrations rather than predicting the intended future from scratch. 
Representative methods, including ICRT~\citep{fu2024icil}, Vid2Robot~\citep{jain2024vid2robot} and RICL~\citep{sridhar2025ricl}, use demonstration trajectories as prompts or retrieved context, allowing policies to adapt their behavior without task-specific fine-tuning. 
Meanwhile, memory-based policies~\citep{shi2025memoryvla,yin2026simplememvla} preserve past execution through retrieved visual evidence or video prefixes to guide subsequent actions.
A reference demonstration, however, describes how the task should proceed, not what the deployed robot has actually completed. 
To address this issue, RoboDemo constructs two independent memory mechanisms, enabling the action expert to benefit from both reference demonstrations and actual execution.

\subsection{Test-Time Training}
TTT-based methodologies \citep{sun2024ttt,dalal2025videottt} have shown that fast weights can retain long-term context for language modeling and temporally coherent video generation. 
Building on this advantage, RoboTTT~\citep{jiang2026robotttt} integrates TTT into VLAs to compress long vision-motion histories into fast weights, enabling human-video in-context imitation and online policy improvement. 
TTT-RM~\citep{sun2026tttrm} complements a bounded memory bank to identify and preserve historical information that cannot be recovered from the current observation and retrieved memory.
WAM-TTT~\citep{feng2026wamttt} adapts lightweight memory within frozen WAM through self-supervised prediction on human videos, with paired human-robot aligning this adaptation with robot motion trajectory.
Together, these methods demonstrate the potential of TTT for long-context control and demonstration-based steering.
RoboDemo addresses a complementary challenge: reconciling demonstrated intent with realized execution progress. 

%% file: segments/03_method.tex
\section{Reconciling Demonstration Intent and Execution Progress with Dual-Cascaded TTT}
\label{sec:method}

\subsection{The Dual-Context Paradigm for Manipulation}
\label{sec:dual_context}
Long-horizon manipulation requires a policy to distinguish how a task should proceed from how execution has actually progressed. 
We therefore formulate long-horizon control around two complementary contexts: demonstration intent and realized interaction progress.
At control step $t$, the robot receives visual observation $o_t$, proprioceptive state $s_t$, and a language instruction $\ell$. 
Let $\mathcal D$ denote a reference video containing $N$ frames, with $d_n$ its $n$-th frame. 
For each earlier step $j<t$, let $a_j^{\mathrm{exec}}$ denote the action actually executed from $o_j$ and $s_j$, where contexts are $\mathcal D=\{d_n\}_{n=1}^{N}$ and $\mathcal H_{<t}^{\mathrm{real}}=\{(o_j,s_j,a_j^{\mathrm{exec}})\}_{j<t}$.
For an action horizon $A$, let $\mathbf a_t=a_{t:t+A-1}$ denote the predicted action chunk. Our policy, parameterized by slow weights $\theta$, models
\begin{equation}
\pi_\theta\left(\mathbf a_t\mid o_t,s_t,\ell,\mathcal D,\mathcal H_{<t}^{\mathrm{real}}\right).
\end{equation}
We instantiate this formulation with a vision-language backbone and a flow-matching action transformer augmented by Retrieval-TTT and Execution-TTT. 
In Action-DiT token sequence, the learned sink token $h_{\mathrm{sink}}$, state embedding $h_{\mathrm{state}}$, and $A$ noised-action embeddings $h_{\mathrm{action}}$ form the token sequence of action expert: $H_t^0=[h_{\mathrm{sink}},h_{\mathrm{state}},h_{\mathrm{action}}]\in\mathbb R^{(A+2)\times D}$ , where $D$ denotes the hidden width. 


\subsection{Learning to Select Reference Demonstrations}
\label{sec:refreward}
RoboDemo uses a task-relevant human demonstration when available. Otherwise, Refreward, an offline-supervised scorer, selects a reference from the robot training corpus.

A useful reference must match both the task instruction and the initial scene. 
At deployment, we form the retrieval query $q=(\ell_q,o_q^0)$ from the target episode's instruction and initial observation. 
Candidate references are drawn from a task-compatible subset of the robot training corpus, indexed by $\mathcal C_q$. 
Each candidate $e_i=(\ell_i,o_i^0,V_i)$ contains its instruction, initial observation, and complete demonstration video. 
We use frozen Qwen3-VL-Embedding-2B \citep{li2026qwen3vlembedding} and DINOv3 \citep{simeoni2025dinov3} as a joint encoder to map the query and each candidate to multimodal representations $\xi_q$ and $\xi_i$, respectively. 
Refreward, a lightweight scorer with parameters $\psi$, then evaluates each candidate's suitability as a reference for the target episode through $r_i=\operatorname{Refreward}_{\psi}(\xi_q,\xi_i)$. 
We select the highest-scoring candidate, with index $i^\star=\arg\max_{i\in\mathcal C_q}r_i$.
For each target episode, a GPT teacher provides suitability grades and pairwise preferences for candidate demonstrations. We use these labels to train Refreward to rank candidates by their suitability as references, emphasizing relative ordering rather than absolute scores.

\subsection{Hierarchical Sampling and Semantic-Motion Joint Encoding for Reference Video}
\label{sec:hybrid12}
\label{sec:joint_encoder}
Encoding a reference demonstration requires complementary representations of scene content and physical motion. 
Individual frames capture object appearance and spatial layout but leave motion direction and continuity ambiguous, while local motion alone does not fully establish task context. 
We therefore combine sparse semantic observations with dense local clips.
Specifically, Hierarchical Global-Event-Clip sampling takes the complete reference video as input and selects uniformly spaced global frames to preserve coarse task progression, event frames to capture pronounced interactions, and a dense clip around the highest-scoring event to capture local motion. 
Events are identified using normalized optical flow and, when available, robot-state and gripper changes. Robot measurements are used only for offline frame selection, not as reference-encoder inputs.

As illustrated by the pink branch in Fig.~\ref{fig:method}, the frozen Qwen3-VL vision tower \citep{qwen3vl2025} encodes global and event frames, while frozen V-JEPA2 \citep{assran2025vjepa2} encodes the local clip. Separate resamplers compress their outputs into fixed-size semantic registers $A^S$ and motion registers $A^M$. We fuse them into demonstration-intent memory through motion residual:
\begin{equation}
U^R = A^S + \tanh(\alpha_M)\,G \odot A^M.
\label{eq:method-reference-tokens}
\end{equation}
Here, $G$ is a learned channel-wise gate, $\alpha_M$ controls the motion contribution, and $\odot$ denotes element-wise multiplication. Initializing $\alpha_M$ near zero keeps the semantic representation dominant at the start of training.
Before rollout, the resulting tokens are written to Retrieval-TTT to construct a reference fast-weight state, which remains fixed and provides persistent guidance throughout execution.

\subsection{State-Action Conditioning and Adaptive Encoding for Execution Video}
\label{sec:execution_encoder}
Unlike the reference demonstration, interaction progress evolves as the robot acts. 
At control step $t\geq1$, we represent the latest completed transition using the observation $o_t$, robot state $s_t$, and preceding executed action $a_{t-1}^{\mathrm{exec}}$.
As illustrated by the purple branch in Fig.~\ref{fig:method}, frozen Qwen3-VL vision tower encodes the observation into visual tokens, which an execution-specific Q-former compresses into a fixed-size set of registers $\widehat U_t^E$.
Visual observations alone may not distinguish similar interaction states. Therefore, we inject state-action information through State-Action-Conditioned Adaptive Layer Normalization. The state and action embeddings are concatenated to predict channel-wise scale $\gamma_t$ and shift $\beta_t$, producing interaction-progress memory tokens:
\begin{equation}
U_t^E =
(1+g_\gamma\odot\gamma_t)\odot
\operatorname{LN}(\widehat U_t^E)
+g_\beta\odot\beta_t.
\label{eq:method-execution-tokens}
\end{equation}
Here, $\operatorname{LN}$ denotes layer normalization, $\odot$ denotes element-wise multiplication, and $g_\gamma,g_\beta$ are learned modulation gates. 
The modulation is broadcast across register positions. Both gates are initialized to zero, so the encoder initially produces normalized visual registers.
The resulting tokens are written to Execution-TTT once per completed transition, before the next action is sampled. And execution memory reflects actual interactions rather than unexecuted action predictions.

\subsection{Dual-Cascaded Test-Time Training}
\label{sec:dcttt}


\subsubsection{Writing and Reading Operations in TTT}
\label{sec:ttt_operations}
TTT stores context in the fixed-size fast-weight state: 1) writing updates fast weights to encode incoming contextual representations, 2) while reading evaluates those weights to recover representations relevant to a query. 
Within each Action-DiT block, Retrieval-TTT and Execution-TTT share the projections $W_K,W_V,W_Q,W_O$ and learned initialization $\phi_0$, but maintain two fast-weights initialized as storage-independent copies, $\phi^r\leftarrow\operatorname{Copy}(\phi_0)$ and $\phi^e\leftarrow\operatorname{Copy}(\phi_0)$, and subsequently evolve independently. 
We denote either state by $\phi^\star$, where $\star\in\{r,e\}$.


\vspace{-10pt}
\paragraph{Writing.}
The fast state $\phi=(W_1,b_1,W_2,b_2)$ parameterizes a two-layer MLP,
$F_\phi(X)=\operatorname{GELU}(XW_1+b_1)W_2+b_2$.
For writer tokens $U_i^\star$, where $i$ indexes token positions, the shared projections form keys $K_i$ and values $V_i$. The MLP learns to reconstruct their normalized residual:
\begin{equation}
\begin{aligned}
K_i&=\operatorname{norm}(W_K(U_i^\star)),
V_i=W_V(U_i^\star),\\
Y_i&=\operatorname{LN}_{\mathrm{in}}(V_i-K_i),
\widehat Y_i=\operatorname{LN}_{\mathrm{in}}(F_{\phi^\star}(K_i)),\\
\mathcal L&_{\mathrm{in}}(\phi^\star,U^\star)
=\frac12\sum_i m_i\|\widehat Y_i-Y_i\|_2^2.
\end{aligned}
\label{eq:method-inner-loss}
\end{equation}
Here, $\operatorname{norm}$ and $\operatorname{LN}_{\mathrm{in}}$ denote vector and layer normalization, and $m_i$ masks invalid tokens. Writer-side role and time conditioning is absorbed into $U_i^\star$. The operation $\operatorname{TTT\!-\!Writer}(\phi^\star,U^\star,\star)$ updates the fast state using token-wise reconstruction gradients and learned token-dependent step sizes, without modifying the shared projections.


\vspace{-10pt}
\paragraph{Reading.}
We add the learned role embedding $e_{\mathrm{read}}^\star$ to the action-side tokens $H$, then project and normalize the result to obtain queries $q^\star$. The fast MLP processes these queries, and its normalized response is added back to them before post-read layer normalization and output projection:
\begin{equation}
\begin{aligned}
q^\star&\textit{=}\operatorname{norm}\!\left(W_Q(H\textit{+}e_{\mathrm{read}}^\star)\right),\\
\operatorname{R}(\phi,H,\star)
&\textit{=}W_O\!\left(\operatorname{LN}_{\mathrm{post}}
\left[q^\star\textit{+}\operatorname{LN}_{\mathrm{in}}(F_\phi(q^\star))\right]\right).
\end{aligned}
\label{eq:method-read-response}
\end{equation}
To isolate the contribution of accumulated context, the TTT-Reader subtracts the initial-state response under same query:
\begin{equation}
\begin{aligned}
\operatorname{TTT\!-\!Reader}(H,\phi^\star,\star)\textit{=}\operatorname{R}(\phi^\star,H,\star)\textit{-}\operatorname{R}(\phi_0,H,\star).
\end{aligned}
\label{eq:method-delta-reader}
\end{equation}
Both evaluations use identical projections and role conditioning. This dynamic-delta read makes an unwritten or reset memory contribute exactly zero, while a written memory contributes its context-induced change. Reading leaves the fast state unchanged.


\subsubsection{Decoupling Context Writing from Action Reading}
\label{sec:cross_memorization}

The write-read distinction enables two different organizations of TTT. In Self-Memorization, the tokens that write and read the fast weights originate from the same sequence. This architecture does not align with the original design intention of RoboDemo: RoboDemo instead requires two contextual streams to guide a separate action-generation stream. Concatenating demonstration and execution videos as prefixes to the noised-action sequence would expose the action expert to their explicit context tokens. Using a single fast state would avoid that concatenation but couple a stable reference representation to continually changing execution representations. Neither organization provides the separation needed here.

We therefore introduce Cross-Memorization, in which memory tokens write fast weights and a different token stream reads them. The demonstration encoder produces demonstration-intent memory tokens for Retrieval-TTT, while the execution encoder produces interaction-progress memory tokens for Execution-TTT. The action expert's sink, state, and noised-action tokens query both memories without becoming their writer inputs. This creates two independent context-writing pipelines alongside a fixed-length action-reading sequence.
For each Action-DiT block $l\in\{1,\ldots,L\}$, let $\phi_R^l$ denote the reference state and $\phi_{E,t}^l$ the execution state before action generation at control step $t$. Reference preloading starts from a storage-independent copy of the block-specific initialization $\phi_0^l$, while $\phi_{E,0}^l$ is initialized from a separate copy. Both use the same TTT-Writer:
\begin{align}
\phi_R^l&=\operatorname{TTT\!-\!Writer}
\left(\operatorname{Copy}(\phi_0^l),U^R,r\right),\\
\phi_{E,t}^{l}&=\operatorname{TTT\!-\!Writer}
\left(\phi_{E,t-1}^{l},U_t^E,e\right),~t\geq1.
\end{align}
Reference preloading finishes before rollout, after which $\phi_R^l$ remains fixed.
Execution writes follow completed transitions, using $U_t^E$ defined in Section~\ref{sec:execution_encoder}.
Cross-Memorization thus separates contextual accumulation from action sampling without appending either video history to the action-token sequence.

\subsubsection{Cascaded Insertion into Action-DiT}
\label{sec:cascaded_insertion}
Within each Action-DiT block, attention is followed by cascaded reads of demonstration intent and interaction progress, then the original feed-forward residual update. The reference state $\phi_R^l$ summarizes the complete demonstration, while $\phi_{E,t}^l$ contains the completed interaction history available at step $t$.
Starting from the attention output $H_{\mathrm{attn}}^l$, the shared TTT-Reader produces a reference-conditioned representation $H_R^l$ and then an execution-refined representation $H_E^l$. Learned gates $g_R^l$ and $g_E^l$ control the residual contribution:
\begin{equation}
\begin{aligned}
H_R^l&\textit{=}H_{\mathrm{attn}}^l\textit{+}\tanh(g_R^l)\operatorname{TTT\!-\!Reader}(H_{\mathrm{attn}}^l,\phi_R^l,r),\\
H_E^l&\textit{=}H_R^l\textit{+}\tanh(g_E^l)\operatorname{TTT\!-\!Reader}(H_R^l,\phi_{E,t}^l,e).
\end{aligned}
\label{eq:method-cascade}
\end{equation}
Unlike a parallel combination, the execution read uses $H_R^l$ as its query, allowing completed interactions to refine an already reference-conditioned representation. This correction changes the action-side representation without overwriting the reference fast weights.
The vision-language context is computed once for the current observation and reused across action-denoising steps. Both fast states remain fixed throughout this sampling process.


\begin{figure*}
\begin{center}
\vspace{-20pt}
\centerline{\includegraphics[width=\linewidth]{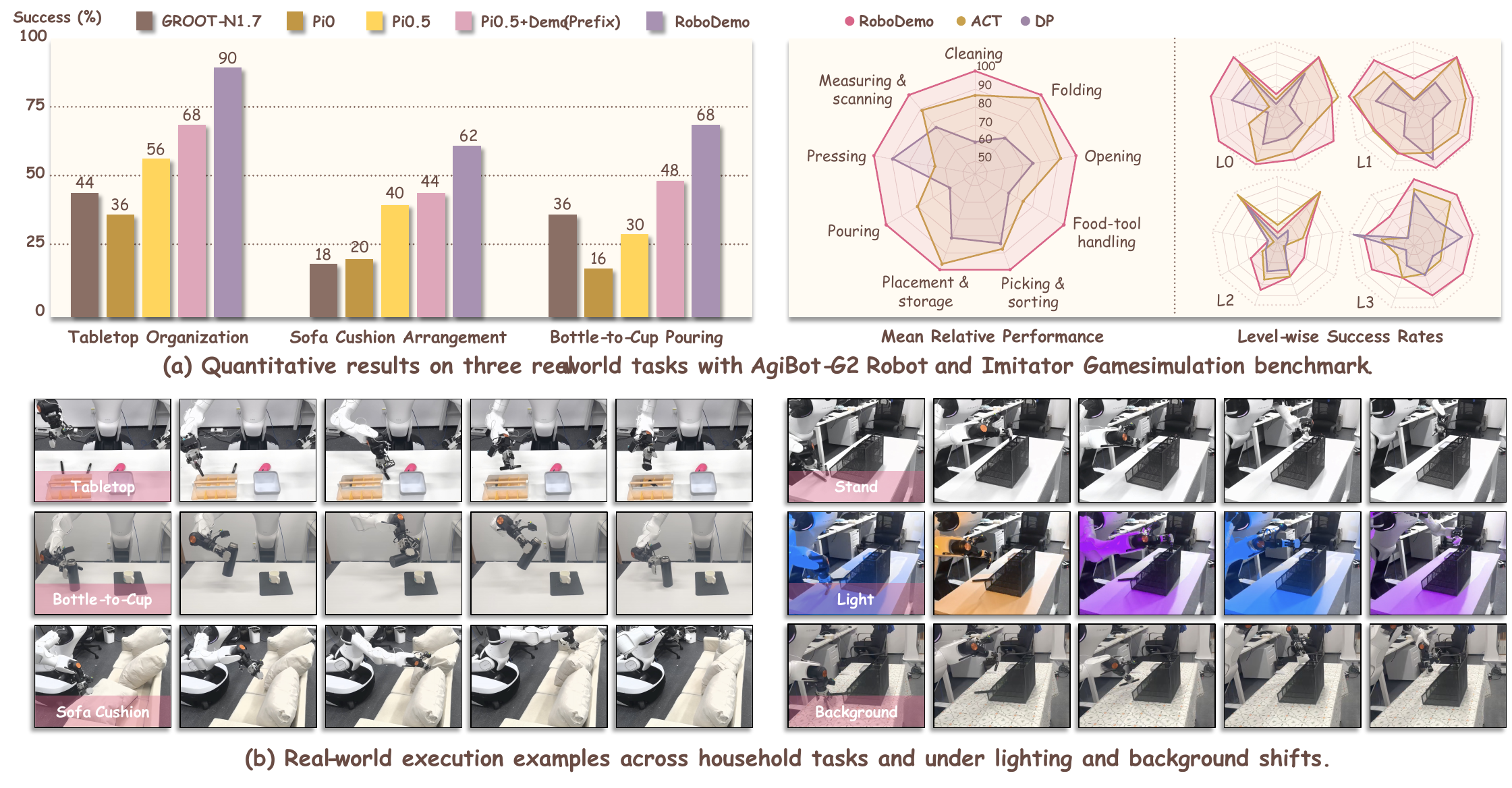}}
\vspace{-10pt}
\caption{\textbf{Real-world evaluation and simulation performance profiles.}
(a) Left: standard-task SR on the AgiBot-G2 humanoid, with 50 trials per method and task.
Right: Imitator Game Full50 profiles over nine operation groups, showing group-relative mean performance and L0-L3 SR.
Among them, the normalized success rate shown in ``Mean Relative Performance'', and the actual success rates shown in``Level-wise Success Rates''.
(b) Real-world execution examples across household tasks and under lighting and background changes.}
\label{fig:real-world-evaluation}
\end{center}
\vspace{-30pt}
\end{figure*}

\subsection{Training and Memory Updates}
\label{sec:training_state}

Training samples provide only a limited observation window, whereas Execution-TTT must retain an episode's accumulated interaction history. This creates two challenges: samples from different episodes require separate fast states, and resetting those states at every forward discards all earlier interaction-progress representations. We address them with episode-based batching and an episode-indexed memory pool. Episodes are shuffled, but their training windows are delivered chronologically. A batch containing $B$ episodes maintains $B$ independent sets of fast states. At each window, the policy restores the corresponding state, writes completed interactions in temporal order, and returns the terminal state to the pool. Memory values persist across windows, while stop-gradient at window boundaries prevents the computation graph from spanning the entire episode. Each sample is an $S$-step window: gradients pass through writes within that window, but not into preceding windows. 
The outer action-loss optimizes the trainable policy and memory-encoding parameters, including the shared TTT projections and fast-weight initialization. The inner reconstruction objective updates only the running fast-weights.


%% file: segments/04_experiment_design.tex
\section{Experiments}
\label{sec:experiments}

Imitator Game~\citep{zhou2026imitatorgamebenchmarkingrobot} evaluates whether a robot can reproduce the intent of a human demonstration as the execution scene changes. 
We use its 50-task simulation suite, comprising 10,000 robot demonstrations paired with human references (IG-10K). 
L0 preserves the demonstration scene, while L1-L3 change object positions, appearance, and functional affordances, respectively.
The nested C15/C30/C45 corpora contain 15/30/45 training tasks and share five Seen and five Unseen evaluation tasks. 
Full50 instead trains and evaluates on all 50 tasks. 
Most experiments labeled \textbf{RoboDemo} use human demonstrations as context, while those explicitly labeled \textbf{RoboDemo-wo} use retrieved robot videos instead. 
For the scale, zero-shot, and Full50 benchmarks, we report success rate (SR) over ten trials per task-level combination with a 500-control-step horizon: 200 Seen episodes per training scale, 200 Unseen episodes under C45, and 2,000 episodes for full50.
Further simulation details are provided in Appendix~\ref{app:benchmark}. Section~\ref{sec:real-world} separately reports real-world evaluation with 50 trials per task and condition.

\input{tables/imitationgame_full50}
\subsection{More Challenging Full-Suite Evaluation}
\label{sec:full50}
We evaluate RoboDemo on Full50, the complete 50-task simulation suite of Imitator Game, using the officially reported DP and ACT (SigLIP2) results as baselines. 
Table~\ref{tab:main} reports overall success rates together with breakdowns across seven task categories and four scene levels.

On Full50, RoboDemo achieves an overall SR of 84.6\%, surpassing DP and ACT by 20.5 and 9.7 percentage points, respectively. 
Beyond the overall average, RoboDemo outperforms both baselines in all seven task categories and at every scene level, establishing a consistent advantage across both evaluation dimensions. 
Against the stronger ACT baseline, it delivers double-digit gains in four categories: pouring (+16.6 points), tool use and scanning (+13.2), rack storage (+12.0), and tabletop manipulation (+11.5). 
The advantage also persists across L0-L3, with improvements over ACT ranging from 6.2 to 13.6 points, and the largest margin occurs at L3, where RoboDemo achieves 81.8\% SR versus ACT's 68.2\%. 
Appendix~\ref{app:full50-results} provides complete per-task results.

\input{tables/imitationgame_scales}
\subsection{Consistent Performance across Training Scales}
\label{sec:scales}
The nested training corpora test whether RoboDemo remains effective as the range of training tasks changes, while keeping the five Seen tasks fixed. 
Table~\ref{tab:scales} compares language-conditioned, video-skill, video-conditioned, and in-context baselines, with results from official simulation release.

RoboDemo achieves 89.5\%, 94.0\%, and 88.5\% on C15, C30, and C45, exceeding the strongest baseline at each scale by 7.5, 12.5, and 6.5 percentage points.
Its three-scale average reaches 90.7\%, compared with 80.8\% for ACT/DINOv2. We use this equally weighted C15/C30/C45 Seen-task mean as the summary success rate, reporting Full50 separately.
RoboDemo performs particularly well on rack-placement tasks: PlacePlateRack maintains at least 97.5\% success across all three settings, while PlaceMugRack reaches 92.5\%, 95.0\%, and 92.5\% on C15, C30, and C45.
For PlaceMugRack, success under position changes (L1) increases from 80.0\% on C15 to 90.0\% on both C30 and C45, while success under appearance changes (L2) rises from 90.0\% to 100\%, demonstrating sustained gains as the training corpus expands. 
Similarly, StirSpoon improves from 87.5\% on C15 to 97.5\% and 90.0\% on C30 and C45.
These results demonstrate strong performance across training scales, with sustained task-level gains and effective generalization to spatial and visual variations.
Complete results are in Appendix~\ref{app:selected-results}.

\input{tables/imitationgame_zero_shot}
\subsection{Generalization beyond the Training Tasks}
\label{sec:zero-shot}
We evaluate RoboDemo on five unseen tasks under the C45 setting to assess transfer beyond the tasks used for policy training. 
All five tasks are excluded from the three policy-training corpora, while task-relevant human demonstrations are provided at inference. 
This protocol tests whether RoboDemo can use demonstrations to perform unseen-tasks.

As shown in Table~\ref{tab:zero-shot}, RoboDemo achieves an average SR of 11.5\%, ranking second among the methods shown, behind Zero-WAM (22.0\%).
The gap to Zero-WAM partly reflect its large-scale pretraining with paired human videos and robot trajectories~\citep{zeroWAM2026}, which can provide transferable priors for interpreting human demonstrations and mapping them to robot actions across diverse tasks.
RoboDemo nevertheless outperforms most evaluated language-conditioned, video-skill, and video-conditioned baselines.
It exceeds the displayed VideoMAE-based ACT, DP, and VQ-BeT variants by 9.0, 8.5, and 4.0 points. Appendix~\ref{app:baseline-details} includes all encoder variants, including VQ-BeT/DINOv2 at 17.5\%.

Task-level results further highlight RoboDemo's capacity for demonstration-guided generalization. 
On FoldBox, RoboDemo achieves 47.5\% SR despite the task being excluded from all three policy-training corpora. 
This result illustrates how task-relevant human demonstrations provided at inference can support successful execution beyond the policy's training tasks, without task-specific policy retraining. 
Detailed results are in Appendices~\ref{app:selected-results} and~\ref{app:baseline-details}.

\subsection{Ablation Studies and Efficiency}
\label{sec:ablation-efficiency}
\input{tables/ablation_components}
\input{tables/inference_efficiency}

 Table~\ref{tab:ablation-components} compares six variants of reference conditioning and execution memory on C15-Seen across four scene levels.
Demonstration prefixes and Retrieval-TTT improve SR from 62.0\% to 68.5\% and 71.5\%, respectively, supporting the value of reference conditioning. TTT's additional 3.0-point gain may reflect persistent, action-query-dependent access to demonstration intent. Writing demonstration and execution contexts into one TTT yields only 73.0\% (+1.5 points), potentially due to interference between stable procedural intent and evolving interaction progress. Separate parallel TTT modules reach 82.0\% (+9.0 points), suggesting that distinct memory states better preserve these complementary contexts. Cascading Retrieval-TTT and Execution-TTT further increases SR to 89.5\% (+7.5 points), allowing execution evidence to refine reference-conditioned queries rather than contribute an independent parallel residual. Overall, RoboDemo gains 27.5 points over the base VLA, supporting separate context storage followed by conditional integration.
Table~\ref{tab:inference-efficiency} reports RoboDemo's parameter counts and inference latency on H200 with batch size one.

\subsection{Replacing Human Videos with Robot Videos}
\label{sec:reference-source}
\input{tables/imitationgame_reference_source}
RoboDemo-wo shares RoboDemo's architecture and training objective but replaces human demonstrations with robot videos retrieved from the C15 training bank. Crucially, training-reference retrieval excludes the query episode, selecting a similar but distinct episode rather than its own target trajectory. Evaluation references likewise come from the training bank.
As shown in Table~\ref{tab:reference-source}, RoboDemo-wo achieves 90.0\% SR on C15-Seen, exceeding RoboDemo's 89.5\% by 0.5 percentage points and ACT by 7.5 points. The results suggest that retrieved, cross-episode robot demonstrations can provide useful procedural guidance without a human reference at policy training or inference.

\input{segments/real_world_evaluation}

%% file: tables/imitationgame_full50.tex
\begingroup
\input{tables/experiment_style}
\begin{table*}[t]
\vspace{-20pt}
\caption{\textbf{Full-suite evaluation on Imitator Game.}
We report SR (\%) across all 50 tasks, grouped by task category and scene level.
Category results are averaged over L0-L3, while level results are averaged over all 50 tasks, and parentheses indicate task counts.
}
\label{tab:main}
\vspace{-8pt}
\centering
\small
\resizebox{1.0\textwidth}{!}{%
\tabcolsep=0.3cm
\renewcommand\arraystretch{1.0}
\begin{tabular}{l|*{7}{c}|*{4}{c}|c}
\toprule
 & \multicolumn{7}{c|}{By task category} & \multicolumn{4}{c|}{By level} & Overall\\
\cmidrule(lr){2-8}\cmidrule(lr){9-12}\cmidrule(l){13-13}
Method & Containers (13) & Racks (5) & Tabletop (10) & Pouring (6) & Cleaning (4) & Open/fold (5) & Tools/scan (7) & L0 & L1 & L2 & L3 & Avg.\\
\midrule
DP & 71.7 & 73.0 & 59.8 & 49.2 & 48.1 & 66.5 & 69.6 & 65.6 & 67.2 & 60.0 & 63.4 & 64.1\\
ACT & {78.3} & {80.5} & {73.3} & {66.3} & {65.0} & {86.0} & {71.8} & {80.8} & {82.0} & {68.4} & {68.2} & {74.9}\\
\textbf{RoboDemo} & \best{83.3} & \best{92.5} & \best{84.8} & \best{82.9} & \best{71.9} & \best{91.5} & \best{85.0} & \best{92.6} & \best{88.2} & \best{75.8} & \best{81.8} & \best{84.6}\\
\bottomrule
\end{tabular}
}
\vspace{-10pt}
\end{table*}
\endgroup

%% file: tables/experiment_style.tex
\definecolor{resultblue}{RGB}{48,126,190}
\newcommand{\best}[1]{\cellcolor{blue!15}#1}
\newcommand{\second}[1]{\cellcolor{blue!5}#1}

%% file: tables/imitationgame_scales.tex
\begingroup
\input{tables/experiment_style}
\begin{table}[t]
\vspace{-5pt}
\caption{\textbf{Seen-task performance across training scales.} SR (\%) on the same five Seen tasks covers C15/C30/C45 equally. Blue and light blue mark the best and second-best results per column.}
\label{tab:scales}
\vspace{-8pt}
\centering
\footnotesize
\resizebox{1.0\columnwidth}{!}{%
\tabcolsep=0.4cm
\renewcommand\arraystretch{1.0}
\begin{tabular}{l|*{4}{>{\centering\arraybackslash}p{0.84cm}}}
\toprule
Method & C15 & C30 & C45 & Avg.\\
\midrule
\multicolumn{5}{l}{\textit{Language-conditioned policies}}\\
\midrule
OpenVLA & 39.0 & 25.5 & 22.0 & 28.8\\
RDT-1B & 37.0 & 50.5 & 41.0 & 42.8\\
GR00T-N1.6 & 29.5 & 48.0 & 74.5 & 50.7\\
$\pi_{0.5}$ & 62.5 & 74.0 & \second{82.0} & 72.8\\
\midrule
\multicolumn{5}{l}{\textit{Video-skill policies}}\\
\midrule
UniSkill & 78.0 & 69.5 & 76.5 & 74.7\\
XSkill & 79.5 & 76.5 & 81.5 & 79.2\\
\midrule
\multicolumn{5}{l}{\textit{Video-conditioned policies}}\\
\midrule
VQ-BeT / VideoMAE & 44.0 & 24.5 & 20.5 & 29.7\\
VQ-BeT / SigLIP2 & 58.5 & 56.0 & 58.0 & 57.5\\
VQ-BeT / DINOv2 & 45.0 & 52.0 & 58.5 & 51.8\\
DP / VideoMAE & 23.5 & 17.0 & 2.5 & 14.3\\
DP / SigLIP2 & 60.5 & 58.5 & 63.0 & 60.7\\
DP / DINOv2 & 65.0 & 68.0 & 68.5 & 67.2\\
ACT / VideoMAE & 80.5 & 69.0 & 67.0 & 72.2\\
ACT / SigLIP2 & \second{82.0} & 76.5 & 77.5 & 78.7\\
ACT / DINOv2 & 80.5 & \second{81.5} & 80.5 & \second{80.8}\\
\midrule
\multicolumn{5}{l}{\textit{In-context policies}}\\
\midrule
Zero-WAM & 68.0 & 77.0 & 68.0 & 71.0\\
\textbf{RoboDemo(Ours)} & \best{89.5} & \best{94.0} & \best{88.5} & \best{90.7}\\
\bottomrule
\end{tabular}
}
\vspace{-5pt}
\end{table}
\endgroup

%% file: tables/imitationgame_zero_shot.tex
\begingroup
\input{tables/experiment_style}
\begin{table}[t]
\vspace{-5pt}
\caption{\textbf{Zero-shot generalization to unseen tasks.}
We report SR (\%) on five unseen tasks under the C45 setting, with results broken down by scene level (L0--L3).
And ``Avg.'' denotes the mean SR across all five tasks and four scene levels.
}
\label{tab:zero-shot}
\vspace{-8pt}
\centering
\footnotesize
\resizebox{1.0\columnwidth}{!}{%
\tabcolsep=0.4cm
\renewcommand\arraystretch{1.0}
\begin{tabular}{l|*{5}{>{\centering\arraybackslash}p{0.74cm}}}
\toprule
Method & L0 & L1 & L2 & L3 & Avg.\\
\midrule
GR00T-N1.6 & 2.0 & 0.0 & \best{20.0} & 0.0 & 5.5\\
OpenVLA & 4.0 & 0.0 & \best{20.0} & 0.0 & 6.0\\
$\pi_{0.5}$ & 2.0 & \best{20.0} & 0.0 & 18.0 & 10.0\\
RDT-1B & 0.0 & 0.0 & \second{18.0} & 0.0 & 4.5\\
\midrule
UniSkill & 18.0 & 0.0 & 0.0 & 0.0 & 4.5\\
XSkill & 10.0 & 6.0 & 2.0 & 2.0 & 5.0\\
\midrule
ACT & 4.0 & 0.0 & 0.0 & 6.0 & 2.5\\
DP & 2.0 & 0.0 & 10.0 & 0.0 & 3.0\\
VQ-BeT & 0.0 & \best{20.0} & 8.0 & 2.0 & 7.5\\
\midrule
Zero-WAM & \best{30.0} & \second{14.0} & \second{18.0} & \best{26.0} & \best{22.0}\\
\textbf{RoboDemo} & \second{22.0} & 2.0 & 2.0 & \second{20.0} & \second{11.5}\\
\bottomrule
\end{tabular}
}
\vspace{-15pt}
\end{table}
\endgroup

%% file: tables/ablation_components.tex
\begingroup
\input{tables/experiment_style}
\newcommand{\abyes}{\ensuremath{\checkmark}}
\newcommand{\abno}{\ensuremath{\times}}
\begin{table}[t]
\vspace{-5pt}
\caption{\textbf{Context and memory organization on C15-Seen.}
Demo.(Pref.) denotes demonstration tokens as a VLM prefix.
Demo.(TTT) and Exec.(TTT) denote demonstration memory and execution progress in TTT, respectively.
In row 4, Demo. and Exec. share a single TTT module. 
In rows 5 and 6, they use two separate TTT modules connected in parallel and in cascade, respectively.}
\label{tab:ablation-components}
\vspace{-8pt}
\centering
\footnotesize
\resizebox{1.0\columnwidth}{!}{%
\tabcolsep=0.2cm
\renewcommand\arraystretch{1.0}
\begin{tabular}{ccccc|c}
\toprule
+Demo.(Pref.) & +Demo.(TTT) & +Exec.(TTT) & Parallel & Cascade & C15(Seen)\\
\midrule
\abno & \abno & \abno & -- & -- & 62.0\\
\abyes & \abno & \abno & -- & -- & 68.5\\
\abno & \abyes & \abno & -- & -- & 71.5\\
\abno & \abyes & \abyes & \abno & \abno & 73.0\\
\abno & \abyes & \abyes & \abyes & \abno & \second{82.0}\\
\abno & \abyes & \abyes & \abno & \abyes & \best{89.5}\\
\bottomrule
\end{tabular}
}
\vspace{-15pt}
\end{table}
\endgroup

%% file: tables/inference_efficiency.tex
\begingroup
\input{tables/experiment_style}
\begin{table}[t]
\vspace{-15pt}
\caption{\textbf{Model size and inference time on H200 (batch size 1).}
Q, E, and P denote costs per policy query, episode, and retrieval pair, respectively.
Retrieval uses precomputed features.
The total includes reference initialization and one query, excluding retrieval.}
\label{tab:inference-efficiency}
\vspace{-8pt}
\centering
\footnotesize
\resizebox{1.0\columnwidth}{!}{%
\tabcolsep=0.4cm
\renewcommand\arraystretch{1.0}
\begin{tabular}{l|cc}
\toprule
Component & Params. & Time (ms)\\
\midrule
VLM backbone\textsuperscript{Q} & 4.438B & 64.7\\
Original action expert\textsuperscript{Q} & 0.675B & $28.8\times5$\\
Dual-TTT\textsuperscript{Q} & \shortstack{Slow: 189.7M\\Fast: $37.9\mathrm{M}\times2$} & \shortstack[l]{Write: $55.1\times1$\\Read: $59.0\times5$}\\
Reference motion encoder\textsuperscript{E} & 1.012B & 290.5\\
Semantic encoding + preload\textsuperscript{E} & Shared & 364.8\\
Refreward\textsuperscript{P} & 7.4M & 5.3\\
\midrule
\textbf{RoboDemo (E + Q)} & 6.35B & 1.22s\\
\bottomrule
\end{tabular}
}
\vspace{-5pt}
\end{table}
\endgroup

%% file: tables/imitationgame_reference_source.tex
\begingroup
\input{tables/experiment_style}
\begin{table}[t]
\vspace{-5pt}
\caption{\textbf{Human versus retrieved robot references on C15-Seen.}
}
\label{tab:reference-source}
\vspace{-8pt}
\centering
\footnotesize
\resizebox{1.0\columnwidth}{!}{%
\tabcolsep=0.15cm
\renewcommand\arraystretch{1.0}
\begin{tabular}{l|cccccc}
\toprule
Method & Fold-towel & Place-file & Hang-mug & Place-plate & stir-spoon & Avg.\\
\midrule
$\pi_{0.5}$ & 43.5 & 40.0 & 90.5 & 85.5 & 55.0 & 62.9\\
ACT & \second{90.0} & \second{80.5} & \second{93.5} & 80.0 & 68.5 & 82.5\\
Zero-WAM & 88.0 & 43.5 & 43.0 & 80.5 & \best{88.0} & 68.6\\
\textbf{RoboDemo} & \best{92.5} & 75.0 & 92.5 & \best{100.0} & \second{87.5} & \second{89.5}\\
\textbf{RoboDemo-wo} & \best{92.5} & \best{90.0} & \best{95.0} & \second{97.5} & 75.0 & \best{90.0}\\
\bottomrule
\end{tabular}
}
\vspace{-15pt}
\end{table}
\endgroup

%% file: segments/real_world_evaluation.tex
\subsection{Real-World Evaluation}
\label{sec:real-world}
\paragraph{Task design, setup and standard results.}
We construct three common household scenarios on the AgiBot-G2 humanoid.
\emph{Tabletop Organization} places pens, tea bags, and toy cars into a storage rack, testing pick-and-place across varied object shapes and grasp affordances.
\emph{Bottle-to-Cup Pouring} pours water from a bottle into a teacup, testing grasp stability, bottle--cup alignment, and continuous orientation control.
\emph{Sofa Cushion Arrangement} restores displaced cushions to their intended positions and orientations, testing rearrangement of bulky, deformable objects over a larger workspace.
Together, these scenarios cover object transfer, continuous manipulation, and larger-scale scene organization beyond tabletop pick-and-place.
We evaluate each method over 50 trials per standard task or generalization.
Figure~\ref{fig:real-world-evaluation}(a) shows that RoboDemo leads on all three tasks, achieving 90\% on tabletop organization, 68\% on pouring, and 62\% on cushion arrangement.
Its average SR reaches 73.3\%, exceeding the strongest baseline, $\pi_{0.5}$+Demo(Prefix), by 20.0 percentage points.
These results support demonstration-guided execution across diverse household tasks, although pouring and cushion arrangement remain more challenging.

\input{tables/real_world_generalization}

\vspace{-10pt}
\paragraph{Generalization transfer.}
We use the broadly applicable Tabletop Organization task to test visual generalization, keeping the manipulation objective fixed while changing illumination or the tabletop background.
RoboDemo and all baselines are trained in the standard scene, then directly evaluated under each changed condition for 50 trials with all learned parameters frozen.
This protocol evaluates direct generalization from the standard scene to altered lighting and backgrounds.
As shown in Table~\ref{tab:real-world-generalization}, RoboDemo reaches 46\% under changed lighting and 76\% under a changed background, exceeding the strongest baseline by 14 and 34 points.
Relative to its 90\% standard-scene score, it retains 51.1\% and 84.4\% of its performance, respectively. The demonstration-prefix baseline retains 47.1\% and 61.8\% relative to its own 68\% reference.
RoboDemo outperforms the baselines under both visual shifts, demonstrating improved visual generalization. 
Performance is more stable under background changes, with a 14 points drop from the standard scene compared with 44 points under lighting changes.

%% file: tables/real_world_generalization.tex
\begingroup
\input{tables/experiment_style}
\begin{table}[t]
\vspace{-15pt}
\caption{\textbf{Real-world visual generalization.}
Tabletop Organization SR (\%), with 50 trials per method and condition.
All policies are trained in the standard scene and evaluated with learned parameters frozen, without target-condition fine-tuning.
$\pi_{0.5}$+Demo prepends demonstration tokens as a prefix.}
\label{tab:real-world-generalization}
\vspace{-8pt}
\centering
\footnotesize
\resizebox{1.0\columnwidth}{!}{%
\tabcolsep=0.3cm
\renewcommand\arraystretch{1.0}
\begin{tabular}{l|ccccc}
\toprule
Condition & GR00T-N1.7 & $\pi_0$ & $\pi_{0.5}$ & $\pi_{0.5}$+Demo & RoboDemo\\
\midrule
Lighting & 20 & 18 & 32 & 32 & 46\\
Background & 24 & 26 & 38 & 42 & 76\\
\bottomrule
\end{tabular}
}
\vspace{-15pt}
\end{table}
\endgroup

%% file: segments/05_discussion.tex
\section{Conclusion and Limitation}\label{sec:conclusion}

\paragraph{Conclusion.}
We present RoboDemo, which separates demonstration intent and interaction progress through Dual-Cascaded Test-Time Training.
Cascaded memory reads integrate both contexts for action prediction while keeping learned policy parameters frozen.
RoboDemo achieves 90.7\% mean Seen-task success on Imitator Game and 73.3\% across three real-world household tasks.

\vspace{-8pt}
\paragraph{Limitation.}
RoboDemo uses fixed reference-sampling budgets and memory-register counts.
Adapting these budgets to task complexity and developing lightweight memory diagnostics could improve flexibility and interpretability.

%% file: segments/statements.tex
\subsection*{AI Use Statement}
Generative AI assisted manuscript drafting and editing, literature discovery and source cross-checking, experimental-protocol design, figure, and document-code scaffolding.

\subsection*{Reproducibility Statement}
The method and appendix specify tensor shapes, update order, gating, optimization, data conversion, seeds, metrics, and retrieval supervision. The reported C15/C30/C45 and Full50 results and retained retrieval-annotation counts are tied to saved experiment records.

%% file: segments/appendix_layout.tex
\setlength{\multicolsep}{6pt plus 2pt minus 1pt}
\setlength{\premulticols}{48pt}
\setlength{\postmulticols}{12pt}

\newcommand{\appendixsupplementimage}[1]{%
  \par\begingroup
  \ifdefined\nolinenumbers\nolinenumbers\fi
  \noindent
  \ifdefined\AppendixPrivacyPreview
    \includegraphics[draft,width=\linewidth]{images/#1}%
  \else
    \includegraphics[width=\linewidth]{images/#1}%
  \fi
  \par\endgroup
  \vspace{6pt}%
}

\makeatletter
\@ifpackageloaded{lineno}{%
  \newcommand{\appendixcolumnlinenumber}{%
    \logtheLineNumber
    \getLineNumber
    \docolaction{\makeLineNumberLeft}%
      {\makeLineNumberLeft}{\makeLineNumberRight}%
  }
  \AtBeginEnvironment{multicols}{%
    \let\makeLineNumber\appendixcolumnlinenumber
  }
}{}
\makeatother

\newenvironment{appendixalgorithm}{%
  \par\addvspace{\intextsep}%
  \begingroup
  \captionsetup{type=algorithm,position=top,skip=4pt}%
  \footnotesize
  \hrule height 0.8pt\relax
  \nopagebreak
}{%
  \par\nopagebreak
  \kern 2pt\hrule height 0.8pt\relax
  \endgroup
  \par\addvspace{\intextsep}%
}

\newenvironment{appendixresulttable}{%
  \par\addvspace{\intextsep}%
  \begingroup
  \ifdefined\nolinenumbers\nolinenumbers\fi
  \noindent\begin{minipage}{\linewidth}%
  \captionsetup{type=table,position=top,width=\linewidth,skip=6pt}%
}{%
  \par\end{minipage}\par
  \endgroup
  \addvspace{\intextsep}%
}

%% file: segments/06_more_related_work.tex
\section*{Extended Related Work}
\label{app:more-related-work}

\subsection*{Future progress modeling in VLAs and WAMs}
Vision--language--action policies map visual observations and language instructions to autoregressive, diffusion, or flow-based actions~\citep{brohan2023rt2,octo2024,kim2024openvla,black2024pi0,liu2024rdt1b}. Embodied Chain-of-Thought and related reasoning policies introduce intermediate language, spatial grounding, and subtask predictions to structure action generation~\citep{zawalski2024ecot,zhang2025cotvla}. Video-based context complements these symbolic descriptions with continuous geometry, object motion, and contact evolution.

World action models couple action prediction with future visual dynamics. GR-2, DreamGen, and unified video-action models learn future observations together with inverse dynamics or joint action generation~\citep{cheang2024gr2,jang2025dreamgen,li2025unifiedvideoaction}. Subsequent approaches improve temporal consistency, masked visual-action modeling, latent dynamics, and progress prediction~\citep{yu2026maskwam,lyu2026lda,zhu2026prowam}. Fast-WAM and LaWAM investigate efficient predictive representations~\citep{yuan2026fastwam,li2026lawa}, while other studies examine the relationship between generated visual trajectories and executable robot actions~\citep{yang2026genvid2robot,li2026badwam}. RoboDemo obtains procedural guidance from an observed reference video and grounds it in completed robot interactions through a separate execution memory.

\subsection*{Reference-video and retrieval-augmented control}
Demonstration-conditioned policies use observed behavior to communicate a task's intended evolution. MimicPlay and Vid2Robot condition control on human or robot videos, and R+X retrieves human clips containing task-relevant motion~\citep{wang2023mimicplay,jain2024vid2robot,papagiannis2025rx}. StellaVLA organizes retrieved trajectories into procedural context~\citep{xu2026stellavla}, while ICI-VLA retrieves spatiotemporally aligned micro-demonstrations~\citep{yang2026icivla}. RA-VLA combines behavior-aligned retrieval with execution~\citep{jang2026ravla}. ReCAP retrieves cheap-embodiment trajectories to extend a frozen policy~\citep{park2026recap}. WAM-TTT absorbs human-video observations into adaptive fast weights through self-supervised video prediction~\citep{feng2026wamttt}. Zero-WAM scales video prompting with paired human--robot video pretraining and a causal world action model~\citep{zeroWAM2026}.

RoboDemo uses a provided human demonstration or a robot demonstration selected by Refreward. H-GEC samples the complete reference video, and semantic--motion encoding compresses the selected evidence into a fixed-size demonstration-intent memory. The reference remains available throughout execution, while an independent interaction-progress memory records completed transitions. The action expert reads reference intent first and then refines that representation using realized progress. This organization connects demonstration guidance to closed-loop control through explicit state ownership and serial memory reads.

\subsection*{Long-context memory and test-time compression}
Transformer-XL propagates segment-level hidden states, the Memorizing Transformer retrieves earlier key--value pairs, and Recurrent Memory Transformer carries learned memory tokens across segments~\citep{dai2019transformerxl,wu2022memorizing,bulatov2022rmt}. TTT layers instead use the parameters of an inner learner as recurrent state, compressing observations into fast weights queried by subsequent tokens~\citep{sun2024ttt}. PI-Mem performs bounded iterative aggregation over chunks processed in parallel~\citep{liu2026pimem}. These approaches offer complementary mechanisms for retaining and accessing long-context information.

Robot-memory methods specialize these ideas for partial observability and long-horizon manipulation. MEM combines video-based short-term memory with textual long-term summaries~\citep{tang2026mem}. NativeMEM compresses historical frames into native VLA tokens~\citep{wang2026nativemem}. MemoryWAM combines recent frames, event anchors, and long-range summaries~\citep{yang2026memorywam}. LaMem-VLA separates short- and long-term latent stores~\citep{qu2026lamemvla}, while UniMem integrates multimodal history into control~\citep{osterberg2026unimem}. WorldScape Policy 2.0 and MaP-WAM connect history to progress-aware planning~\citep{wu2026worldscape,zhao2026mapwam}. SimpleMemVLA retains timestamped video and amortizes its processing through prefix caching~\citep{yin2026simplememvla}.

In-context imitation methods use demonstrations to specify target behavior within a contextual inference space~\citep{jiang2023vima,fu2024icil,ding2026contextflow}. RoboTTT extends this approach to long visuomotor streams using fast-weight compression, sequence action forcing, and truncated backpropagation~\citep{jiang2026robotttt}. RoboDemo organizes its two memories by semantic role: the reference stores the intended procedure, and the execution stream stores realized interaction. Both use the same write/read operations with independent numerical states. Video-derived registers write those states. The action expert's sink, state, and noised-action tokens query them through cascaded reads. This separation preserves a fixed-length action interface while execution history accumulates.

%% file: segments/appendix.tex
\makeatletter
\providecommand{\theHALG@line}{}
\renewcommand{\theHALG@line}{\thealgorithm.\arabic{ALG@line}}
\makeatother
\section{Exact Architecture and Configuration}
\label{app:architecture}

This appendix records the architecture, memory updates, and implementation settings of RoboDemo. The names RetWriter/ExeWriter and RetReader/ExeReader identify the reference/execution applications of the shared TTT-Writer and TTT-Reader. The distinct video encoders precede the common write operation.

\subsection{Configuration summary}

Table~\ref{tab:all-hyperparameters} groups the configuration into model architecture, training, and benchmark/data settings. Derived tensor lengths and dataset counts are included as implementation references rather than tuned hyperparameters.

\input{tables/hyperparameters}

At one control step, the Action-DiT input is
\begin{equation}
H^0=[h_{\mathrm{sink}},h_{\mathrm{state}},h_{\mathrm{action},1:30}]
\in\mathbb R^{B\times32\times1024}.
\end{equation}
The current RGB observation and instruction are encoded by frozen Qwen3-VL and exposed as per-layer keys and values to the base attention path. The reference and execution writers use the same RGB view.

\subsection{Motivation: complementary evidence under a fixed writer budget}
\label{app:semantic-motion-motivation}

A reference demonstration must communicate both \emph{what} is present and \emph{how} it evolves. An isolated image can reveal object identity, scene layout, grasp geometry, and the robot's instantaneous pose, but in general it cannot identify the sign or magnitude of motion. For example, the same frame can be consistent with a gripper that is closing or opening. Velocity requires at least a temporal difference, $v_i\approx(x_i-x_{i-1})/\Delta t_i$. Conversely, a video backbone is well suited to direction, velocity, contact change, and short trajectory geometry, but need not retain the task-aligned object semantics of a vision--language model. We therefore treat appearance semantics and physical dynamics as complementary evidence rather than asking one encoder to approximate both.

The resulting design has two constraints. First, RetWriter remains prompt-independent and visual-only: language, robot state, and noised Action-DiT tokens never define the memory target that they subsequently query. Second, increasing the number of sampled frames must not increase the online TTT interface. A frozen Qwen3-VL branch supplies spatial, vision--language-pretrained evidence. A frozen V-JEPA2 branch supplies local spatiotemporal evidence. Learned Resamplers reduce both variable token sets to $R_R=32$ register tokens before fusion. V-JEPA2 is the temporal encoder~\citep{qwen3vl2025,assran2025vjepa2}.

\clearpage
\appendixsupplementimage{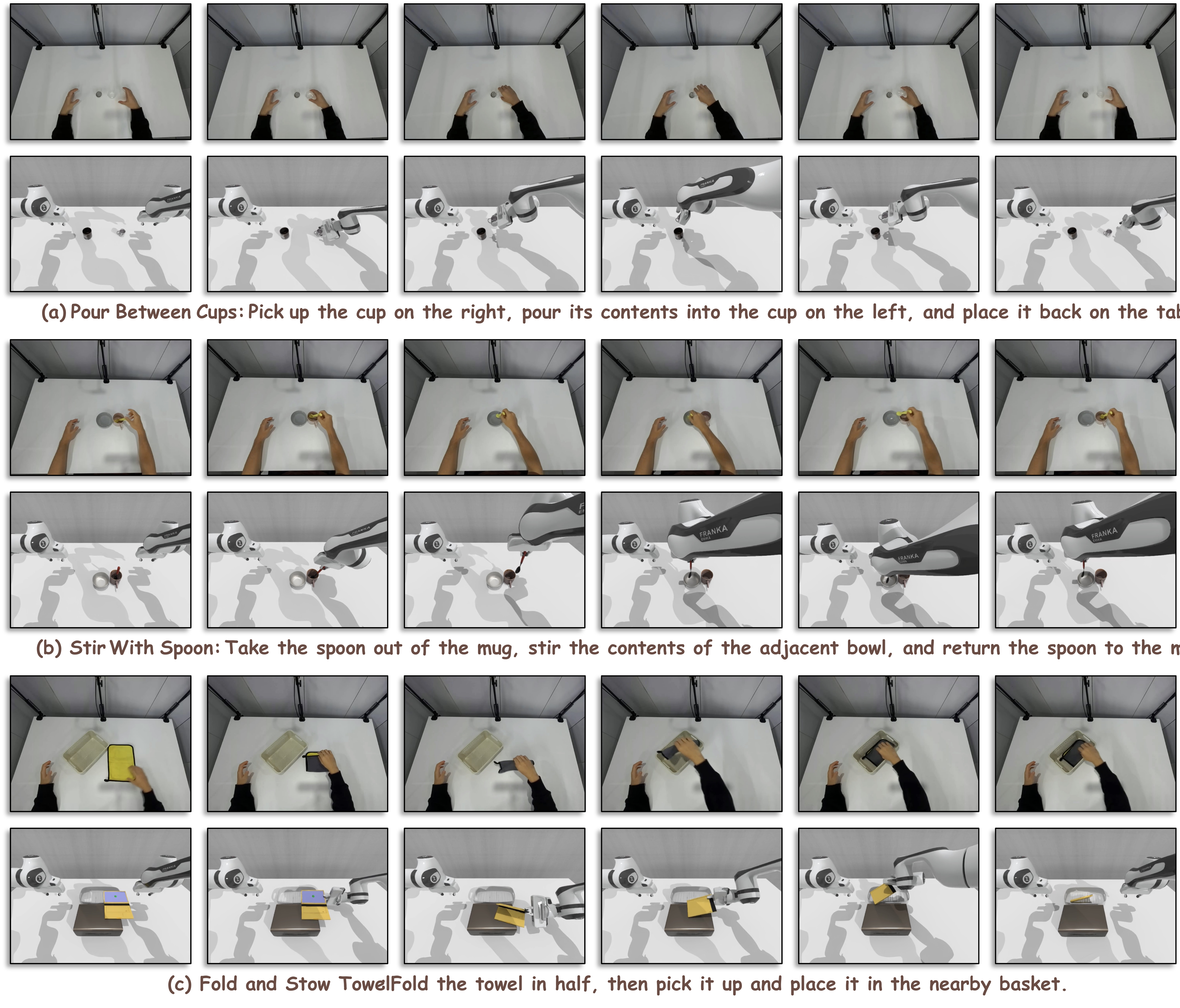}
\subsection{Hierarchical Global--Event--Clip sampling}
\label{app:hybrid12}

H-GEC takes the complete reference video containing $N$ chronological observations as input. For frame index $i\in\{0,\ldots,N-1\}$, let $I_i$ be the ego-view RGB image, $t_i$ its timestamp in seconds, $s_i$ the synchronized robot-state vector when available, and $g_i$ the gripper coordinate when available. The reference observations are collected in $\mathcal V$:
\begin{equation}
\mathcal V=\{(I_i,t_i,s_i,g_i)\}_{i=0}^{N-1},
\qquad t_{i+1}\ge t_i,
\end{equation}
Global and event frames are selected over the full temporal extent of $\mathcal V$, together with an ordered local clip around a selected event.

\paragraph{Event evidence.}
We compute an offline event score from visual motion, robot-state change, and gripper change. For adjacent frames, define the elapsed time $\Delta t_i=\max(t_i-t_{i-1},10^{-6})$. Convert the two ego images to grayscale and resize them to height $H_{\mathrm{img}}=96$ and width $W_{\mathrm{img}}=96$. Let $F_i(x,y)$ be their Farneb\"ack dense optical-flow vector at pixel coordinates $(x,y)$. The visual-motion rate $v_i^{\mathrm{flow}}$ is the mean flow magnitude divided by elapsed time.

\clearpage
\appendixsupplementimage{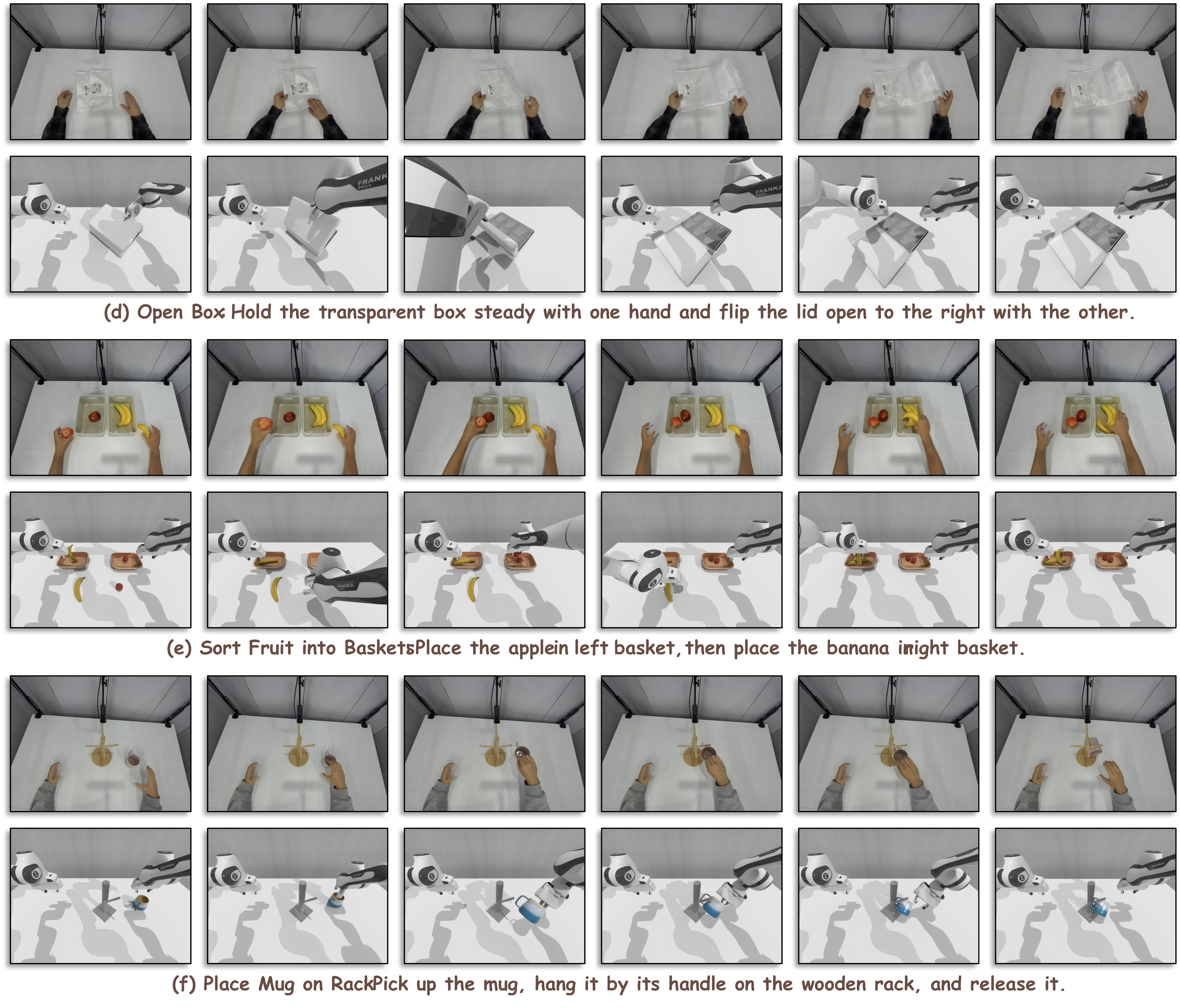}
The robot-motion rate $v_i^{\mathrm{robot}}$ and gripper-change rate $v_i^{\mathrm{grip}}$ use the corresponding temporal differences:
\begin{align}
v_i^{\mathrm{flow}}&=\frac{\sum_{x=1}^{W_{\mathrm{img}}}\sum_{y=1}^{H_{\mathrm{img}}}\|F_i(x,y)\|_2}
{\Delta t_iH_{\mathrm{img}}W_{\mathrm{img}}},\\
v_i^{\mathrm{robot}}&=\frac{\|s_i-s_{i-1}\|_2}{\Delta t_i}, &
v_i^{\mathrm{grip}}&=\frac{|g_i-g_{i-1}|}{\Delta t_i},
\end{align}
All three rates are zero for $i=0$. Optical flow measures image displacement, and the three rates prioritize moments of pronounced motion. For cue $x\in\{\mathrm{flow},\mathrm{robot},\mathrm{grip}\}$, let $v^x=(v_0^x,\ldots,v_{N-1}^x)$ collect its video-wide rates, and let $P_{10}(v^x)$ and $P_{90}(v^x)$ denote their 10th and 90th percentiles. The normalized cue $s_i^x$ uses clipping to $[0,1]$:
\begin{equation}
s_i^x=
\begin{cases}
\operatorname{clip}\!\left(\dfrac{v_i^x-P_{10}(v^x)}{P_{90}(v^x)-P_{10}(v^x)},0,1\right),
&P_{90}(v^x)>P_{10}(v^x),\\[6pt]
0,&\text{otherwise,}
\end{cases}
\end{equation}
The combined event score $s_i^{\mathrm{event}}$ is:
\begin{equation}
s_i^{\mathrm{event}}=
0.40s_i^{\mathrm{flow}}+0.35s_i^{\mathrm{robot}}+0.25s_i^{\mathrm{grip}}.
\label{eq:hybrid-event-score}
\end{equation}
A high score indicates pronounced visual motion, robot-state change, or gripper movement and is used to prioritize potentially informative interactions. When telemetry is unavailable, event selection uses the visual-motion cue alone. Telemetry affects only offline frame selection. It is not passed to either Retrieval-TTT encoder branch or the TTT writer.

\paragraph{Global level.}
Global frames preserve video-wide scene evolution and the coarse manipulation process. Let $K_G$ be their count and $j\in\{1,\ldots,K_G\}$ index the target timestamps $\tau_j^{G}$. We use interior video-relative targets, not the endpoints:
\begin{equation}
\tau_j^{G}=t_0+\frac{j}{K_G+1}(t_{N-1}-t_0),
\qquad j=1,\ldots,K_G.
\end{equation}
Each target selects its nearest source timestamp. Repeated nearest neighbours are removed, then the unused frame farthest in time from the current selected set is inserted until $K_G$ unique frames are obtained. Only a genuinely short video is padded, and padded slots are marked invalid rather than treated as repeated observations. The resulting set is $\mathcal G$.

\paragraph{Event level.}
Event frames concentrate on sparse moments of pronounced interaction. Let $K_E$ be the desired count and $\mathcal E$ the selected set. Candidate frames are stably sorted by decreasing $s_i^{\mathrm{event}}$. Frames already in $\mathcal G$ are skipped, and greedy temporal non-maximum suppression initially requires every pair of selected event frames to be separated by at least $0.4$\,s. If this yields fewer than $K_E$ events, the separation constraint is relaxed while uniqueness with respect to global and previously selected event frames is retained. Any remaining shortfall is filled from the largest uncovered temporal gaps. The highest-scoring selected event defines the clip centre $t^\star$. If no event candidate remains, the highest-scoring valid video frame is used.

\paragraph{Clip level.}
Global and event frames may show an open and a closed gripper without resolving whether the local motion is opening or closing. The temporal branch therefore receives a chronological local clip rather than sparse events falsely presented as adjacent video. Let $K_C$ be the clip-frame count and $f_c$ the target sampling rate. The first-to-last target span $T_c$ is
\begin{equation}
T_c=\frac{K_C-1}{f_c}.
\end{equation}
Let $a_{\mathrm{clip}}$ denote the video-bounded clip start and $\tau_k^{C}$ its target timestamp at position $k\in\{0,\ldots,K_C-1\}$:
\begin{align}
a_{\mathrm{clip}}&=\operatorname{clip}\!\left(t^\star-\frac{T_c}{2},\ t_0,\ \max(t_0,t_{N-1}-T_c)\right),\\
\tau_k^{C}&=a_{\mathrm{clip}}+\frac{k}{f_c},\qquad k=0,\ldots,K_C-1.
\end{align}
Nearest physical frames form the ordered clip $\mathcal C$. A clip-validity mask $m^C$ invalidates boundary overflow and repeated nearest neighbours. The clip never crosses the video boundary. The target timestamps are equally spaced, although nearest-frame matching can introduce deviations when source timestamps are irregular. Sparse global/event frames supply Qwen3-VL with scene, object, and pose evidence, while the ordered clip supplies V-JEPA2 with local motion direction, relative speed, and trajectory-continuity cues.

\paragraph{Unique physical observations.}
The three H-GEC counts describe complementary roles, not necessarily disjoint initial selections: the strongest event can also appear in $\mathcal C$. The sparse semantic set initially contains $\mathcal G\cup\mathcal E$. Whenever it overlaps the clip, additional unused semantic backfill frames are chosen in decreasing event-score order, with largest-gap fallback, until
\begin{equation}
\left|\mathcal G\cup\mathcal E\cup\mathcal B\cup\mathcal C\right|
=\min(N,K_G+K_E+K_C).
\end{equation}
The Qwen branch therefore receives the valid sparse semantic rows $\mathcal G\cup\mathcal E\cup\mathcal B$, while V-JEPA2 receives the $K_C$ ordered clip slots. Arrays use explicit masks for unavailable observations. Padding never writes fast weights. Let $F_S$ denote the padded sparse-semantic frame count and $F_M=K_C$ the ordered clip length for the complete reference input.

\subsection{Sparse-semantic, dense-motion joint encoding}
\label{app:dual-stream-writer}

\paragraph{Qwen3-VL semantic branch.}
Every valid sparse frame is encoded independently by the frozen Qwen3-VL-4B vision tower. We tap the visual stack after its spatial merger and before the language Transformer, yielding prompt-independent spatial tokens rather than a pooled image vector. Under deterministic image preprocessing, one RGB frame gives
\begin{equation}
Z_k^{Q}\in\mathbb R^{N_Q\times D_v},
\qquad D_v=2560.
\end{equation}
Here $N_Q$ is the post-merger spatial token count determined by image preprocessing. The vision tower is permanently frozen and in evaluation mode. Cache construction, training, and rollout use the same preprocessing, output interface, and BF16 storage contract. For patch $n$ in sparse frame $k$, we form
\begin{equation}
X_{k,n}^{S}=P_ZZ_{k,n}^{Q}
+e_{\mathrm{view}}(v)
+e_{\mathrm{patch}}(r_{k,n},c_{k,n})
+e_{\mathrm{time}}(t_k-t_{\mathrm{demo},0}),
\label{eq:semantic-token-embedding}
\end{equation}
where $P_Z:\mathbb R^{2560}\rightarrow\mathbb R^D$, patch row and column are normalized to $[0,1]$, and time uses a continuous sinusoidal code followed by a learned projection. Let $S$ denote the flattened valid sequence of these embedded Qwen tokens and $m^S$ its mask. Separate learned semantic queries $Q_{\mathrm{reg}}^S\in\mathbb R^{R_R\times D}$ jointly attend to all selected sparse frames and their patches:
\begin{align}
H^S&=Q_{\mathrm{reg}}^S+\operatorname{CrossAttn}\!\left(
\operatorname{LN}_q(Q_{\mathrm{reg}}^S),
\operatorname{LN}_{kv}(S),m^S\right),\\
A^S&=\operatorname{LN}_{\mathrm{out}}\!\left(
H^S+\operatorname{FFN}(\operatorname{LN}_{\mathrm{post}}(H^S))\right)
\in\mathbb R^{B\times32\times D}.
\label{eq:semantic-resampler}
\end{align}
The learned register tokens can specialize to different objects, contact regions, poses, or procedural summaries without imposing hand-written roles.

\paragraph{V-JEPA2 temporal branch.}
The ordered clip is stacked as
\begin{equation}
C\in\mathbb R^{B\times F_M\times3\times H\times W}.
\end{equation}
Frames are mapped to RGB $[0,1]$, resized with short side $256/224$ times the target size, centre-cropped to $256\times256$, normalized by ImageNet statistics, and rearranged to $[B,3,F_M,256,256]$. A frozen V-JEPA2 ViT-G with tubelet size two returns pre-pool spatiotemporal tokens
\begin{equation}
M\in\mathbb R^{B\times N_M\times D_m},
\qquad D_m=1408.
\end{equation}
Here $N_M$ is the number of spatiotemporal tokens, determined by the temporal tubelet count and the $16\times16$ spatial grid. The encoder runs in deterministic evaluation and inference mode and its output is detached. After LayerNorm and projection to $D$, separate learned motion queries $Q_{\mathrm{reg}}^M\in\mathbb R^{R_R\times D}$ produce
\begin{align}
\widetilde M&=P_M\operatorname{LN}_M(M),\\
A^M&=\operatorname{LN}_{M,\mathrm{out}}\!\left(
Q_{\mathrm{reg}}^M+\operatorname{CrossAttn}(Q_{\mathrm{reg}}^M,\widetilde M,\widetilde M,m^M)
\right)
\in\mathbb R^{B\times32\times D}.
\label{eq:motion-resampler}
\end{align}
Unlike sparse event images, the fixed-rate clip permits the temporal backbone to represent motion direction, relative speed, gripper opening or closing, contact evolution, and local path geometry.

\paragraph{Dimension-wise gated residual fusion.}
The semantic and motion resamplers use separate learned query banks
$Q_{\mathrm{reg}}^S,Q_{\mathrm{reg}}^M
\in\mathbb R^{R_R\times D}$.
Their validity masks exclude missing or padded source tokens.
The two branches compress features sampled from the complete reference
video into equal-shape registers
$A^S,A^M\in\mathbb R^{B\times R_R\times D}$,
where $B$ is the batch size, $R_R$ is the register count,
and $D$ is the hidden width.

Rather than concatenating these registers into a longer writer sequence,
we retain the semantic registers as the residual base and modulate the
motion contribution with a learned gate:
\begin{align}
G&=\sigma\!\left(W_g[A^S\Vert A^M\Vert A^S\odot A^M]\right),\\
U^R&=A^S+\tanh(\alpha_M)\,G\odot A^M,
\qquad U^R\in\mathbb R^{B\times R_R\times D}.
\label{eq:semantic-motion-fusion}
\end{align}
Here, $[\,\Vert\,]$ denotes channel-wise concatenation,
$\odot$ denotes element-wise multiplication,
and $\sigma$ is the sigmoid function.
The learned projection
$W_g:\mathbb R^{3D}\rightarrow\mathbb R^D$
is applied at each register position, producing a gate for every
register and hidden channel.
The interaction term $A^S\odot A^M$ allows the gate to depend on
both semantic and motion content.
The learned scalar $\alpha_M$ controls the overall motion contribution.
We initialize $\tanh(\alpha_M)$ to $0.01$ so that the semantic
representation initially dominates.

If the temporal stream is unavailable, the output reduces to $A^S$.
If only the motion stream is available, it uses $A^M$.
If neither stream is valid, the output and its validity mask are zero,
and the TTT write is skipped.
Both frozen encoders are prompt-independent:
their visual representations are related to the current instruction
through subsequent policy memory reads.

The complete reference-encoding path is therefore
\begin{equation}
\mathcal V
\xrightarrow{\mathrm{H\text{-}GEC}}
(\mathcal G\cup\mathcal E\cup\mathcal B,\mathcal C)
\xrightarrow{\mathrm{Qwen3\text{-}VL},\,\mathrm{V\text{-}JEPA2}}
(S,M)
\xrightarrow{\mathrm{Resample+Fuse}}
U^R
\xrightarrow{\mathrm{RetWriter}}
\phi_R.
\label{eq:retrieval-writer-flow}
\end{equation}
RetWriter writes the fused tokens $U^R$ once during reference preloading and retains the resulting state. Frame count and encoder token count affect offline representation quality, while the writer interface remains fixed in size.

\subsection{Algorithmic specification of sampling and encoding}
\label{app:hybrid12-algorithm}

\begin{appendixalgorithm}
\caption{H-GEC Sampling and joint encoding for the complete reference video}
\label{alg:hybrid12-writer}
\begin{algorithmic}[1]
\Require Reference video $\mathcal V=\{(I_i,t_i,s_i,g_i)\}_{i=0}^{N-1}$, counts $(K_G,K_E,K_C)$, clip rate $f_c$, frozen Qwen encoder $E_Q$, frozen V-JEPA2 encoder $E_M$, learned Resamplers and gate
\Ensure Retrieval-TTT register tokens $U^R\in\mathbb R^{B\times32\times D}$ and validity mask $m^R$
\State Compute the available motion cues from adjacent observations
\State Initialize the first-frame rates to zero
\State Robust-normalize each cue with $P_{10}/P_{90}$
\State Use Eq.~(\ref{eq:hybrid-event-score}) when telemetry is available and the visual cue otherwise
\State $\mathcal G\gets$ nearest unique frames to $t_0+\frac{j}{K_G+1}(t_{N-1}-t_0)$ for $j=1,\ldots,K_G$
\State Backfill a short $\mathcal G$ using frames with maximum distance to the already selected timestamps
\State $\mathcal E\gets\operatorname{GreedyNMS}(\operatorname{sortdown}(s^{\mathrm{event}}),K_E,0.4\,\mathrm{s},\,\mathcal G)$
\State If $|\mathcal E|<K_E$, relax NMS while preserving uniqueness, then use largest-gap backfill
\State $t^\star\gets$ timestamp of the highest-scoring event in $\mathcal E$, or of the highest-scoring valid video frame if $\mathcal E$ is empty
\State Compute video-bounded $a_{\mathrm{clip}}$ and targets $\tau_k^{C}=a_{\mathrm{clip}}+k/f_c$ for $k=0,\ldots,K_C-1$
\State Map the targets to nearest source frames to obtain ordered $\mathcal C$ and mask $m^C$
\State Add unused, high-score backfill frames $\mathcal B$ until $|\mathcal G\cup\mathcal E\cup\mathcal B\cup\mathcal C|=\min(N,K_G+K_E+K_C)$
\State $S\gets\operatorname{SemEmbed}(E_Q(\mathcal G\cup\mathcal E\cup\mathcal B))$ using Eq.~(\ref{eq:semantic-token-embedding})
\State $A^S\gets\operatorname{SemResampler}(Q_{\mathrm{reg}}^S,S,m^S)$ by Eq.~(\ref{eq:semantic-resampler})
\State $M\gets E_M(\mathcal C)$
\State $A^M\gets\operatorname{MotionResampler}(Q_{\mathrm{reg}}^M,M,m^M)$ by Eq.~(\ref{eq:motion-resampler})
\If{$m^S$ and $m^M$ both contain valid evidence}
  \State Compute $G$ and $U^R$ by Eq.~(\ref{eq:semantic-motion-fusion})
\ElsIf{only $m^S$ is valid} \State $U^R\gets A^S$
\ElsIf{only $m^M$ is valid} \State $U^R\gets A^M$
\Else \State $U^R\gets0$ and $m^R\gets0$
  \State Skip the TTT write
\EndIf
\State Set $m^R$ to one for registers constructed from valid evidence and zero for invalid outputs
\State \Return $U^R,m^R$
\end{algorithmic}
\end{appendixalgorithm}

\subsection{Online observation encoding and State-Action Adaptation}

At control step $t\geq1$, the execution encoder receives the latest
observation $o_t$ and robot state $s_t$, together with the action
$a_{t-1}^{\mathrm{exec}}$ actually executed during the preceding transition.
The frozen Qwen3-VL vision tower maps $o_t$ to visual tokens
$Z_t^{\mathrm{online}}$.
An execution-specific Q-former uses learned queries
$Q_{\mathrm{reg}}^E\in\mathbb R^{R_E\times D}$ to produce
a fixed-size visual representation:
\begin{equation}
\widehat U_t^E
=\operatorname{QFormer}_E
\left(Z_t^{\mathrm{online}},Q_{\mathrm{reg}}^E\right)
\in\mathbb R^{B\times R_E\times D}.
\end{equation}

State-Action-Conditioned Adaptive Layer Normalization (SA-AdaLN)
injects proprioception and executed-action information into these
visual registers.
Learned projections $P_s$ and $P_a$ map each input to width $D$.
The projection $P_{sa}:\mathbb R^{2D}\rightarrow\mathbb R^D$
fuses their concatenation, and learned heads $G_\gamma,G_\beta$
predict channel-wise scale and shift:
\begin{align}
h_t^s&=P_s(s_t),\qquad
h_t^a=P_a(a_{t-1}^{\mathrm{exec}}),\\
h_t^{sa}&=P_{sa}\!\left([h_t^s\Vert h_t^a]\right),\\
\gamma_t&=G_\gamma(h_t^{sa}),\qquad \beta_t=G_\beta(h_t^{sa}),\\
U_t^E&=(1+g_\gamma\odot\gamma_t)
\odot\operatorname{LN}(\widehat U_t^E)
+g_\beta\odot\beta_t.
\end{align}
The projections operate independently on each batch element.
Both $\gamma_t$ and $\beta_t$ have shape $B\times D$ and are
broadcast across the $R_E$ register positions.
The learned modulation gates $g_\gamma,g_\beta$ are initialized
to zero, so the initial output is the normalized visual representation.
Metadata dropout is applied before the output heads.

This conditioning changes register content without extending the
writer-token sequence.
The resulting $U_t^E$ is written to Execution-TTT once for the
completed transition, before sampling the current action.
Neither the current predicted action nor a future action target
is used as the executed-action input.

\section{Dual-Cascaded Test-Time Training}
\label{app:fastweight}

\subsection{Self-Memorization and Cross-Memorization}

\textbf{Self-Memorization} writes representations from the backbone sequence tokens into fast weights and then uses tokens from that same backbone sequence to query and read the adapted weights. This describes the recurring sequence-TTT template rather than every implementation detail of an individual prior method.

\textbf{Cross-Memorization} writes representations from video-derived memory tokens into fast weights and uses separate backbone sequence tokens to query and read the adapted weights. RoboDemo uses Cross-Memorization for both Retrieval-TTT and Execution-TTT: demonstration-intent and interaction-progress memory tokens write, while current sink, state, and noised-action tokens only read. The two upstream encoders have separate learned query banks. The TTT modules share the write/read operations but maintain independent fast states and lifetimes.

This organization forms Dual-Cascaded TTT (DC-TTT). Retrieval-TTT constructs a demonstration-intent memory from the reference demonstration video. Execution-TTT constructs an interaction-progress memory from the realized interaction history. The two memories are not merged into one state. RetReader is applied first, and ExeReader subsequently revises the reference-conditioned representation without overwriting Retrieval-TTT fast weights.

\subsection{Writing: RetWriter and ExeWriter}

Every Action-DiT block owns one TTT-MLP slow parameterization. Within a block, Retrieval-TTT and Execution-TTT share its key/value/query/output projections, initial two-layer fast MLP, inner normalization, and writer-dependent learning-rate predictor. They maintain distinct runtime fast states, role embeddings, learning-rate offsets, and bounded residual gates. Fast state is never shared across Action-DiT blocks, and both gates initialize elementwise at $0.1$ before the $\tanh$ bound. The shared TTT-Writer consumes either the fused demonstration-intent memory tokens or the SA-AdaLN-conditioned interaction-progress memory tokens. Semantic-motion fusion and state-action conditioning belong to the preceding encoders, not to different TTT write rules.

For one role and one Action-DiT block, let $U\in\mathbb R^{B\times R_r\times D}$ be writer registers and $m\in\{0,1\}^{B\times R_r}$ their mask. After splitting into $H$ heads,
\begin{align}
K_i&=\operatorname{norm}(U_iW_K), & V_i&=U_iW_V,\\
Y_i&=\operatorname{LN}(V_i-K_i), &
\widehat Y_i&=\operatorname{LN}(F_\phi(K_i)),\\
\eta_i^r&=\frac{\eta_0\,\sigma(g_\eta(U_i)+\delta_\eta^r)m_i}
{d_h\max(1,\sum_j m_j)}.
\end{align}
The two-layer fast model is
\begin{equation}
F_\phi(X)=\operatorname{GELU}(XW_1+b_1)W_2+b_2.
\end{equation}
The raw inner reconstruction losses are summed. Normalization is applied once through $\eta_i^r$. One write is
\begin{equation}
\phi^+=\phi-\sum_i\eta_i^r\nabla_\phi\frac12\|\widehat Y_i-Y_i\|_2^2.
\end{equation}
The implementation evaluates this gradient analytically with explicit differentiable tensor operations, so outer-loss gradients still reach the writer and slow TTT parameters. Fast parameters, inner normalization, and accumulated updates use FP32. The per-head update is jointly clipped by relative RMS to at most $0.05$ of the initial fast-parameter scale, using a $10^{-3}$ reference-scale floor. The clip scale is detached.

\subsection{Reading: RetReader and ExeReader}

Let $H$ denote the action-side tokens and
$\star\in\{r,e\}$ identify the reference or execution role.
We suppress the Action-DiT block index for clarity.
The learned role embedding $e_{\mathrm{read}}^\star$ conditions
the shared query projection:
\begin{equation}
q^\star=
\operatorname{norm}\!\left(W_Q(H+e_{\mathrm{read}}^\star)\right).
\end{equation}
The fast MLP $F_\phi$ processes these normalized queries.
Its response is normalized and added to the queries before
post-read normalization and projection back to the backbone width:
\begin{equation}
\operatorname{R}(\phi,H,\star)
=
W_O\!\left(
\operatorname{LN}_{\mathrm{post}}
\left[q^\star+
\operatorname{LN}_{\mathrm{in}}(F_\phi(q^\star))\right]
\right).
\end{equation}
Here, $\operatorname{LN}_{\mathrm{in}}$ denotes the inner
normalization used by the fast learner, and
$\operatorname{LN}_{\mathrm{post}}$ denotes post-read layer normalization.

The TTT-Reader returns the change relative to the learned
initialization $\phi_0$:
\begin{equation}
\operatorname{TTT\!-\!Reader}(H,\phi^\star,\star)
=
\operatorname{R}(\phi^\star,H,\star)
-
\operatorname{R}(\phi_0,H,\star).
\end{equation}
Both evaluations use identical queries, projections, normalization
parameters, and role embeddings. Only the fast state differs.
Consequently, an unwritten or reset state
$\phi^\star=\phi_0$ contributes exactly zero.
A written state contributes the response change induced by its
stored context, rather than the static response of the initialization.
Neither evaluation modifies the fast state.

RetReader and ExeReader apply this same operation to independent
reference and execution states.
Missing writer rows skip the update rather than evaluating
an all-masked resampler or changing the effective memory age.

\subsection{DC-TTT contract relative to sequence TTT}
\label{app:ttt-specialization}

Table~\ref{tab:ttt-specialization} compares a self-context sequence-TTT template with RoboDemo. Cross-context compression specifies the source of writer evidence, state ownership, lifetime, readout, and update timing for embodied control.

\begin{table}[H]
\centering
\footnotesize
\caption{How the implemented RoboDemo contract specializes a common sequence-TTT template.}
\label{tab:ttt-specialization}
\begin{tabularx}{\linewidth}{@{}>{\raggedright\arraybackslash}p{0.16\linewidth}>{\raggedright\arraybackslash}p{0.30\linewidth}>{\raggedright\arraybackslash}X@{}}
\toprule
Axis & Self-context sequence-TTT template & RoboDemo cross-context implementation \\
\midrule
Write ownership & tokens from the active sequence supply the inner objective & reference visual registers and state-action-conditioned execution registers write. Action-side sink, state, and noised-action tokens read \\
State layout & one chronological recurrent learner state & one shared slow learner per Action-DiT block, with one demonstration-intent state and one storage-independent interaction-progress state \\
State lifetime & state advances with one token stream & Retrieval-TTT is built once and immutable during rollout. Execution-TTT resets per episode and grows only from completed transitions \\
Inner step & self-supervised gradient update & raw-sum reconstruction update with role-conditioned per-token step sizes normalized by head width and the number of valid writers \\
Readout & absolute adapted-learner response & dynamic delta subtracts the response of the initial fast weights under the identical query and role embedding \\
Multiple memories & not intrinsic to the single-state template & RetReader reads the complete-reference state. ExeReader then queries the resulting representation \\
Mask and stability & implementation dependent & invalid writers are exact no-ops. Fast states and accumulation are FP32. A detached joint relative-RMS scale clips each per-head update \\
Online timing & update timing follows the incoming stream & predict from $\phi_{E,t}$, execute $a_t$, then commit the realized transition exactly once to obtain $\phi_{E,t+1}$ \\
\bottomrule
\end{tabularx}
\end{table}

The fast state is not a copy of the full policy and an online write does not run AdamW. Outer training differentiates through the analytic inner update to learn the initial fast MLP, projections, Resampler, step-size predictor, and residual gates. At rollout, those slow parameters are fixed. Only the per-stream tensors $\phi=(W_1,b_1,W_2,b_2)$ change through RetWriter or ExeWriter. This distinction is why RoboDemo can expose an explicit memory reset and commit boundary without treating every action as a full-policy fine-tuning step.

The changes above address different failure modes. \emph{Cross-Memorization} prevents the current query---which contains state and noised future-action tokens---from defining the memory target it immediately consumes. \emph{Independent state ownership} prevents the stable demonstration-intent memory from being overwritten by imperfect online experience. \emph{Dynamic-delta readout} removes the static response of the auxiliary TTT path, so an unwritten or reset memory is functionally neutral. An all-invalid write leaves an existing memory unchanged. It does not erase earlier valid writes. DC-TTT lets interaction progress revise the reference-conditioned representation rather than compete with it through a parallel sum.

Algorithms~\ref{alg:ttt-operations} and~\ref{alg:ttt-lifecycle} specify the TTT operations and their training and rollout lifecycles. The mask enters the learned step size itself. Consequently an all-invalid writer gives a zero analytic update before clipping. The clip is applied jointly over $(W_1,b_1,W_2,b_2)$ for each sample and head, with its scale detached while the analytic inner update remains connected to the outer graph. During training, an episode-indexed pool restores a detached numerical state before the current segment. Within that segment, each prediction reads after all preceding valid completed transitions have been committed. The terminal state is detached again before it is returned to the pool.

\input{segments/appendix_ttt_algorithms}

\section{Demonstration-Intent Memory}
\label{app:retrieval}

H-GEC samples the complete reference demonstration video, and the joint encoder compresses the selected observations into fixed-shape Retrieval-TTT register tokens $U^R$. In each Action-DiT block $l$, RetWriter starts from a storage-independent copy of the learned initialization and writes these tokens before rollout:
\begin{equation}
\phi_R^l=\operatorname{TTT\!-\!Writer}
\left(\operatorname{Copy}(\phi_0^l),U^R,r\right).
\end{equation}
The resulting reference state $\phi_R^l$ is retained for subsequent action generation and remains frozen throughout the rollout. Frame budget affects reference encoding, but does not change the retained fast-state shape.

\section{Interaction-Progress Memory and Causality}
\label{app:execution}

\subsection{Episode-indexed interaction-progress state carry}
\label{app:state-carry}

Training shuffles episodes but never their internal order. If episode $e$ contains segments $S_{e,0},\ldots,S_{e,K_e-1}$, segment $k$ is delivered only after segment $k-1$. A stable episode lane and an episode-indexed pool $\mathcal C^l[e]$ are maintained independently for every Action-DiT block $l$. At the start of a segment,
\begin{equation}
\phi_{e,k,0}^l=\begin{cases}
\phi_0^l,&k=0\text{ or reset},\\
\operatorname{StopGrad}(\mathcal C^l[e]),&k>0.
\end{cases}
\end{equation}
For the current eight-step segment, let $U^E_{e,k,j}$ encode the transition ending at local observation $j$, pairing that observation and state with the preceding executed action. The restored state includes any required boundary transition committed exactly once before the first prediction. Within the segment, transition $j-1$ is written through its endpoint-aligned tokens before target $j$ is read:
\begin{align}
\widehat a_{e,k,0}&=\pi_\theta(o_{e,k,0},\mathcal D,\phi_{e,k,0}),
\phi_{e,k,j}=\operatorname{TTT\!-\!Writer}(\phi_{e,k,j-1},U^E_{e,k,j},e),\\
\widehat a_{e,k,j}&=\pi_\theta(o_{e,k,j},\mathcal D,\phi_{e,k,j}),\qquad j=1,\ldots,7.
\end{align}
After the final valid transition, the terminal state is detached and committed:
\begin{equation}
\mathcal C^l[e]\leftarrow\operatorname{StopGrad}(\phi_{e,k,\mathrm{terminal}}^l).
\end{equation}
Numerical fast state can therefore persist for the complete episode, while the computation graph never crosses a segment boundary. Gradients reach all analytic writes in the current segment and no earlier segment.

\subsection{State-pool invariants}

Each pool record stores episode ID, expected next segment index, effective committed age, terminal flags, and the detached $(W_1,b_1,W_2,b_2)$ tensors for all layers and heads. A batch row may own at most one episode, and an episode may occupy at most one row/rank at a time. Restore fails closed on a missing noninitial record, duplicate episode ID, stale or skipped segment index, tensor-shape/configuration mismatch, or nonfinite state. Episode start initializes $\phi_0$. Episode termination and explicit reset evict the record. Retries may replace a pool entry only when the segment identity and input fingerprint match.

Distributed sampling assigns each complete episode lane to one rank and preserves chronological delivery. Each pool record identifies the next segment and the last committed transition, linking the stored numerical state to its position in the episode.

\begin{table}[H]
\centering
\footnotesize
\caption{Episode-indexed state-pool contract.}
\label{tab:state-pool-contract}
\begin{tabularx}{\linewidth}{@{}>{\raggedright\arraybackslash}p{0.24\linewidth}>{\raggedright\arraybackslash}p{0.34\linewidth}>{\raggedright\arraybackslash}X@{}}
\toprule
Boundary & Required invariant & Failure handling \\
\midrule
restore & matching episode and next segment with detached finite tensors & reject missing, stale, duplicate, or incompatible record \\
current segment & commit transition $t-1$, then predict $t$ within an eight-step graph & mask invalid writers without attaching a prior cached graph \\
commit & terminal state detached exactly once & atomic replace under episode ID and expected index \\
episode end & no state survives reset or terminal marker & evict all per-layer records for that episode \\
distributed lane & one episode has one chronological owner & sampler prevents cross-rank or cross-row migration \\
\bottomrule
\end{tabularx}
\end{table}

\subsection{Causality argument}

\paragraph{Proposition (no same-step or future-action leakage).}
Assume the runtime commits transition $t-1$ before predicting step $t$, and that current action-side tokens are readers only. Then the interaction-progress state queried at step $t$ is measurable with respect to completed transitions $0{:}t-1$. Its construction graph has no causal path from the action emitted at step $t$ or from any later transition.

\paragraph{Argument.}
Define the completed transition record $m_j=(o_{j+1},s_{j+1},a_j^{\mathrm{exec}})$ and let $\mathcal U$ denote its encoding followed by the execution-role TTT-Writer. It becomes available only after action $a_j^{\mathrm{exec}}$ has been executed and its resulting observation and state have arrived. With
\begin{equation}
\phi_{E,0}=\phi_0,\qquad
\phi_{E,t}=\mathcal U(\phi_{E,t-1},m_{t-1}),
\end{equation}
induction gives that $\phi_{E,t}$ depends only on $m_0,\ldots,m_{t-1}$. The policy reads $\phi_{E,t}$ before sending $a_t$, so neither $a_t^{\mathrm{exec}}$ nor later transitions can influence the prediction. The segmented training recurrence implements the same order explicitly rather than relying on a convention in tensor indexing.

\subsection{Persistent-state size}

One fast state stores two matrices and two biases per block and head. One demonstration-intent state and one interaction-progress state require
\begin{equation}
2LH(2d_hd_f+d_h+d_f)
=75{,}866{,}112
\end{equation}
FP32 scalars, approximately $289.4$ MiB per active stream excluding allocator overhead and temporary activations. This persistent allocation does not grow with elapsed episode length or reference length. Total runtime is not constant: reference preprocessing is linear in reference frames, and one ExeWriter update is performed for each completed control transition.

The added trainable components comprise the dual-memory blocks, reference semantic and motion resamplers, execution Q-former, semantic--motion fusion gate, and SA-AdaLN conditioner. The base action expert is also optimized by the outer action objective. The Qwen3-VL and V-JEPA2 backbones remain frozen.

\subsection{Memory and compute scaling}

Let $N_q=32$ be the Action-DiT query length and let one fast-MLP evaluation cost $\Theta(d_hd_f)$ per head and token. Let $C_{\mathrm{ref}}$ denote the cost of H-GEC sampling, semantic and temporal encoding, and register fusion for the complete reference video. Building its demonstration-intent state costs
\begin{equation}
C_{\mathrm{ref}}+\Theta(L\,H\,R_R\,d_h d_f)
\end{equation}
once per reference demonstration video. The preprocessing cost depends on the source video and sampled frame budget. The TTT term performs one write of the fused registers per Action-DiT block. One completed-transition ExeWriter commit costs $\Theta(L\,H\,R_E\,d_h d_f)$. A single Action-DiT evaluation reads one demonstration-intent state and one interaction-progress state per block, costing
\begin{equation}
\Theta\!\left(2L\,H\,N_qd_hd_f\right).
\end{equation}
Thus elapsed episode length does not enter the per-evaluation memory-read cost. With five Euler evaluations, an action-sampling call invokes $36\times5\times2=360$ logical dynamic-delta read pairs, each vectorized over tokens and heads. Every pair evaluates both the adapted and initial fast state, for 720 logical fast-MLP state evaluations. These are structural counts, not measured latency. Register count changes write cost and transient activation memory, but not persistent fast-state size. By contrast, full-history attention retains memory linear in the number of history tokens and attends over that growing store at every control step.

\section{Insertion into Action-DiT}
\label{app:dit}

In Action-DiT block $l$, let $H_{\mathrm{attn}}^l$ denote the
output of the base timestep-conditioned attention.
The reference state $\phi_R^l$ stores demonstration intent,
while $\phi_{E,t}^l$ stores the completed interaction history
available at control step $t$.
The shared TTT-Reader produces residuals
$\Delta O_R^l$ and $\Delta O_E^l$, whose contributions are
controlled by learned gates $g_R^l$ and $g_E^l$:
\begin{align}
\Delta O_R^l
&=\operatorname{TTT\!-\!Reader}
(H_{\mathrm{attn}}^l,\phi_R^l,r),
H_R^l
=H_{\mathrm{attn}}^l+\tanh(g_R^l)\Delta O_R^l,\\
\Delta O_E^l
&=\operatorname{TTT\!-\!Reader}
(H_R^l,\phi_{E,t}^l,e),
H_E^l
=H_R^l+\tanh(g_E^l)\Delta O_E^l.
\end{align}
The execution reader therefore queries $H_R^l$, which already
contains reference guidance, rather than independently querying
$H_{\mathrm{attn}}^l$.
This serial composition lets realized progress refine the
reference-conditioned representation without modifying the
reference fast weights.

The inherited AdaLN-SwiGLU residual maps $H_E^l$ to block $l+1$.
The inherited two-layer linear--GELU--linear flow action decoder
and five-step Euler integration remain unchanged.

The policy is optimized with the base action objective:
\begin{equation}
\mathcal L_{\mathrm{policy}}=\mathcal L_{\mathrm{base}}.
\end{equation}
The inner reconstruction loss constructs the differentiable fast update. It is not separately added to the outer scalar objective.

For completeness, the inherited base-policy term is also fixed by the implementation. Given clean action $a$, Gaussian noise $\epsilon$, and $u\sim\operatorname{Beta}(1.5,1.0)$, it samples $\tau=0.999(1-u)$. The Action-DiT receives $x_\tau=(1-\tau)\epsilon+\tau a$ and predicts velocity $v^\star=a-\epsilon$. Valid action coordinates have unit weight, so $\mathcal L_{\mathrm{mse}}$ is their masked mean squared velocity error. $\mathcal L_{\mathrm{freq}}$ is the masked mean magnitude of the temporal real-FFT difference between predicted and target velocities, excluding the two gripper coordinates, indexed by 7 and 15 in the zero-based action representation. Separately, each language-side action position predicts five action candidates. The candidate with minimum masked sequence-level $\ell_1$ error defines $\mathcal L_{\mathrm{choice}}$. A score head regresses all five detached candidate errors with mean squared error $\mathcal L_{\mathrm{score}}$. Therefore
\begin{equation}
\mathcal L_{\mathrm{base}}=
0.5\mathcal L_{\mathrm{mse}}+
\mathcal L_{\mathrm{freq}}+
0.5\mathcal L_{\mathrm{choice}}+
0.5\mathcal L_{\mathrm{score}}.
\end{equation}

\subsection{End-to-end tensor schedule and train--rollout parity}
\label{app:tensor-schedule}

Table~\ref{tab:tensor-schedule} records the policy's axes at interfaces where silent reshaping or state mutation would change the method. Here $B$ is the training-window batch, $S=8$ the supervised trajectory length, $K$ the Action-DiT diffusion-training repeat (actively $K=1$), $F_S$ and $F_M$ the sparse-semantic and ordered-clip frame counts for the complete reference input, $N_q=32$ the Action-DiT query length, and $N$ the padded post-merger patch count.

\begin{table}[H]
\centering
\footnotesize
\caption{Exact tensor and state schedule at the RoboDemo interfaces. ``Functional'' means the state is an ordinary differentiable tensor in the outer graph. ``Stored'' means it is detached runtime memory owned by one stream and layer.}
\label{tab:tensor-schedule}
\begin{tabularx}{\linewidth}{@{}>{\raggedright\arraybackslash}p{0.20\linewidth}>{\raggedright\arraybackslash}p{0.34\linewidth}>{\raggedright\arraybackslash}X@{}}
\toprule
Interface & Training & Layer-level rollout contract \\
\midrule
writer evidence & current Qwen $[B,S,1,N,2560]$ and reference Qwen/video $[B,F_S,\cdot]$ and $[B,F_M,\cdot]$ & reference evidence is encoded once on preload. Current ego frames are encoded at control boundaries \\
writer register tokens & $U^R:[B,32,1024]$ after semantic--motion fusion and $U^E:[B,S,16,1024]$ after SA-AdaLN & one $[B,16,1024]$ completed-transition input to ExeWriter per commit \\
Action-DiT readers & the base action expert flattens time-major $[S,B,K,N_q,D]$ to $[SBK,N_q,D]$. Each RoboDemo block restores $[B,S,K,N_q,D]$ & runtime fast states are repeated only across candidate/sample batch copies. The stored originals are not mutated by reads \\
fast states & the preloaded demonstration-intent state is functional. Interaction progress initializes from detached $\mathcal C^l[e]$ and is differentiable only within the current segment & one detached Retrieval-TTT state and one detached Execution-TTT state are owned independently by every block and stream \\
flow computation & one sampled $\tau$ and one velocity prediction per repeated training example & five Euler evaluations reuse the same Qwen cache and fast states. No write occurs inside the solver \\
causal advance & target $j$ reads after all preceding valid completed transitions & transition $t-1$ is committed before predicting $t$. The newly emitted $a_t$ cannot enter memory until the next step \\
\bottomrule
\end{tabularx}
\end{table}

%% file: tables/hyperparameters.tex
\begingroup
\footnotesize
\renewcommand{\arraystretch}{1.1}
\begin{longtable}{@{}>{\raggedright\arraybackslash}p{0.14\linewidth}>{\raggedright\arraybackslash}p{0.29\linewidth}>{\raggedright\arraybackslash}p{0.51\linewidth}@{}}
\caption{RoboDemo and Refreward configuration, grouped by model architecture, training, and benchmark/data settings.}\label{tab:all-hyperparameters}\\
\toprule
Component & Parameter & Value \\
\midrule
\endfirsthead
\multicolumn{3}{l}{Table~\thetable\ continued}\\
\toprule
Component & Parameter & Value\\
\midrule
\endhead
\bottomrule
\endfoot
\multicolumn{3}{l}{\textbf{(a) Model Architecture}}\\
\midrule
RoboDemo & Frozen semantic backbone & Qwen3-VL-4B-Instruct, vision/language blocks 24/36, post-merger width 2560 \\
RoboDemo & Frozen temporal backbone & V-JEPA2-AC ViT-G, output width 1408, tubelet size 2 \\
RoboDemo & Action-DiT depth $L$ & 36 \\
RoboDemo & Hidden width $D$ & 1024 \\
RoboDemo & Attention/TTT heads $H$ & 8, per-head width $d_h=128$ \\
RoboDemo & Fast MLP hidden width $d_f$ & 512, expansion factor 4 \\
RoboDemo & Reference/execution registers $R_R,R_E$ & 32 / 16, resampler heads 8 \\
RoboDemo & Persistent states & One reference + one execution state per layer \\
RoboDemo & Action horizon and reader length & $A=30$, $N_q=A+2=32$ \\
RoboDemo & Normalized input slots & State/action: 60 / 60 \\
RoboDemo & Action sampling & 5 Euler evaluations, 5 auxiliary action candidates \\
Refreward & Frozen encoders & Qwen3-VL-Embedding-2B (2048), DINOv3 ViT-H+ (1280) \\
Refreward & Input image sizes & Qwen: $448\times448$, DINO: $256\times256$, aspect-preserving letterbox \\
Refreward & Feature budgets & $L_t=32$ instruction tokens, $L_v=256$ spatial patches, $L_u=32$ process frames \\
Refreward & Color/position features & Lab: $3\times8=24$ bins, position: 4 coordinates per patch \\
Refreward & Input token lengths & Query 290, candidate 322 \\
Refreward & Hidden width and attention & $d=256$, 8 heads, FFN width 1024 \\
Refreward & Spatial/reward blocks & 2 / 4, reward latents 32, shared evidence K/V \\
Refreward & Relative-position MLP & $4\to32\to8$ \\
Refreward & Output head & $256\to256\to9$: 1 residual, 2 categorical logits, 6 factor logits \\
Refreward & Projection and embedding scales & Content norm $\sqrt d=16$, color/position/positional gain 0.25, role/modality gain 16 \\
Refreward & Modality/role embeddings & 5 modality + 2 role vectors \\
Refreward & Score prior and thresholds & Cosine center 0.9, 4 learned ordinal thresholds \\
Refreward & Normalization and bias & Bias-free linears, non-affine LayerNorm, dropout 0 \\
Refreward & Numerical constants & LayerNorm epsilon $10^{-5}$, L2 normalization floor $10^{-12}$, sinusoidal frequency base 10,000 \\
Refreward & Trainable parameters & 7,359,877 (frozen encoders excluded) \\
\midrule
\multicolumn{3}{l}{\textbf{(b) Training}}\\
\midrule
RoboDemo & Training seed & 42 \\
RoboDemo & Evaluated checkpoints & Full50: 240k, C15: 90k, C30: 160k, C45: 260k, C15-wo: 100k \\
RoboDemo & Optimizer & AdamW, betas $(0.9,0.95)$, epsilon $10^{-8}$, weight decay $10^{-4}$ \\
RoboDemo & Learning-rate schedule & Base $10^{-5}$, floor $10^{-6}$, 1,000-update warm-up, cosine decay \\
RoboDemo & Learning-rate multipliers & Pretrained policy 0.1, TTT/merge gates 1, resamplers/registers/SA-AdaLN 5 \\
RoboDemo & Gradient accumulation/clipping & 1 / 0.1 \\
RoboDemo & Training windows & $S=8$, stride 1, endpoint-aligned TBPTT, repeat $K=1$ \\
RoboDemo & Precision & BF16 forward, FP32 trainable parameters and fast states \\
RoboDemo & Inner update & $\eta_0=0.1$, relative-RMS clip 0.05, reference-scale floor $10^{-3}$, normalization epsilon $10^{-8}$ \\
RoboDemo & Initial residual gates & TTT gates 0.1 before tanh, motion residual 0.01, SA-AdaLN gates 0 \\
RoboDemo & Regularization & Metadata dropout 0.1, no data augmentation \\
RoboDemo & Action objective weights & MSE 0.5, frequency 1, choice 0.5, score 0.5 \\
RoboDemo & Flow time sampling & $u\sim\mathrm{Beta}(1.5,1.0)$, $\tau=0.999(1-u)$ \\
Refreward & Seed and batch organization & Seed 0, one query group per optimizer update, all 15,093 retained conditions \\
Refreward & Optimizer & AdamW, betas $(0.9,0.999)$, epsilon $10^{-8}$, weight decay $10^{-2}$, gradient clip 1 \\
Refreward & Initial schedule & 4,000 updates, peak $10^{-4}$, 200-update warm-up (5\%), cosine decay \\
Refreward & Final learning rate and selected step & $10^{-5}$, selected update 6,000 \\
Refreward & Loss coefficients & Ordinal 1, annotated pairwise 1, gates 0.5, factors 0.25, evidence 0.1, ordinal contrast 0.5 \\
Refreward & Ordinal contrast & Similarity temperature 0.1, denominator floor $10^{-12}$ \\
Refreward & Supervision encoding & Ordinal/factors: 1--5, pairwise: 1/0/0.5, factor BCE target $(y_f-1)/4$ \\
Refreward & Initial score scale & Positive prior scale 20, final head weights zero \\
Refreward & Latent/role initialization & Normal standard deviation 0.02, threshold parameters $(-1.5,0,0,0)$ \\
\midrule
\multicolumn{3}{l}{\textbf{(c) Benchmark And Data}}\\
\midrule
Imitator Game & Full-50 training corpus & 50 tasks, 4 scene levels, 200 episodes/task, 10,000 episodes total \\
Imitator Game & Evaluation budget & 10 trials/task-level, 2,000 total \\
Imitator Game & Observation and raw dimensions & Single RGB, state 18, action 16 \\
Imitator Game & Reference input & H-GEC over the complete reference video, $F_S$ sparse-semantic frames and $F_M$ ordered clip frames \\
Imitator Game & Temporal preprocessing & $256\times256$ center crop, short-side resize factor $256/224$ \\
Imitator Game & Simulator and reference rates & Control: 20 Hz, physics: 100 Hz, reference: 30 fps, horizon: 500 steps \\
Imitator Game & Action ensembling & Store every 4 control steps, retain up to 4 chunks, weight $\exp(-0.1\,\mathrm{age})$ \\
Imitator Game & Gripper coordinates & Zero-based action indices 7 and 15 excluded from frequency loss \\
Retrieval corpus & Fixed queries & 350, 7 per task, stratified across 4 levels \\
Retrieval corpus & Retained labels & Pointwise 12,994, pairwise 2,099, total 15,093 \\
Retrieval corpus & Teacher judgments & 3 ordinary judgments/condition, total 45,279 \\
Retrieval corpus & Candidate-bank size & Training query: 199 other same-task episodes, evaluation query: 200 \\
Retrieval corpus & Teacher evidence & Query: instruction + initial image, candidate: instruction + 32 timestamped frames \\
Retrieval corpus & Text input cap & 512 instruction tokens before order-preserving pooling to 32 \\
H-GEC & Event preprocessing & Optical flow: $96\times96$ grayscale, elapsed-time floor $10^{-6}$ s \\
H-GEC & Event scoring & 10th/90th percentile normalization, flow/state/gripper weights 0.40/0.35/0.25 when available \\
H-GEC & Event spacing & 0.4 s initial separation, relaxed if candidates are insufficient \\
\end{longtable}
\endgroup

%% file: segments/appendix_ttt_algorithms.tex
\begin{appendixalgorithm}
\footnotesize
\caption{Shared TTT write and dynamic-delta read in one Action-DiT block}
\label{alg:ttt-operations}
\begin{algorithmic}[1]
\Require Slow parameters and initialization $\phi_0$, role $\star\in\{r,e\}$
\Function{Write}{$\phi,U,\star,m$}
  \If{$\sum_i m_i=0$}
    \State \Return $\phi$
  \EndIf
  \State $K_i\gets\operatorname{norm}(U_iW_K)$
  \State $V_i\gets U_iW_V$
  \State $Y_i\gets\operatorname{LN}_{\rm in}(V_i-K_i)$
  \State $\widehat Y_i\gets\operatorname{LN}_{\rm in}(F_\phi(K_i))$
  \State $\eta_i^\star\gets
    \eta_0\sigma(g_\eta(U_i)+\delta_\eta^\star)m_i/
    [d_h\max(1,\sum_jm_j)]$
  \State $\Delta\phi\gets-\sum_i\eta_i^\star
    \nabla_\phi\frac12\|\widehat Y_i-Y_i\|_2^2$
    \Comment{Analytic, differentiable update}
  \State $c\gets\min\!\left(1,
    \frac{0.05\max(\operatorname{RMS}(\phi_0),10^{-3})}
    {\operatorname{RMS}(\Delta\phi)+10^{-8}}\right)$
    \Comment{Jointly over each head's fast tensors}
  \State \Return $\phi+\operatorname{StopGrad}(c)\Delta\phi$
    \Comment{FP32 state and accumulation}
\EndFunction
\Function{ReadDelta}{$H,\phi,\star$}
  \State $q^\star\gets\operatorname{norm}(W_Q(H+e_{\rm read}^{\star}))$
  \State $A\gets W_O\operatorname{LN}_{\rm post}
    [q^\star+\operatorname{LN}_{\rm in}(F_\phi(q^\star))]$
  \State $B\gets W_O\operatorname{LN}_{\rm post}
    [q^\star+\operatorname{LN}_{\rm in}(F_{\phi_0}(q^\star))]$
  \State \Return $A-B$ \Comment{Identical queries and unchanged stored state}
\EndFunction
\end{algorithmic}
\end{appendixalgorithm}

\begin{appendixalgorithm}
\footnotesize
\caption{Episode-local training and rollout with dual-cascaded TTT}
\label{alg:ttt-lifecycle}
\begin{algorithmic}[1]
\Require Complete reference $\mathcal D$, block initializations $\{\phi_0^l\}_{l=1}^L$
\Function{PreloadReference}{$\mathcal D$}
  \State $(U^R,m^R)\gets\operatorname{ReferenceEncoder}(\mathcal D)$
  \For{$l=1,\ldots,L$}
    \State $\phi_R^l\gets\Call{Write}{\operatorname{Copy}(\phi_0^l),U^R,r,m^R}$
  \EndFor
  \State \Return $\{\phi_R^l\}_{l=1}^L$
\EndFunction
\Function{Cascade}{$H_{\rm attn}^l,\phi_R^l,\phi_E^l$}
  \State $H_R^l\gets H_{\rm attn}^l+
    \tanh(g_R^l)\Call{ReadDelta}{H_{\rm attn}^l,\phi_R^l,r}$
  \State $H_E^l\gets H_R^l+
    \tanh(g_E^l)\Call{ReadDelta}{H_R^l,\phi_E^l,e}$
  \State \Return $\operatorname{BaseFFN}_l(H_E^l)$
\EndFunction
\Statex \textbf{Training: one chronological window of episode $e$}
\State $\{\phi_R^l\}\gets\Call{PreloadReference}{\mathcal D_e}$
\State At episode start, initialize separate copies of $\phi_0^l$.
  Otherwise, restore detached execution states and last committed transition from $\mathcal C[e]$
\For{each valid step $t$ in the ordered eight-step window}
  \If{transition $t-1$ is completed and has not yet been committed}
    \State $(U_t^E,m_t^E)\gets\operatorname{ExecutionEncoder}
      (o_t,s_t,a_{t-1}^{\rm exec})$
    \State $\phi_E^l\gets\Call{Write}{\phi_E^l,U_t^E,e,m_t^E}$ for all $l$
    \State Advance the committed-transition index
  \EndIf
  \State Predict the training flow velocity using \Call{Cascade}{} in every block
  \State Accumulate the base action loss for this step
\EndFor
\State Store $\operatorname{StopGrad}(\{\phi_E^l\})$ and the transition index in $\mathcal C[e]$
\State Differentiate the action loss through the current window
\State Update slow parameters
\Statex \textbf{Rollout: one episode}
\State $\{\phi_R^l\}\gets\Call{PreloadReference}{\mathcal D}$
\State $\phi_E^l\gets\operatorname{Copy}(\phi_0^l)$ for all $l$
\For{$t=0,\ldots,T-1$}
  \If{$t>0$}
    \State $(U_t^E,m_t^E)\gets\operatorname{ExecutionEncoder}
      (o_t,s_t,a_{t-1}^{\rm exec})$
    \State $\phi_E^l\gets\Call{Write}{\phi_E^l,U_t^E,e,m_t^E}$ for all $l$
  \EndIf
  \State Encode the current image and instruction once into the VLM cache
  \State Generate a 30-action chunk with five Euler evaluations in which
    each block uses \Call{Cascade}{} with fixed $\phi_R^l,\phi_E^l$
  \State Ensemble and bound the action
  \State Execute $a_t^{\rm exec}$ and receive $(o_{t+1},s_{t+1})$
\EndFor
\State Release the episode's reference and execution states
\end{algorithmic}
\end{appendixalgorithm}

%% file: segments/appendix_experiment_reporting.tex
\section{Experimental Details}
\label{app:experiment_reporting}

\subsection{Benchmark, task membership, and data budgets}
\label{app:benchmark}
Imitator Game evaluates whether a robot can reproduce the intent of a demonstration under changes to the execution scene~\citep{zhou2026imitatorgamebenchmarkingrobot}. We use its 50-task IG-10K suite, containing 10,000 robot demonstrations with paired human references. L0 preserves the reference scene, while L1--L3 change object positions, appearance attributes, and functional affordances, respectively. Each simulation task is evaluated at all four levels.

C15, C30, and C45 are nested training corpora with 15, 30, and 45 tasks. Their common Seen subset comprises FoldTowel, PlaceFileFolder, PlaceMugRack, PlacePlateRack, and StirSpoon. The five policy-held-out tasks are FoldBox, PickFood, PickRemoteControl, PourKettle, and ScanMilkBox. Full50 includes all 50 tasks in training and evaluation. C15-wo uses the C15 robot corpus and retrieved robot references. Table~\ref{tab:experiment-budgets} lists the budgets. Table~\ref{tab:task-membership} gives the complete task membership.

\input{tables/experiment_budgets}

\paragraph{Success rate and aggregation.}
For an evaluation set $\mathcal E$, let $y_i$ indicate whether episode $i$ satisfies the task success condition within its 500-control-step horizon. We report
\begin{equation}
\operatorname{SR}(\mathcal E)=100\,\frac{\sum_{i\in\mathcal E}y_i}{|\mathcal E|}.
\label{eq:success-rate}
\end{equation}
Each task--level cell contains ten trials, giving 200 episodes per five-task split and 2,000 for Full50. Task--level cells receive equal weight. The Seen cross-scale average weights C15, C30, and C45 equally. Full50 category scores average their member tasks, and the overall score weights all 50 tasks equally.

\paragraph{Reference identities and random seeds.}
Human--robot training assignments are sampled with seed 42 and held fixed. Human-reference evaluation selects one reference per task using a frozen SHA256-seeded manifest and reuses it across levels and trials. For the zero-based alphabetical task index $j$ and trial index $i\in\{0,\ldots,9\}$, the evaluation seeds are
\begin{equation}
\begin{aligned}
s_{\rm reset}&=20261007+100j+i,\\
s_{\rm policy}&=202610070+10000j+500i.
\end{aligned}
\label{eq:eval-seeds}
\end{equation}
The human references and seeds are shared across C15/C30/C45 and their corresponding Full50 conditions. Each setting uses training seed 42. Human references for policy-held-out tasks are supplied at evaluation. RoboDemo-wo instead selects a task-matched robot reference from the IG training bank, as detailed in Appendix~\ref{app:reference-source}.

\end{multicols}
\input{tables/imitationgame_task_membership}
\begin{multicols}{2}
\raggedcolumns

\subsection{Observation, action, and reference processing}
\label{app:data}
The policy receives one RGB view, an 18-dimensional proprioceptive state, and 16-dimensional action targets. State and action values are normalized using the corresponding training corpus and padded to the policy's 60-dimensional interfaces. Unused action coordinates are masked. The frozen semantic and motion encoders produce 32 fused reference registers. Execution-TTT receives 16 registers per completed transition. Training windows contain eight consecutive steps with unit temporal stride and endpoint-aligned state/action conditioning. Detached episode-indexed storage carries execution states between windows.

Human-reference RoboDemo receives its reference directly. RoboDemo-wo retrieves a robot reference with the frozen scorer in Appendix~\ref{app:refreward}. Both configurations use the same policy reference encoder and dual-memory action expert. The scale, Full50, and memory-ablation comparisons use human references. Table~\ref{tab:reference-source} compares human and retrieved robot references.

\subsection{Optimization and evaluated checkpoints}
\label{app:optimization}
Table~\ref{tab:all-hyperparameters}(b) lists the shared optimization recipe. Each policy is trained on its own task corpus from the pretrained base policy. AdamW uses betas $(0.9,0.95)$, epsilon $10^{-8}$, and weight decay $10^{-4}$. The base learning rate is $10^{-5}$ with a $10^{-6}$ floor, 1,000 warm-up updates, and cosine decay. Learning-rate multipliers are 0.1 for pretrained policy parameters, 1 for TTT slow parameters and merge gates, and 5 for resamplers, register queries, and SA-AdaLN. Qwen3-VL and V-JEPA2 remain frozen. Forward computation uses BF16. Trainable parameters and fast states use FP32.

We evaluate Full50 at 240,000 updates, C15 at 90,000, C30 at 160,000, C45 at 260,000, and C15-wo at 100,000. The checkpoint associated with each result is explicit in the tables.

The outer objective is the base action loss in Appendix~\ref{app:dit}. The inner reconstruction objective defines differentiable fast-state writes. For each ordered window, Retrieval-TTT encodes the complete reference and Execution-TTT restores the preceding episode state. Each completed transition is committed once before subsequent action prediction. The terminal execution state is detached and retained for the next window. Algorithm~\ref{alg:ttt-lifecycle} gives the schedule. Refreward is trained separately with the objective in Appendix~\ref{app:rr_training}.

\subsection{Evaluation deployment}
\label{app:deployment}
Evaluation uses joint-position control, PhysX CPU simulation, OptiX rendering, and episode-level scene reconfiguration. The policy is queried at every control step. Five Euler evaluations generate a candidate action chunk of length 30 with 16 active coordinates. Every fourth step, a new chunk is retained. Up to four retained chunks are aligned to the current step and combined with weights proportional to $\exp(-0.1\,\mathrm{age})$, newer first. The bounded action applied to the robot is fed back to execution memory. A complete 500-step rollout has 500 policy calls and 499 completed-transition writes. Both memories remain fixed inside the Euler solver.

Simulation uses a 20-Hz control rate and a 100-Hz physics rate. Reference videos use a 30-fps time base. Measured policy latency is reported separately in Appendix~\ref{app:timing}. Table~\ref{tab:checkpoint-provenance} lists episode counts for all evaluated checkpoints.

\input{tables/experiment_counts}

\subsection{Complete Seen and Unseen task results}
\label{app:selected-results}
Table~\ref{tab:selected-details} reports all 120 task--level cells for C15 at 90k, C30 at 160k, and C45 at 260k. Their Seen counts are 179/200, 188/200, and 177/200, giving a three-scale average of 90.7\%. The Unseen counts are 0/200, 1/200, and 23/200. Table~\ref{tab:selected-unseen-c45} gives the C45 Unseen profile used in the main comparison.

\input{tables/imitationgame_selected_details}

\paragraph{Seen-task profiles.}
PlacePlateRack achieves 100\%, 100\%, and 97.5\% across C15/C30/C45. PlaceMugRack reaches 92.5\%, 95.0\%, and 92.5\%, while StirSpoon reaches 87.5\%, 97.5\%, and 90.0\%. PlaceFileFolder scores 75.0\%, 85.0\%, and 67.5\%. FoldTowel scores 92.5\%, 92.5\%, and 95.0\%. The C30 level averages are 90.0/92.0/96.0/98.0\%. The C45 averages are 92.0/80.0/92.0/90.0\%.

\paragraph{Held-out-task profiles.}
C30 completes one FoldBox trial at L2. C45 completes 19 FoldBox trials, two PourKettle trials at L0 and L1, and two ScanMilkBox trials at L0 and L2. Its FoldBox profile is 90.0/0.0/0.0/100\% across L0--L3. PickFood and PickRemoteControl score 0.0\%. These profiles identify the procedures and scene conditions in which transfer occurs.

\subsection{Full50 evaluation and aggregation}
\label{app:protocol}
Full50 contains 50 robot demonstrations for each of 50 tasks and four levels, giving 10,000 training demonstrations. The 240k checkpoint succeeds in 1,692 of 2,000 episodes, achieving 84.6\% SR with human references and the deployment protocol above.

\subsection{Complete Full50 task-level results}
\label{app:full50-results}
Table~\ref{tab:full50-all} includes all 50 tasks in five disjoint reporting groups: transport and placement (28), pouring (6), cleaning (4), opening and folding (5), and tool use and scanning (7). For group $\mathcal G$, the level-specific mean is $|\mathcal G|^{-1}\sum_{t\in\mathcal G}s_{m,t,l}$, where $s_{m,t,l}$ is method $m$'s SR on task $t$ at level $l$. Avg.\ averages the four levels.

The main Full50 table subdivides transport and placement into container transfer (13), rack/shelf storage (5), and tabletop placement (10). Membership is shared across methods. DP and ACT (SigLIP2) follow the official full-suite release~\citep{zhou2026imitatorgamebenchmarkingrobot}.

\input{tables/imitationgame_full50_all}

\paragraph{Task and level analysis.}
RoboDemo reaches 100\% on FoldBox, FoldTowel, PlaceBrushRest, PlaceChipsRack, PlacePlateRack, PourLiquidMug, and TransFood. Its pouring average is 82.9\%, compared with ACT's 66.3\%, and tool use/scanning reaches 85.0\%, compared with 71.8\%. CleanDesk and PlaceFruitBox each score 37.5\%, while ACT scores 95.0\% and 62.5\%. RoboDemo's level averages are 92.6/88.2/75.8/81.8\%. Full50 includes FoldBox in training. The nested-corpus evaluations hold this task out.

\subsection{Baseline sources and complete comparisons}
\label{app:baseline-details}
\hypertarget{exp-source-benchmark}{}The benchmark follows Imitator Game~\citep{zhou2026imitatorgamebenchmarkingrobot}. \hypertarget{exp-source-scales}{[E2]} identifies the official simulation task-detail release accessed on November 8, 2026: \url{https://imitator-game.github.io/leaderboard.html}. Tables~\ref{tab:baseline-aggregates}--\ref{tab:baseline-zs-45} retain all 16 released variants: four language-conditioned policies, two video-skill policies, ACT/DP/VQ-BeT with DINOv2, SigLIP2, or VideoMAE encoders, and Zero-WAM.

The original 15 baseline recipes use 50 demonstrations per task--level pair and frozen visual/task encoders. Zero-WAM uses scale-specific SR values from the leaderboard's ``By task'' view for the Seen and Zero-shot regimes at L0--L3. Full50 uses the official DP and ACT full-suite results. All task--level values are preserved, and aggregates follow the averaging rules above.

\input{tables/imitationgame_baseline_aggregates}

RoboDemo leads the Seen comparison at all three scales. ACT/DINOv2 has the highest baseline three-scale average, 80.8\%, and $\pi_{0.5}$ is the strongest C45 baseline at 82.0\%. Zero-WAM scores 68.0/77.0/68.0\% on Seen tasks. On C45 Unseen tasks, Zero-WAM, VQ-BeT/DINOv2, RoboDemo, and VQ-BeT/SigLIP2 score 22.0\%, 17.5\%, 11.5\%, and 10.5\%, respectively. Zero-WAM reaches 80.0\% on FoldBox and 22.5\% on PourKettle. RoboDemo scores 47.5\% and 5.0\%, and additionally completes two ScanMilkBox trials. Tables~\ref{tab:baseline-seen-15}--\ref{tab:baseline-zs-45} give all 1,280 official baseline cells alongside RoboDemo.

\input{tables/imitationgame_baseline_details}

\subsection{Context and memory ablations on C15-Seen}
\label{app:ablation-protocol}
We compare the six configurations in Table~\ref{tab:ablation-components}, derived from the same RoboDemo architecture. The base VLA removes the RoboDemo-specific additions while retaining the vision-language backbone and action expert. Each configuration uses C15 training and the common 200-episode Seen evaluation.

\paragraph{Reference access and execution context.}
Rows 2 and 3 introduce the human reference through alternative interfaces. Row 2 supplies the concatenated tokens $[Z_{\rm ref}\Vert Z_{\rm current}]$ to the VLM. Row 3 writes the reference into Retrieval-TTT for the action expert to read. Rows 4--6 also consume actual execution-video context up to the current decision.

\paragraph{One state versus two states.}
Row 4 writes reference and execution tokens into one numerical fast state. Rows 5 and 6 allocate independent states: the reference state is preloaded and fixed during rollout, while execution memory accumulates completed transitions. These rows compare context organization and numerical state allocation under shared TTT projections and initialization.

\paragraph{Parallel versus cascaded reads.}
The parallel configuration queries both memories from the original attention output:
\begin{equation}
\begin{aligned}
H_{\rm par}^{l}={}&H_{\rm attn}^{l}\\
&+\tanh(g_R^l)\operatorname{TTT\!-\!Reader}(H_{\rm attn}^{l},\phi_R^{l},r)\\
&+\tanh(g_E^l)\operatorname{TTT\!-\!Reader}(H_{\rm attn}^{l},\phi_{E,t}^{l},e).
\end{aligned}
\end{equation}
RoboDemo uses Eq.~\eqref{eq:method-cascade}: the execution reader queries the reference-conditioned $H_R^l$. Execution context refines the reference read while the reference fast state remains fixed.

\paragraph{Effect sizes.}
Reference-prefix and reference-TTT conditioning improve the 62.0\% base VLA by 6.5 and 9.5 percentage points. Adding execution context to the shared state gives another 1.5 points. Independent parallel states gain 9.0 points over the shared state. Cascading adds 7.5 points and reaches 89.5\%. The full architecture gains 27.5 points over the base VLA.

\paragraph{Refreward development evaluation.}
The selector's fixed development set contains 50 queries with 60 candidates each and is used for checkpoint selection. Its top-ranked candidate is teacher-labeled suitable for 45 of the 50 queries. All 50 top-ranked candidates have known suitability labels. This evaluates reference suitability. The reference-source comparison below reports robot control SR.

\subsection{Component sizes and measured inference costs}
\label{app:timing}
We profile the C45 policy and frozen Refreward scorer on NVIDIA H200 GPUs at batch size one. Policy forwards use BF16 autocast with FP32 trainable parameters and fast states. Refreward uses FP32. Timings start after model loading and offline candidate preprocessing.

\paragraph{Policy components.}
The policy contains 6.347B unique parameters: 4.438B in the VLM, 0.675B in the base action expert, 0.190B in Dual-TTT, 0.032B in memory adapters, and 1.012B in the reference motion encoder. Dual-TTT has 189.7M slow parameters, including 37.9M learned initialization parameters. Reference and execution fast states each contain 37.9M entries across 36 layers, totaling 75.9M runtime entries per episode. Adapter counts cover both reference and execution paths. Online adapter timing measures execution-token construction.

\paragraph{Policy-query timing.}
We replay recorded $224\times224$ RGB observations, proprioception, and preceding executed actions from FoldBox, PlaceMugRack, and StirSpoon. Each sequence receives ten warm-up calls and 30 measured calls, giving 90 measurements. CUDA events measure component intervals, including host-launch gaps. The VLM interval includes writer-side vision and current image--text preparation. One completed-transition write across 36 layers takes 55.1\,ms. Cascaded memory reads total 295.1\,ms over five Euler rounds, or 59.0\,ms per round. The base action expert totals 144.1\,ms, or 28.8\,ms per round.

Separate synchronized wall-time measurements cover preprocessing, session checks, memory updates, and action sampling, ending before action ensembling, native conversion, rendering, and simulation. Mean query latency is 563.6\,ms, median 559.9\,ms, and p95 588.1\,ms. Instrumented calls average 578.7\,ms. Table~\ref{tab:timing-protocol} gives the distributions.

\paragraph{Single-pair Refreward timing.}
The scorer has 7.4M parameters and takes 5.3\,ms per query--candidate pair. We use ten distinct-instruction query pairs, each with a same-instruction candidate from a different episode group. Each pair receives 50 warm-up and 200 measured calls, giving 2,000 synchronized wall-time measurements. Every call recomputes the learned query/candidate transformations and scoring forward from GPU-resident frozen features. The interval covers the scorer. Frozen encoder extraction, input transfer, candidate recall, and sorting occur outside it.

\input{tables/experiment_timing_protocol}

\subsection{C15 human-reference and robot-reference comparison}
\label{app:reference-source}
\paragraph{Training-reference assignment.}
RoboDemo and RoboDemo-wo share the architecture and action-learning objective. RoboDemo uses human references. RoboDemo-wo uses Refreward-selected robot demonstrations. For training episode $q$, selection over its task-compatible C15 bank is
\begin{equation}
i^\star(q)=\arg\max_{\substack{i\in\mathcal B_{\mathrm{C15}}(q)\\i\ne q}}
\operatorname{Refreward}_{\psi}\!\left(\xi(\ell_q,o_q^0),\xi_i\right).
\end{equation}
Episode-level exclusion gives 199 candidates per training query. Scoring uses the instruction and initial observation. The highest-scoring distinct episode supplies the video. This reference enters the semantic--motion encoder and Retrieval-TTT, while Execution-TTT tracks actual execution. Policy training retains robot-action supervision.

\paragraph{Evaluation references.}
At evaluation, Refreward selects from 200 robot training demonstrations of the query task, and the reference stays fixed during rollout. Seen-task candidates belong to C15. For policy-held-out tasks, a task-matched robot reference is provided from the IG training bank. Table~\ref{tab:additional-checkpoints} reports all 400 C15-wo episodes at 100k.

\paragraph{Results and task differences.}
The five labels in Table~\ref{tab:reference-source} correspond to FoldTowel, PlaceFileFolder, PlaceMugRack, PlacePlateRack, and StirSpoon. Human-reference and robot-reference policies score 89.5\% and 90.0\%, exceeding the dedicated comparison's ACT score of 82.5\% by 7.0 and 7.5 percentage points. Robot-reference task scores are 92.5/90.0/95.0/97.5/75.0\%, giving differences of 0.0/+15.0/+2.5/$-2.5$/$-12.5$ points relative to human references. The overall difference is +0.5 points. Retrieved cross-episode robot references provide effective procedural guidance within the same dual-memory policy.

The dedicated comparison uses task-level baseline aggregates of 62.9\% for $\pi_{0.5}$, 82.5\% for ACT, and 68.6\% for Zero-WAM. These aggregates belong to Table~\ref{tab:reference-source}. The official-release comparisons use Tables~\ref{tab:scales} and~\ref{tab:baseline-aggregates}.

\subsection{Complete robot-reference results}
\label{app:additional-checkpoints}
Table~\ref{tab:additional-checkpoints} gives the complete C15-wo 100k evaluation, with 200 Seen and 200 policy-held-out episodes. C15-wo scores 90.0\% Seen and 0.0\% on policy-held-out tasks, with 180 and zero successes.

\subsection{Real-world evaluation}
\label{app:real-world}
Each standard task and each Tabletop Organization generalization condition uses 50 trials per method. The standard-task mean equally weights Tabletop Organization, Sofa Cushion Arrangement, and Bottle-to-Cup Pouring. Visual generalization changes lighting or the tabletop background while preserving the task objective. Policies trained in the standard scene are evaluated directly with learned parameters frozen. RoboDemo retains its episode-local fast-state operations.

Performance retention is $100\,\operatorname{SR}_{\mathrm{shift}}/\operatorname{SR}_{\mathrm{standard}}$, using each method's own Tabletop Organization score as reference. RoboDemo retains 51.1\% under changed lighting and 84.4\% under a changed background.

\paragraph{Figure panels and radial scales.}
Figure~\ref{fig:real-world-evaluation}(a) combines real-world standard-task scores with Full50 simulation profiles regrouped into nine operation families. In the relative-score radar, 100 denotes the strongest displayed method within a family. Scores 50--100 map to 12--100\% of the radius, with lower values placed inward. In the raw-SR radar, intervals 0--50, 50--90, and 90--100\% map to 0--15, 15--85, and 85--100\% of the radius. Numeric axes define these piecewise scales. Panel (b) shows task photographs and the Standard, Light, and Background tabletop conditions.

%% file: tables/experiment_budgets.tex
\begingroup
\input{tables/experiment_style}
\begin{table}[H]
\centering
\captionsetup{width=\linewidth}
\scriptsize
\setlength{\tabcolsep}{1.5pt}
\renewcommand{\arraystretch}{1.05}
\begin{tabularx}{\linewidth}{@{}lrrrrl>{\raggedright\arraybackslash}X@{}}
\toprule
Setting & Tasks & \shortstack{Robot\\demos} & Updates & \shortstack{Trials/\\cell} & Ref. & \shortstack[l]{Evaluation\\episodes}\\
\midrule
C15 & 15 & 3,000 & 90k & 10 & Human & 200 Seen + 200 Unseen\\
C30 & 30 & 6,000 & 160k & 10 & Human & 200 Seen + 200 Unseen\\
C45 & 45 & 9,000 & 260k & 10 & Human & 200 Seen + 200 Unseen\\
Full50 & 50 & 10,000 & 240k & 10 & Human & 2,000 across all tasks\\
C15-wo & 15 & 3,000 & 100k & 10 & Robot & 200 Seen + 200 held-out\\
\bottomrule
\end{tabularx}
\caption{\textbf{Training and evaluation budgets.} Each training task contributes 50 robot demonstrations per scene level. The nested human-reference pools contain 937, 1,831, and 2,765 videos for C15, C30, and C45. C15-wo trains on robot demonstrations and uses retrieved robot references. The evaluated checkpoints are shared by the main-text and appendix comparisons.}
\label{tab:experiment-budgets}
\end{table}
\endgroup

%% file: tables/imitationgame_task_membership.tex
\begingroup
\input{tables/experiment_style}
\begin{table}[H]
\centering
\fontsize{7pt}{8pt}\selectfont
\tabcolsep=6pt
\renewcommand\arraystretch{1}
\begin{tabularx}{\linewidth}{@{}r>{\raggedright\arraybackslash}p{0.32\linewidth}|*{4}{>{\centering\arraybackslash}X}>{\raggedright\arraybackslash}p{0.18\linewidth}@{}}
\toprule
$j$ & Task & C15 & C30 & C45 & \shortstack{Test\\split} & \shortstack{Human\\reference}\\
\midrule
\multicolumn{7}{l}{\textit{Transport and placement}}\\
\midrule
6 & KnifeBowlFork & -- & Y & Y & -- & human\_H47:9\\
10 & PickAppleBananaToBaskets & -- & -- & Y & -- & human\_H8:0\\
11 & PickAppleBasket & -- & -- & Y & -- & human\_H7:9\\
12 & PickAppleToScale & -- & -- & Y & -- & human\_H9:3\\
13 & PickFood & -- & -- & -- & unseen & human\_H21:0\\
14 & PickFruitsToPlate & -- & -- & Y & -- & human\_H16:5\\
15 & PickPillToRegions & -- & Y & Y & -- & human\_H31:8\\
16 & PickRemoteControl & -- & -- & -- & unseen & human\_H4:4\\
17 & PickTennisBallGolfBall & -- & Y & Y & -- & human\_H26:9\\
19 & PlaceBookBookcase & -- & -- & Y & -- & human\_H2:9\\
20 & PlaceBrushRest & Y & Y & Y & -- & human\_H35:9\\
21 & PlaceBurgerTray & -- & -- & Y & -- & human\_H56:1\\
22 & PlaceChipsRack & Y & Y & Y & -- & human\_H10:5\\
23 & PlaceClothBasket & Y & Y & Y & -- & human\_H3:7\\
24 & PlaceCommodityRack & -- & Y & Y & -- & human\_H11:1\\
25 & PlaceCupPlate & -- & Y & Y & -- & human\_H36:0\\
26 & PlaceFileFolder & Y & Y & Y & seen & human\_H32:4\\
27 & PlaceFoodScale & -- & -- & Y & -- & human\_H23:7\\
28 & PlaceFruitBox & Y & Y & Y & -- & human\_H12:1\\
29 & PlaceMagazineFolder & Y & Y & Y & -- & human\_H5:9\\
30 & PlaceMugRack & Y & Y & Y & seen & human\_H57:2\\
31 & PlacePillBox & -- & -- & Y & -- & human\_H34:1\\
32 & PlacePlateRack & Y & Y & Y & seen & human\_H14:3\\
33 & PlaceScrewdriver & -- & Y & Y & -- & human\_H58:7\\
34 & PlaceShoeBox & -- & -- & Y & -- & human\_H28:0\\
43 & PutBox & -- & Y & Y & -- & human\_H46:4\\
44 & PutCubeOnScale & -- & Y & Y & -- & human\_H40:1\\
48 & TransFood & -- & Y & Y & -- & human\_H22:4\\
\midrule
\multicolumn{7}{l}{\textit{Pouring}}\\
\midrule
35 & PourCup & -- & Y & Y & -- & human\_H41:0\\
36 & PourKetchupFries & -- & Y & Y & -- & human\_H17:1\\
37 & PourKettle & -- & -- & -- & unseen & human\_H20:5\\
38 & PourLiquidCup & -- & Y & Y & -- & human\_H33:7\\
39 & PourLiquidFilter & -- & -- & Y & -- & human\_H45:7\\
40 & PourLiquidMug & -- & -- & Y & -- & human\_H39:2\\
\midrule
\multicolumn{7}{l}{\textit{Cleaning}}\\
\midrule
0 & CleanCup & Y & Y & Y & -- & human\_H42:5\\
1 & CleanDesk & -- & Y & Y & -- & human\_H19:2\\
18 & PickWash & Y & Y & Y & -- & human\_H6:1\\
49 & WipePot & -- & Y & Y & -- & human\_H18:1\\
\midrule
\multicolumn{7}{l}{\textit{Opening and folding}}\\
\midrule
3 & FoldBox & -- & -- & -- & unseen & human\_H30:8\\
4 & FoldTowel & Y & Y & Y & seen & human\_H49:5\\
7 & LiftLidFromSkillet & Y & Y & Y & -- & human\_H48:9\\
8 & OpenBox & -- & -- & Y & -- & human\_H29:9\\
9 & OpenLiquidCap & -- & -- & Y & -- & human\_H43:1\\
\midrule
\multicolumn{7}{l}{\textit{Tool use and scanning}}\\
\midrule
2 & CutFruit & Y & Y & Y & -- & human\_H15:0\\
5 & GrindFood & Y & Y & Y & -- & human\_H44:9\\
41 & PressJuicer & -- & Y & Y & -- & human\_H50:4\\
42 & PressStapler & -- & -- & Y & -- & human\_H24:8\\
45 & ScanMilkBox & -- & -- & -- & unseen & human\_H13:1\\
46 & ScanPillBottle & -- & -- & Y & -- & human\_H27:7\\
47 & StirSpoon & Y & Y & Y & seen & human\_H1:5\\
\bottomrule
\end{tabularx}
\caption{\textbf{Task membership and evaluation references.} All 50 tasks belong to full50. Y denotes membership in each nested training corpus. Seen/Unseen identify the common evaluation split. Index $j$ specifies the evaluation seeds in Eq.~\eqref{eq:eval-seeds}.}
\label{tab:task-membership}
\end{table}
\endgroup

%% file: tables/experiment_counts.tex
\begingroup
\input{tables/experiment_style}
\begin{table}[H]
\centering
\footnotesize
\setlength{\tabcolsep}{3pt}
\renewcommand{\arraystretch}{1.05}
\begin{tabular}{@{}lrr@{}}
\toprule
Checkpoint & Seen & Unseen\\
\midrule
C15 90k & 179/200 & 0/200\\
C30 160k & 188/200 & 1/200\\
C45 260k & 177/200 & 23/200\\
C15-wo 100k & 180/200 & 0/200\\
Full50 240k & 1,692/2,000 & --\\
\bottomrule
\end{tabular}
\caption{\textbf{Successful episodes / evaluated episodes.} C15-wo Unseen denotes policy-held-out tasks supplied with robot references. Full50 evaluates all 50 training tasks.}
\label{tab:checkpoint-provenance}
\end{table}
\endgroup

%% file: tables/imitationgame_selected_details.tex
\end{multicols}
\begingroup
\input{tables/experiment_style}
\begin{table}[H]
\centering
\scriptsize
\setlength{\tabcolsep}{8pt}
\renewcommand{\arraystretch}{1}
\resizebox{\linewidth}{!}{%
\begin{tabular}{@{}l|rrrrr|rrrrr|rrrrr@{}}
\toprule
 & \multicolumn{5}{c}{C15 90k} & \multicolumn{5}{c}{C30 160k} & \multicolumn{5}{c}{C45 260k}\\
Task & L0 & L1 & L2 & L3 & Avg. & L0 & L1 & L2 & L3 & Avg. & L0 & L1 & L2 & L3 & Avg.\\
\midrule
\multicolumn{16}{l}{\textit{Seen tasks}}\\
\midrule
FoldTowel & 100.0 & 100.0 & 80.0 & 90.0 & 92.5 & 100.0 & 90.0 & 80.0 & 100.0 & 92.5 & 100.0 & 100.0 & 90.0 & 90.0 & 95.0\\
PlaceFileFolder & 70.0 & 50.0 & 90.0 & 90.0 & 75.0 & 70.0 & 80.0 & 100.0 & 90.0 & 85.0 & 80.0 & 20.0 & 100.0 & 70.0 & 67.5\\
PlaceMugRack & 100.0 & 80.0 & 90.0 & 100.0 & 92.5 & 90.0 & 90.0 & 100.0 & 100.0 & 95.0 & 80.0 & 90.0 & 100.0 & 100.0 & 92.5\\
PlacePlateRack & 100.0 & 100.0 & 100.0 & 100.0 & 100.0 & 100.0 & 100.0 & 100.0 & 100.0 & 100.0 & 100.0 & 90.0 & 100.0 & 100.0 & 97.5\\
StirSpoon & 100.0 & 90.0 & 90.0 & 70.0 & 87.5 & 90.0 & 100.0 & 100.0 & 100.0 & 97.5 & 100.0 & 100.0 & 70.0 & 90.0 & 90.0\\
\midrule
\textbf{Seen average} & 94.0 & 84.0 & 90.0 & 90.0 & 89.5 & 90.0 & 92.0 & 96.0 & 98.0 & 94.0 & 92.0 & 80.0 & 92.0 & 90.0 & 88.5\\
\midrule
\multicolumn{16}{l}{\textit{Unseen tasks}}\\
\midrule
FoldBox & 0.0 & 0.0 & 0.0 & 0.0 & 0.0 & 0.0 & 0.0 & 10.0 & 0.0 & 2.5 & 90.0 & 0.0 & 0.0 & 100.0 & 47.5\\
PickFood & 0.0 & 0.0 & 0.0 & 0.0 & 0.0 & 0.0 & 0.0 & 0.0 & 0.0 & 0.0 & 0.0 & 0.0 & 0.0 & 0.0 & 0.0\\
PickRemoteControl & 0.0 & 0.0 & 0.0 & 0.0 & 0.0 & 0.0 & 0.0 & 0.0 & 0.0 & 0.0 & 0.0 & 0.0 & 0.0 & 0.0 & 0.0\\
PourKettle & 0.0 & 0.0 & 0.0 & 0.0 & 0.0 & 0.0 & 0.0 & 0.0 & 0.0 & 0.0 & 10.0 & 10.0 & 0.0 & 0.0 & 5.0\\
ScanMilkBox & 0.0 & 0.0 & 0.0 & 0.0 & 0.0 & 0.0 & 0.0 & 0.0 & 0.0 & 0.0 & 10.0 & 0.0 & 10.0 & 0.0 & 5.0\\
\midrule
\textbf{Unseen average} & 0.0 & 0.0 & 0.0 & 0.0 & 0.0 & 0.0 & 0.0 & 2.0 & 0.0 & 0.5 & 22.0 & 2.0 & 2.0 & 20.0 & 11.5\\
\bottomrule
\end{tabular}}\par
\caption{\textbf{Complete nested-corpus results with human references.} SR (\%) for C15 at 90k, C30 at 160k, and C45 at 260k on ten tasks and four scene levels. Each cell has ten trials. Each split has 200 episodes. These checkpoints are used in the main scale comparisons.}
\label{tab:selected-details}
\end{table}
\endgroup

\begin{multicols}{2}[][340pt]
\raggedcolumns
\input{tables/imitationgame_additional_checkpoints}

\begingroup
\input{tables/experiment_style}
\begin{table}[H]
\centering
\scriptsize
\setlength{\tabcolsep}{3pt}
\renewcommand{\arraystretch}{1.05}
\begin{tabular}{@{}lrrrrr@{}}
\toprule
Task & L0 & L1 & L2 & L3 & Avg.\\
\midrule
FoldBox & 90.0 & 0.0 & 0.0 & 100.0 & 47.5\\
PickFood & 0.0 & 0.0 & 0.0 & 0.0 & 0.0\\
PickRemoteControl & 0.0 & 0.0 & 0.0 & 0.0 & 0.0\\
PourKettle & 10.0 & 10.0 & 0.0 & 0.0 & 5.0\\
ScanMilkBox & 10.0 & 0.0 & 10.0 & 0.0 & 5.0\\
\midrule
Unseen average & 22.0 & 2.0 & 2.0 & 20.0 & 11.5\\
\bottomrule
\end{tabular}
\caption{\textbf{C45 260k Unseen results.} SR (\%) for the checkpoint used in the main text, with ten trials per task--level cell.}
\label{tab:selected-unseen-c45}
\end{table}
\endgroup

%% file: tables/imitationgame_additional_checkpoints.tex
\begingroup
\input{tables/experiment_style}
\begin{table}[H]
\centering
\scriptsize
\setlength{\tabcolsep}{10pt}
\renewcommand{\arraystretch}{1}
\resizebox{\linewidth}{!}{%
\begin{tabular}{@{}l|rrrrr@{}}
\toprule
 & \multicolumn{5}{c}{C15-wo 100k (robot reference)}\\
Task & L0 & L1 & L2 & L3 & Avg.\\
\midrule
\multicolumn{6}{l}{\textit{Seen tasks}}\\
FoldTowel & 100.0 & 100.0 & 70.0 & 100.0 & 92.5\\
PlaceFileFolder & 100.0 & 60.0 & 100.0 & 100.0 & 90.0\\
PlaceMugRack & 90.0 & 90.0 & 100.0 & 100.0 & 95.0\\
PlacePlateRack & 90.0 & 100.0 & 100.0 & 100.0 & 97.5\\
StirSpoon & 90.0 & 80.0 & 90.0 & 40.0 & 75.0\\
\midrule
\textbf{Seen average} & 94.0 & 86.0 & 92.0 & 88.0 & 90.0\\
\midrule
\multicolumn{6}{l}{\textit{Policy-held-out tasks}}\\
FoldBox & 0.0 & 0.0 & 0.0 & 0.0 & 0.0\\
PickFood & 0.0 & 0.0 & 0.0 & 0.0 & 0.0\\
PickRemoteControl & 0.0 & 0.0 & 0.0 & 0.0 & 0.0\\
PourKettle & 0.0 & 0.0 & 0.0 & 0.0 & 0.0\\
ScanMilkBox & 0.0 & 0.0 & 0.0 & 0.0 & 0.0\\
\midrule
\textbf{Held-out average} & 0.0 & 0.0 & 0.0 & 0.0 & 0.0\\
\bottomrule
\end{tabular}
}
\caption{\textbf{Complete C15-wo 100k results with robot references.} SR (\%) on ten tasks and four scene levels. Each cell has ten trials. Each split has 200 episodes. Task-matched robot references come from the IG training bank, including references for policy-held-out tasks.}
\label{tab:additional-checkpoints}
\end{table}
\endgroup

%% file: tables/imitationgame_full50_all.tex
\end{multicols}
\begingroup
\input{tables/experiment_style}
\fontsize{7pt}{8pt}\selectfont
\tabcolsep=0.01cm
\renewcommand\arraystretch{0.96}
\setlength{\LTleft}{0pt}
\setlength{\LTright}{0pt plus 1fill}
\setlength{\LTcapwidth}{\linewidth}
\begin{longtable}{@{}>{\raggedright\arraybackslash}p{3.65cm}|*{15}{>{\centering\arraybackslash}p{\dimexpr(\linewidth-3.65cm-30\tabcolsep-\arrayrulewidth-0.01pt)/15\relax}}@{}}
\caption{\textbf{Complete full50 results.} SR (\%) for all 50 tasks, grouped by operation. Each level compares DP, ACT, and RoboDemo. Avg. averages L0--L3. All score columns have equal width. Only the Avg. cells are shaded: blue/light blue indicate the best/second-best among the three models, including ties.}
\label{tab:full50-all}\\
\toprule
Task & \multicolumn{3}{c}{L0} & \multicolumn{3}{c}{L1} & \multicolumn{3}{c}{L2} & \multicolumn{3}{c}{L3} & \multicolumn{3}{c}{Avg.}\\
\cmidrule(lr){2-4}\cmidrule(lr){5-7}\cmidrule(lr){8-10}\cmidrule(lr){11-13}\cmidrule(l){14-16}
 & DP & ACT & {\fontsize{5pt}{5.5pt}\selectfont RoboDemo} & DP & ACT & {\fontsize{5pt}{5.5pt}\selectfont RoboDemo} & DP & ACT & {\fontsize{5pt}{5.5pt}\selectfont RoboDemo} & DP & ACT & {\fontsize{5pt}{5.5pt}\selectfont RoboDemo} & DP & ACT & {\fontsize{5pt}{5.5pt}\selectfont RoboDemo}\\
\midrule
\endfirsthead
\multicolumn{16}{l}{Table~\thetable\ continued}\\
\toprule
Task & \multicolumn{3}{c}{L0} & \multicolumn{3}{c}{L1} & \multicolumn{3}{c}{L2} & \multicolumn{3}{c}{L3} & \multicolumn{3}{c}{Avg.}\\
\cmidrule(lr){2-4}\cmidrule(lr){5-7}\cmidrule(lr){8-10}\cmidrule(lr){11-13}\cmidrule(l){14-16}
 & DP & ACT & {\fontsize{5pt}{5.5pt}\selectfont RoboDemo} & DP & ACT & {\fontsize{5pt}{5.5pt}\selectfont RoboDemo} & DP & ACT & {\fontsize{5pt}{5.5pt}\selectfont RoboDemo} & DP & ACT & {\fontsize{5pt}{5.5pt}\selectfont RoboDemo} & DP & ACT & {\fontsize{5pt}{5.5pt}\selectfont RoboDemo}\\
\midrule
\endhead
\bottomrule
\endfoot
\bottomrule
\endlastfoot
\multicolumn{16}{l}{\textit{Transport and placement}}\\*
\midrule
KnifeBowlFork & 0.0 & 100 & 100 & 60.0 & 100 & 100 & 60.0 & 80.0 & 90.0 & 90.0 & 50.0 & 100 & 52.5 & \second{82.5} & \best{97.5}\\
PickAppleBananaToBaskets & 50.0 & 90.0 & 100 & 100 & 90.0 & 100 & 50.0 & 50.0 & 70.0 & 100 & 50.0 & 100 & \second{75.0} & 70.0 & \best{92.5}\\
PickAppleBasket & 70.0 & 100 & 90.0 & 70.0 & 90.0 & 100 & 100 & 100 & 100 & 80.0 & 100 & 100 & \second{80.0} & \best{97.5} & \best{97.5}\\
PickAppleToScale & 40.0 & 90.0 & 100 & 90.0 & 10.0 & 100 & 40.0 & 90.0 & 80.0 & 80.0 & 80.0 & 100 & 62.5 & \second{67.5} & \best{95.0}\\
PickFood & 40.0 & 100 & 100 & 100 & 70.0 & 100 & 10.0 & 30.0 & 60.0 & 90.0 & 90.0 & 100 & 60.0 & \second{72.5} & \best{90.0}\\
PickFruitsToPlate & 90.0 & 20.0 & 90.0 & 90.0 & 40.0 & 90.0 & 100 & 100 & 90.0 & 50.0 & 0.0 & 90.0 & \second{82.5} & 40.0 & \best{90.0}\\
PickPillToRegions & 40.0 & 90.0 & 80.0 & 60.0 & 90.0 & 90.0 & 50.0 & 70.0 & 50.0 & 10.0 & 80.0 & 50.0 & 40.0 & \best{82.5} & \second{67.5}\\
PickRemoteControl & 100 & 100 & 100 & 100 & 100 & 100 & 100 & 100 & 100 & 40.0 & 80.0 & 90.0 & 85.0 & \second{95.0} & \best{97.5}\\
PickTennisBallGolfBall & 90.0 & 60.0 & 70.0 & 90.0 & 80.0 & 100 & 90.0 & 70.0 & 30.0 & 100 & 50.0 & 90.0 & \best{92.5} & 65.0 & \second{72.5}\\
PlaceBookBookcase & 100 & 100 & 90.0 & 100 & 80.0 & 100 & 90.0 & 100 & 100 & 90.0 & 30.0 & 100 & \second{95.0} & 77.5 & \best{97.5}\\
PlaceBrushRest & 90.0 & 80.0 & 100 & 70.0 & 100 & 100 & 100 & 100 & 100 & 100 & 100 & 100 & 90.0 & \second{95.0} & \best{100}\\
PlaceBurgerTray & 80.0 & 90.0 & 70.0 & 90.0 & 80.0 & 80.0 & 80.0 & 90.0 & 100 & 70.0 & 70.0 & 80.0 & \second{80.0} & \best{82.5} & \best{82.5}\\
PlaceChipsRack & 90.0 & 70.0 & 100 & 80.0 & 70.0 & 100 & 100 & 70.0 & 100 & 80.0 & 100 & 100 & \second{87.5} & 77.5 & \best{100}\\
PlaceClothBasket & 100 & 100 & 100 & 100 & 100 & 40.0 & 80.0 & 90.0 & 100 & 10.0 & 90.0 & 50.0 & \second{72.5} & \best{95.0} & \second{72.5}\\
PlaceCommodityRack & 40.0 & 100 & 100 & 50.0 & 100 & 100 & 30.0 & 90.0 & 80.0 & 30.0 & 0.0 & 0.0 & 37.5 & \best{72.5} & \second{70.0}\\
PlaceCupPlate & 40.0 & 50.0 & 70.0 & 20.0 & 60.0 & 60.0 & 70.0 & 70.0 & 60.0 & 30.0 & 0.0 & 50.0 & 40.0 & \second{45.0} & \best{60.0}\\
PlaceFileFolder & 80.0 & 80.0 & 80.0 & 70.0 & 80.0 & 70.0 & 100 & 60.0 & 100 & 80.0 & 100 & 80.0 & \best{82.5} & \second{80.0} & \best{82.5}\\
PlaceFoodScale & 40.0 & 60.0 & 100 & 20.0 & 100 & 100 & 90.0 & 100 & 100 & 0.0 & 0.0 & 60.0 & 37.5 & \second{65.0} & \best{90.0}\\
PlaceFruitBox & 80.0 & 70.0 & 80.0 & 20.0 & 30.0 & 30.0 & 90.0 & 90.0 & 30.0 & 30.0 & 60.0 & 10.0 & \second{55.0} & \best{62.5} & 37.5\\
PlaceMagazineFolder & 80.0 & 100 & 100 & 100 & 100 & 100 & 0.0 & 0.0 & 0.0 & 70.0 & 70.0 & 80.0 & 62.5 & \second{67.5} & \best{70.0}\\
PlaceMugRack & 100 & 100 & 100 & 70.0 & 100 & 90.0 & 80.0 & 80.0 & 90.0 & 90.0 & 100 & 100 & \second{85.0} & \best{95.0} & \best{95.0}\\
PlacePillBox & 80.0 & 100 & 100 & 70.0 & 60.0 & 100 & 60.0 & 80.0 & 90.0 & 30.0 & 70.0 & 60.0 & 60.0 & \second{77.5} & \best{87.5}\\
PlacePlateRack & 90.0 & 100 & 100 & 40.0 & 90.0 & 100 & 30.0 & 50.0 & 100 & 80.0 & 80.0 & 100 & 60.0 & \second{80.0} & \best{100}\\
PlaceScrewdriver & 30.0 & 100 & 100 & 20.0 & 80.0 & 90.0 & 30.0 & 40.0 & 100 & 30.0 & 100 & 10.0 & 27.5 & \best{80.0} & \second{75.0}\\
PlaceShoeBox & 90.0 & 100 & 100 & 90.0 & 100 & 40.0 & 60.0 & 90.0 & 100 & 100 & 90.0 & 100 & \second{85.0} & \best{95.0} & \second{85.0}\\
PutBox & 100 & 100 & 100 & 90.0 & 90.0 & 100 & 90.0 & 100 & 100 & 10.0 & 100 & 90.0 & \second{72.5} & \best{97.5} & \best{97.5}\\
PutCubeOnScale & 100 & 100 & 100 & 90.0 & 100 & 100 & 90.0 & 100 & 90.0 & 60.0 & 70.0 & 70.0 & 85.0 & \best{92.5} & \second{90.0}\\
TransFood & 50.0 & 60.0 & 100 & 90.0 & 90.0 & 100 & 30.0 & 0.0 & 100 & 30.0 & 20.0 & 100 & \second{50.0} & 42.5 & \best{100}\\
\midrule
\textbf{Transport and placement (Avg.)} & 70.7 & 86.1 & 93.6 & 72.9 & 81.4 & 88.6 & 67.9 & 74.6 & 82.5 & 59.3 & 65.4 & 77.1 & 67.7 & \second{76.9} & \best{85.4}\\
\midrule
\multicolumn{16}{l}{\textit{Pouring}}\\*
\midrule
PourCup & 10.0 & 100 & 100 & 100 & 100 & 100 & 100 & 100 & 90.0 & 90.0 & 100 & 90.0 & 75.0 & \best{100} & \second{95.0}\\
PourKetchupFries & 50.0 & 100 & 100 & 50.0 & 100 & 100 & 100 & 80.0 & 100 & 30.0 & 60.0 & 50.0 & 57.5 & \second{85.0} & \best{87.5}\\
PourKettle & 90.0 & 100 & 100 & 80.0 & 90.0 & 80.0 & 0.0 & 0.0 & 0.0 & 80.0 & 100 & 100 & 62.5 & \best{72.5} & \second{70.0}\\
PourLiquidCup & 10.0 & 10.0 & 100 & 0.0 & 10.0 & 90.0 & 0.0 & 0.0 & 90.0 & 10.0 & 0.0 & 60.0 & \second{5.0} & \second{5.0} & \best{85.0}\\
PourLiquidFilter & 60.0 & 100 & 100 & 60.0 & 100 & 10.0 & 20.0 & 90.0 & 30.0 & 50.0 & 90.0 & 100 & 47.5 & \best{95.0} & \second{60.0}\\
PourLiquidMug & 60.0 & 0.0 & 100 & 10.0 & 90.0 & 100 & 100 & 70.0 & 100 & 20.0 & 0.0 & 100 & \second{47.5} & 40.0 & \best{100}\\
\midrule
\textbf{Pouring (Avg.)} & 46.7 & 68.3 & 100 & 50.0 & 81.7 & 80.0 & 53.3 & 56.7 & 68.3 & 46.7 & 58.3 & 83.3 & 49.2 & \second{66.3} & \best{82.9}\\
\midrule
\multicolumn{16}{l}{\textit{Cleaning}}\\*
\midrule
CleanCup & 40.0 & 30.0 & 60.0 & 50.0 & 30.0 & 100 & 20.0 & 70.0 & 100 & 100 & 90.0 & 100 & 52.5 & \second{55.0} & \best{90.0}\\
CleanDesk & 10.0 & 80.0 & 0.0 & 60.0 & 100 & 0.0 & 40.0 & 100 & 50.0 & 60.0 & 100 & 100 & \second{42.5} & \best{95.0} & 37.5\\
PickWash & 70.0 & 90.0 & 80.0 & 100 & 100 & 100 & 30.0 & 60.0 & 80.0 & 80.0 & 90.0 & 80.0 & \second{70.0} & \best{85.0} & \best{85.0}\\
WipePot & 10.0 & 20.0 & 100 & 0.0 & 0.0 & 100 & 0.0 & 0.0 & 0.0 & 100 & 80.0 & 100 & \second{27.5} & 25.0 & \best{75.0}\\
\midrule
\textbf{Cleaning (Avg.)} & 32.5 & 55.0 & 60.0 & 52.5 & 57.5 & 75.0 & 22.5 & 57.5 & 57.5 & 85.0 & 90.0 & 95.0 & 48.1 & \second{65.0} & \best{71.9}\\
\midrule
\multicolumn{16}{l}{\textit{Opening and folding}}\\*
\midrule
FoldBox & 100 & 100 & 100 & 100 & 100 & 100 & 100 & 100 & 100 & 100 & 100 & 100 & \best{100} & \best{100} & \best{100}\\
FoldTowel & 60.0 & 100 & 100 & 40.0 & 100 & 100 & 10.0 & 100 & 100 & 40.0 & 80.0 & 100 & 37.5 & \second{95.0} & \best{100}\\
LiftLidFromSkillet & 0.0 & 100 & 90.0 & 80.0 & 100 & 90.0 & 80.0 & 100 & 100 & 100 & 100 & 100 & 65.0 & \best{100} & \second{95.0}\\
OpenBox & 100 & 100 & 100 & 90.0 & 100 & 100 & 0.0 & 0.0 & 0.0 & 100 & 100 & 90.0 & \second{72.5} & \best{75.0} & \second{72.5}\\
OpenLiquidCap & 60.0 & 90.0 & 80.0 & 50.0 & 60.0 & 90.0 & 70.0 & 90.0 & 100 & 50.0 & 0.0 & 90.0 & 57.5 & \second{60.0} & \best{90.0}\\
\midrule
\textbf{Opening and folding (Avg.)} & 64.0 & 98.0 & 94.0 & 72.0 & 92.0 & 96.0 & 52.0 & 78.0 & 80.0 & 78.0 & 76.0 & 96.0 & 66.5 & \second{86.0} & \best{91.5}\\
\midrule
\multicolumn{16}{l}{\textit{Tool use and scanning}}\\*
\midrule
CutFruit & 100 & 100 & 100 & 50.0 & 100 & 100 & 0.0 & 0.0 & 0.0 & 100 & 100 & 100 & \second{62.5} & \best{75.0} & \best{75.0}\\
GrindFood & 100 & 90.0 & 100 & 100 & 100 & 100 & 100 & 100 & 100 & 100 & 70.0 & 90.0 & \best{100} & 90.0 & \second{97.5}\\
PressJuicer & 60.0 & 20.0 & 100 & 80.0 & 100 & 100 & 0.0 & 0.0 & 0.0 & 100 & 100 & 100 & \second{60.0} & 55.0 & \best{75.0}\\
PressStapler & 100 & 50.0 & 100 & 70.0 & 90.0 & 100 & 100 & 40.0 & 100 & 90.0 & 40.0 & 70.0 & \second{90.0} & 55.0 & \best{92.5}\\
ScanMilkBox & 100 & 100 & 100 & 100 & 100 & 90.0 & 100 & 100 & 70.0 & 90.0 & 100 & 40.0 & \second{97.5} & \best{100} & 75.0\\
ScanPillBottle & 90.0 & 100 & 100 & 50.0 & 100 & 90.0 & 90.0 & 90.0 & 100 & 10.0 & 0.0 & 100 & 60.0 & \second{72.5} & \best{97.5}\\
StirSpoon & 20.0 & 50.0 & 100 & 0.0 & 50.0 & 90.0 & 40.0 & 40.0 & 70.0 & 10.0 & 80.0 & 70.0 & 17.5 & \second{55.0} & \best{82.5}\\
\midrule
\textbf{Tool use and scanning (Avg.)} & 81.4 & 72.9 & 100 & 64.3 & 91.4 & 95.7 & 61.4 & 52.9 & 62.9 & 71.4 & 70.0 & 81.4 & 69.6 & \second{71.8} & \best{85.0}\\
\midrule
\textbf{Overall average} & 65.6 & 80.8 & 92.6 & 67.2 & 82.0 & 88.2 & 60.0 & 68.4 & 75.8 & 63.4 & 68.2 & 81.8 & 64.1 & \second{74.9} & \best{84.6}\\
\end{longtable}
\endgroup
\begin{multicols}{2}
\raggedcolumns

%% file: tables/imitationgame_baseline_aggregates.tex
\begingroup
\input{tables/experiment_style}
\begin{table}[H]
\centering
\scriptsize
\tabcolsep=0.1cm
\resizebox{1.0\linewidth}{!}{%
\tabcolsep=10pt
\renewcommand\arraystretch{1.0}
\begin{tabular}{@{}l|rrrr|r@{}}
\toprule
 & \multicolumn{4}{c|}{Seen} & Unseen\\
\cmidrule(lr){2-5}\cmidrule(l){6-6}
Method & C15 & C30 & C45 & Avg. & C45\\
\midrule
GR00T-N1.6 & 29.5 & 48.0 & 74.5 & 50.7 & 5.5\\
OpenVLA & 39.0 & 25.5 & 22.0 & 28.8 & 6.0\\
$\pi_{0.5}$ & 62.5 & 74.0 & \second{82.0} & 72.8 & 10.0\\
RDT-1B & 37.0 & 50.5 & 41.0 & 42.8 & 4.5\\
UniSkill & 78.0 & 69.5 & 76.5 & 74.7 & 4.5\\
XSkill & 79.5 & 76.5 & 81.5 & 79.2 & 5.0\\
ACT / DINOv2 & 80.5 & \second{81.5} & 80.5 & \second{80.8} & 1.0\\
ACT / SigLIP2 & \second{82.0} & 76.5 & 77.5 & 78.7 & 2.0\\
ACT / VideoMAE & 80.5 & 69.0 & 67.0 & 72.2 & 2.5\\
DP / DINOv2 & 65.0 & 68.0 & 68.5 & 67.2 & 5.5\\
DP / SigLIP2 & 60.5 & 58.5 & 63.0 & 60.7 & 3.0\\
DP / VideoMAE & 23.5 & 17.0 & 2.5 & 14.3 & 3.0\\
VQ-BeT / DINOv2 & 45.0 & 52.0 & 58.5 & 51.8 & \second{17.5}\\
VQ-BeT / SigLIP2 & 58.5 & 56.0 & 58.0 & 57.5 & 10.5\\
VQ-BeT / VideoMAE & 44.0 & 24.5 & 20.5 & 29.7 & 7.5\\
Zero-WAM & 68.0 & 77.0 & 68.0 & 71.0 & \best{22.0}\\
\midrule
\textbf{RoboDemo} & \best{89.5} & \best{94.0} & \best{88.5} & \best{90.7} & 11.5\\
\bottomrule
\end{tabular}
}
\caption{\textbf{Complete aggregate comparison.} SR (\%) for all 16 official baselines and RoboDemo. Seen Avg. weights the three training scales equally. C45 uses the 260k checkpoint. Blue/light blue mark the best/second-best per column.}
\label{tab:baseline-aggregates}
\end{table}
\endgroup

%% file: tables/imitationgame_baseline_details.tex
\begingroup
\input{tables/experiment_style}
\begin{appendixresulttable}
\caption{\textbf{Complete C15 Seen results.} SR (\%) for every task and scene level. D/S/V denote DINOv2, SigLIP2, and VideoMAE. RoboDemo is repeated in both panels for comparison. Blue/light blue mark the best/second-best among all 17 methods for the same row, across both panels.}
\label{tab:baseline-seen-15}
\nopagebreak
\centering
\fontsize{7pt}{8pt}\selectfont
\resizebox{\linewidth}{!}{%
\tabcolsep=1pt
\renewcommand\arraystretch{1.0}
\begin{tabular}{@{}lc|*{8}{c}@{}}
\toprule
\multicolumn{10}{c}{(a) Language-conditioned, video-skill, and in-context policies}\\
\midrule
Task & Level & \shortstack{GR00T-\\N1.6} & \shortstack{Open\\VLA} & $\pi_{0.5}$ & \shortstack{RDT-\\1B} & \shortstack{Uni\\Skill} & XSkill & \shortstack{Zero-\\WAM} & \shortstack{Robo\\Demo}\\
\midrule
FoldTowel & L0 & 0.0 & 40.0 & \best{100} & 50.0 & \best{100} & 80.0 & \best{100} & \best{100}\\
 & L1 & 0.0 & 0.0 & 10.0 & 0.0 & \best{100} & \best{100} & \best{100} & \best{100}\\
 & L2 & 0.0 & 20.0 & 10.0 & 0.0 & \second{90.0} & 80.0 & 60.0 & 80.0\\
 & L3 & 0.0 & 50.0 & 50.0 & 40.0 & 60.0 & 70.0 & \second{90.0} & \second{90.0}\\
\midrule
PlaceFileFolder & L0 & 0.0 & \best{90.0} & 50.0 & 70.0 & \best{90.0} & \best{90.0} & 60.0 & 70.0\\
 & L1 & 0.0 & 0.0 & 10.0 & 0.0 & 70.0 & 80.0 & 0.0 & 50.0\\
 & L2 & 60.0 & 20.0 & 30.0 & 10.0 & 40.0 & 20.0 & 80.0 & \second{90.0}\\
 & L3 & 40.0 & \second{90.0} & 70.0 & \best{100} & \best{100} & 80.0 & 30.0 & \second{90.0}\\
\midrule
PlaceMugRack & L0 & 0.0 & 70.0 & \second{90.0} & 10.0 & \second{90.0} & \second{90.0} & 50.0 & \best{100}\\
 & L1 & 0.0 & 40.0 & \best{100} & 0.0 & \second{90.0} & \best{100} & 40.0 & 80.0\\
 & L2 & 20.0 & 0.0 & 80.0 & 60.0 & \best{100} & \second{90.0} & 80.0 & \second{90.0}\\
 & L3 & \best{100} & \second{90.0} & \second{90.0} & \second{90.0} & \best{100} & \second{90.0} & 0.0 & \best{100}\\
\midrule
PlacePlateRack & L0 & 60.0 & 10.0 & \best{100} & 60.0 & 80.0 & \second{90.0} & \best{100} & \best{100}\\
 & L1 & 70.0 & 30.0 & 70.0 & 70.0 & 80.0 & 70.0 & 80.0 & \best{100}\\
 & L2 & 30.0 & 50.0 & \second{90.0} & 50.0 & 50.0 & 60.0 & 50.0 & \best{100}\\
 & L3 & 80.0 & 60.0 & 80.0 & 50.0 & \best{100} & \second{90.0} & \second{90.0} & \best{100}\\
\midrule
StirSpoon & L0 & 60.0 & 60.0 & \second{90.0} & 0.0 & 70.0 & \second{90.0} & \best{100} & \best{100}\\
 & L1 & 0.0 & 60.0 & 50.0 & 0.0 & 60.0 & 80.0 & \best{100} & \second{90.0}\\
 & L2 & 40.0 & 0.0 & 70.0 & 20.0 & 60.0 & \best{100} & 80.0 & \second{90.0}\\
 & L3 & 30.0 & 0.0 & 10.0 & 60.0 & 30.0 & 40.0 & \second{70.0} & \second{70.0}\\
\midrule
\textbf{Overall} &  & 29.5 & 39.0 & 62.5 & 37.0 & 78.0 & 79.5 & 68.0 & \best{89.5}\\
\bottomrule
\end{tabular}
}
\par\vspace{7pt}
\fontsize{7pt}{8pt}\selectfont
\resizebox{\linewidth}{!}{%
\tabcolsep=1pt
\renewcommand\arraystretch{1.0}
\begin{tabular}{@{}lc|*{10}{c}@{}}
\toprule
\multicolumn{12}{c}{(b) Video-conditioned policies}\\
\midrule
Task & Level & \shortstack{ACT\\D} & \shortstack{ACT\\S} & \shortstack{ACT\\V} & \shortstack{DP\\D} & \shortstack{DP\\S} & \shortstack{DP\\V} & \shortstack{VQ-BeT\\D} & \shortstack{VQ-BeT\\S} & \shortstack{VQ-BeT\\V} & \shortstack{Robo\\Demo}\\
\midrule
FoldTowel & L0 & 80.0 & \best{100} & \best{100} & \second{90.0} & 50.0 & 40.0 & \best{100} & \best{100} & \second{90.0} & \best{100}\\
 & L1 & \best{100} & \best{100} & \best{100} & 70.0 & 50.0 & 30.0 & 70.0 & \second{90.0} & \best{100} & \best{100}\\
 & L2 & \best{100} & \best{100} & \second{90.0} & 30.0 & 10.0 & 10.0 & 10.0 & 30.0 & 20.0 & 80.0\\
 & L3 & \second{90.0} & 60.0 & 80.0 & \best{100} & 80.0 & 10.0 & 50.0 & 70.0 & 70.0 & \second{90.0}\\
\midrule
PlaceFileFolder & L0 & \best{90.0} & 60.0 & \second{80.0} & \second{80.0} & \best{90.0} & \second{80.0} & 50.0 & \second{80.0} & 60.0 & 70.0\\
 & L1 & \best{100} & \best{100} & \second{90.0} & \second{90.0} & 50.0 & 50.0 & 80.0 & \second{90.0} & 70.0 & 50.0\\
 & L2 & \best{100} & 80.0 & 60.0 & 60.0 & 80.0 & 40.0 & 70.0 & \best{100} & 60.0 & \second{90.0}\\
 & L3 & \best{100} & 80.0 & \second{90.0} & \second{90.0} & \second{90.0} & 10.0 & 60.0 & 80.0 & 80.0 & \second{90.0}\\
\midrule
PlaceMugRack & L0 & \best{100} & \best{100} & \best{100} & 80.0 & \best{100} & 0.0 & 0.0 & 30.0 & 0.0 & \best{100}\\
 & L1 & 70.0 & 80.0 & 80.0 & \best{100} & \best{100} & 20.0 & \second{90.0} & 60.0 & 0.0 & 80.0\\
 & L2 & \best{100} & \second{90.0} & 80.0 & 80.0 & \second{90.0} & 0.0 & 0.0 & 10.0 & 0.0 & \second{90.0}\\
 & L3 & \best{100} & \best{100} & \second{90.0} & 80.0 & \best{100} & \second{90.0} & 50.0 & 80.0 & 10.0 & \best{100}\\
\midrule
PlacePlateRack & L0 & 80.0 & 70.0 & \second{90.0} & 80.0 & \best{100} & 0.0 & 20.0 & 80.0 & \second{90.0} & \best{100}\\
 & L1 & 70.0 & 80.0 & \best{100} & 60.0 & 40.0 & 0.0 & 70.0 & \second{90.0} & 60.0 & \best{100}\\
 & L2 & 50.0 & 70.0 & 50.0 & 30.0 & 40.0 & 0.0 & 0.0 & 30.0 & 10.0 & \best{100}\\
 & L3 & \best{100} & \best{100} & \best{100} & 80.0 & 50.0 & 10.0 & 10.0 & 50.0 & 60.0 & \best{100}\\
\midrule
StirSpoon & L0 & 40.0 & 80.0 & \second{90.0} & 20.0 & 20.0 & 0.0 & 20.0 & 20.0 & 10.0 & \best{100}\\
 & L1 & 10.0 & 60.0 & 40.0 & 10.0 & 10.0 & 10.0 & 50.0 & 0.0 & 70.0 & \second{90.0}\\
 & L2 & 50.0 & 80.0 & 40.0 & 50.0 & 50.0 & 40.0 & 20.0 & 40.0 & 10.0 & \second{90.0}\\
 & L3 & \best{80.0} & 50.0 & 60.0 & 20.0 & 10.0 & 30.0 & \best{80.0} & 40.0 & 10.0 & \second{70.0}\\
\midrule
\textbf{Overall} &  & 80.5 & \second{82.0} & 80.5 & 65.0 & 60.5 & 23.5 & 45.0 & 58.5 & 44.0 & \best{89.5}\\
\bottomrule
\end{tabular}
}
\end{appendixresulttable}
\endgroup
\begingroup
\input{tables/experiment_style}
\begin{appendixresulttable}
\caption{\textbf{Complete C30 Seen results.} SR (\%) for every task and scene level. D/S/V denote DINOv2, SigLIP2, and VideoMAE. RoboDemo is repeated in both panels for comparison. Blue/light blue mark the best/second-best among all 17 methods for the same row, across both panels.}
\label{tab:baseline-seen-30}
\nopagebreak
\centering
\fontsize{7pt}{8pt}\selectfont
\resizebox{\linewidth}{!}{%
\tabcolsep=1pt
\renewcommand\arraystretch{1.0}
\begin{tabular}{@{}lc|*{8}{c}@{}}
\toprule
\multicolumn{10}{c}{(a) Language-conditioned, video-skill, and in-context policies}\\
\midrule
Task & Level & \shortstack{GR00T-\\N1.6} & \shortstack{Open\\VLA} & $\pi_{0.5}$ & \shortstack{RDT-\\1B} & \shortstack{Uni\\Skill} & XSkill & \shortstack{Zero-\\WAM} & \shortstack{Robo\\Demo}\\
\midrule
FoldTowel & L0 & 30.0 & 10.0 & \best{100} & \second{90.0} & \best{100} & 80.0 & \best{100} & \best{100}\\
 & L1 & 70.0 & 0.0 & \best{100} & 20.0 & \second{90.0} & \best{100} & 80.0 & \second{90.0}\\
 & L2 & 0.0 & 0.0 & \second{90.0} & 0.0 & 80.0 & \best{100} & 70.0 & 80.0\\
 & L3 & 0.0 & 0.0 & \second{90.0} & 80.0 & 80.0 & 30.0 & 80.0 & \best{100}\\
\midrule
PlaceFileFolder & L0 & \second{80.0} & 50.0 & 70.0 & 10.0 & 0.0 & \best{90.0} & \best{90.0} & 70.0\\
 & L1 & 70.0 & 10.0 & 20.0 & 60.0 & 0.0 & \best{100} & 20.0 & 80.0\\
 & L2 & 30.0 & 0.0 & 70.0 & 10.0 & 0.0 & 50.0 & 80.0 & \best{100}\\
 & L3 & 50.0 & \second{90.0} & 60.0 & \second{90.0} & \best{100} & 60.0 & \second{90.0} & \second{90.0}\\
\midrule
PlaceMugRack & L0 & 40.0 & 30.0 & \second{90.0} & 60.0 & \best{100} & \second{90.0} & \best{100} & \second{90.0}\\
 & L1 & 70.0 & 30.0 & \best{100} & 30.0 & \best{100} & \best{100} & 50.0 & \second{90.0}\\
 & L2 & 60.0 & 0.0 & \second{80.0} & 40.0 & \best{100} & \best{100} & \best{100} & \best{100}\\
 & L3 & \second{90.0} & \second{90.0} & \second{90.0} & \best{100} & 70.0 & 20.0 & 10.0 & \best{100}\\
\midrule
PlacePlateRack & L0 & \second{90.0} & 40.0 & \second{90.0} & 60.0 & 80.0 & \second{90.0} & \best{100} & \best{100}\\
 & L1 & 60.0 & 0.0 & \second{90.0} & 80.0 & \second{90.0} & \second{90.0} & 80.0 & \best{100}\\
 & L2 & \best{100} & 0.0 & 50.0 & 70.0 & \second{80.0} & 70.0 & 60.0 & \best{100}\\
 & L3 & \best{100} & 80.0 & \second{90.0} & 50.0 & \best{100} & \second{90.0} & \second{90.0} & \best{100}\\
\midrule
StirSpoon & L0 & 20.0 & 20.0 & 60.0 & 40.0 & 70.0 & 80.0 & \best{100} & \second{90.0}\\
 & L1 & 0.0 & 40.0 & 60.0 & 0.0 & 60.0 & 40.0 & 60.0 & \best{100}\\
 & L2 & 0.0 & 0.0 & 50.0 & 30.0 & 60.0 & \second{90.0} & \second{90.0} & \best{100}\\
 & L3 & 0.0 & 20.0 & 30.0 & \second{90.0} & 30.0 & 60.0 & \second{90.0} & \best{100}\\
\midrule
\textbf{Overall} &  & 48.0 & 25.5 & 74.0 & 50.5 & 69.5 & 76.5 & 77.0 & \best{94.0}\\
\bottomrule
\end{tabular}
}
\par\vspace{7pt}
\fontsize{7pt}{8pt}\selectfont
\resizebox{\linewidth}{!}{%
\tabcolsep=1pt
\renewcommand\arraystretch{1.0}
\begin{tabular}{@{}lc|*{10}{c}@{}}
\toprule
\multicolumn{12}{c}{(b) Video-conditioned policies}\\
\midrule
Task & Level & \shortstack{ACT\\D} & \shortstack{ACT\\S} & \shortstack{ACT\\V} & \shortstack{DP\\D} & \shortstack{DP\\S} & \shortstack{DP\\V} & \shortstack{VQ-BeT\\D} & \shortstack{VQ-BeT\\S} & \shortstack{VQ-BeT\\V} & \shortstack{Robo\\Demo}\\
\midrule
FoldTowel & L0 & \best{100} & \second{90.0} & \best{100} & \best{100} & 50.0 & 0.0 & 20.0 & 70.0 & 0.0 & \best{100}\\
 & L1 & \best{100} & \best{100} & \best{100} & 70.0 & 0.0 & 0.0 & \second{90.0} & \best{100} & 10.0 & \second{90.0}\\
 & L2 & \second{90.0} & \second{90.0} & 50.0 & 10.0 & 10.0 & 0.0 & 0.0 & 30.0 & 0.0 & 80.0\\
 & L3 & \second{90.0} & 80.0 & \second{90.0} & 80.0 & 70.0 & 0.0 & 0.0 & 40.0 & 20.0 & \best{100}\\
\midrule
PlaceFileFolder & L0 & 70.0 & \best{90.0} & 40.0 & \best{90.0} & 70.0 & 0.0 & 60.0 & 60.0 & 30.0 & 70.0\\
 & L1 & \second{90.0} & 60.0 & 60.0 & \second{90.0} & 50.0 & 10.0 & \second{90.0} & 80.0 & 0.0 & 80.0\\
 & L2 & 60.0 & 30.0 & 0.0 & \second{90.0} & 80.0 & 40.0 & \second{90.0} & 60.0 & \best{100} & \best{100}\\
 & L3 & \best{100} & \best{100} & \best{100} & 80.0 & \best{100} & 20.0 & 80.0 & 70.0 & 0.0 & \second{90.0}\\
\midrule
PlaceMugRack & L0 & \best{100} & \second{90.0} & \best{100} & \best{100} & 80.0 & 10.0 & 60.0 & 50.0 & 0.0 & \second{90.0}\\
 & L1 & \second{90.0} & 70.0 & \second{90.0} & \second{90.0} & \second{90.0} & 70.0 & 70.0 & 80.0 & 0.0 & \second{90.0}\\
 & L2 & \second{80.0} & \best{100} & \best{100} & \best{100} & \best{100} & 60.0 & 70.0 & 60.0 & 0.0 & \best{100}\\
 & L3 & \best{100} & \best{100} & \second{90.0} & \second{90.0} & \best{100} & \best{100} & 80.0 & \best{100} & 40.0 & \best{100}\\
\midrule
PlacePlateRack & L0 & \best{100} & 70.0 & 80.0 & \second{90.0} & 80.0 & 20.0 & 80.0 & \second{90.0} & 80.0 & \best{100}\\
 & L1 & 80.0 & \second{90.0} & 80.0 & 60.0 & 60.0 & 0.0 & 70.0 & 70.0 & 50.0 & \best{100}\\
 & L2 & \second{80.0} & \second{80.0} & 60.0 & 60.0 & 20.0 & 0.0 & 10.0 & 30.0 & 0.0 & \best{100}\\
 & L3 & \second{90.0} & \second{90.0} & \second{90.0} & 80.0 & \second{90.0} & 0.0 & 20.0 & 30.0 & 60.0 & \best{100}\\
\midrule
StirSpoon & L0 & 70.0 & 50.0 & 60.0 & 10.0 & 20.0 & 0.0 & 30.0 & 20.0 & 10.0 & \second{90.0}\\
 & L1 & 60.0 & 60.0 & 40.0 & 30.0 & 20.0 & 10.0 & 30.0 & 0.0 & \second{70.0} & \best{100}\\
 & L2 & 20.0 & 20.0 & 10.0 & 30.0 & 60.0 & 0.0 & 70.0 & 40.0 & 10.0 & \best{100}\\
 & L3 & 60.0 & 70.0 & 40.0 & 10.0 & 20.0 & 0.0 & 20.0 & 40.0 & 10.0 & \best{100}\\
\midrule
\textbf{Overall} &  & \second{81.5} & 76.5 & 69.0 & 68.0 & 58.5 & 17.0 & 52.0 & 56.0 & 24.5 & \best{94.0}\\
\bottomrule
\end{tabular}
}
\end{appendixresulttable}
\endgroup
\begingroup
\input{tables/experiment_style}
\begin{appendixresulttable}
\caption{\textbf{Complete C45 Seen results.} SR (\%) for every task and scene level. D/S/V denote DINOv2, SigLIP2, and VideoMAE. RoboDemo is repeated in both panels for comparison. Blue/light blue mark the best/second-best among all 17 methods for the same row, across both panels.}
\label{tab:baseline-seen-45}
\nopagebreak
\centering
\fontsize{7pt}{8pt}\selectfont
\resizebox{\linewidth}{!}{%
\tabcolsep=1pt
\renewcommand\arraystretch{1.0}
\begin{tabular}{@{}lc|*{8}{c}@{}}
\toprule
\multicolumn{10}{c}{(a) Language-conditioned, video-skill, and in-context policies}\\
\midrule
Task & Level & \shortstack{GR00T-\\N1.6} & \shortstack{Open\\VLA} & $\pi_{0.5}$ & \shortstack{RDT-\\1B} & \shortstack{Uni\\Skill} & XSkill & \shortstack{Zero-\\WAM} & \shortstack{Robo\\Demo}\\
\midrule
FoldTowel & L0 & \second{90.0} & 0.0 & \best{100} & 70.0 & \best{100} & \best{100} & \best{100} & \best{100}\\
 & L1 & 70.0 & 0.0 & \best{100} & 80.0 & \second{90.0} & \best{100} & \second{90.0} & \best{100}\\
 & L2 & 60.0 & 0.0 & \second{90.0} & 30.0 & \best{100} & 60.0 & 70.0 & \second{90.0}\\
 & L3 & 50.0 & 0.0 & 80.0 & 40.0 & 70.0 & 80.0 & 70.0 & \second{90.0}\\
\midrule
PlaceFileFolder & L0 & \second{90.0} & 50.0 & 80.0 & 40.0 & \second{90.0} & \best{100} & 30.0 & 80.0\\
 & L1 & \second{90.0} & 30.0 & 40.0 & 0.0 & 20.0 & 60.0 & 30.0 & 20.0\\
 & L2 & 70.0 & 0.0 & 80.0 & 0.0 & 40.0 & 80.0 & 70.0 & \best{100}\\
 & L3 & \second{90.0} & 60.0 & 70.0 & \second{90.0} & \best{100} & \best{100} & 20.0 & 70.0\\
\midrule
PlaceMugRack & L0 & \best{100} & 0.0 & 80.0 & 80.0 & \second{90.0} & \second{90.0} & 70.0 & 80.0\\
 & L1 & 60.0 & 60.0 & \best{100} & 0.0 & \best{100} & \best{100} & 70.0 & \second{90.0}\\
 & L2 & 60.0 & 0.0 & \second{90.0} & 0.0 & \best{100} & \best{100} & 70.0 & \best{100}\\
 & L3 & \second{90.0} & 60.0 & \second{90.0} & \best{100} & \second{90.0} & 40.0 & 0.0 & \best{100}\\
\midrule
PlacePlateRack & L0 & 60.0 & 0.0 & 70.0 & 40.0 & 80.0 & \best{100} & \second{90.0} & \best{100}\\
 & L1 & \best{100} & 0.0 & \second{90.0} & 80.0 & 60.0 & \best{100} & \best{100} & \second{90.0}\\
 & L2 & 70.0 & 10.0 & \second{90.0} & 60.0 & \second{90.0} & 80.0 & 40.0 & \best{100}\\
 & L3 & \second{90.0} & 30.0 & \best{100} & 40.0 & \best{100} & \best{100} & 70.0 & \best{100}\\
\midrule
StirSpoon & L0 & \second{80.0} & \second{80.0} & \best{100} & 20.0 & 40.0 & 40.0 & \best{100} & \best{100}\\
 & L1 & 80.0 & 30.0 & 70.0 & 0.0 & 70.0 & 50.0 & \second{90.0} & \best{100}\\
 & L2 & 30.0 & 10.0 & \second{80.0} & 10.0 & \second{80.0} & \second{80.0} & \best{100} & 70.0\\
 & L3 & 60.0 & 20.0 & 40.0 & 40.0 & 20.0 & 70.0 & \second{80.0} & \best{90.0}\\
\midrule
\textbf{Overall} &  & 74.5 & 22.0 & \second{82.0} & 41.0 & 76.5 & 81.5 & 68.0 & \best{88.5}\\
\bottomrule
\end{tabular}
}
\par\vspace{7pt}
\fontsize{7pt}{8pt}\selectfont
\resizebox{\linewidth}{!}{%
\tabcolsep=1pt
\renewcommand\arraystretch{1.0}
\begin{tabular}{@{}lc|*{10}{c}@{}}
\toprule
\multicolumn{12}{c}{(b) Video-conditioned policies}\\
\midrule
Task & Level & \shortstack{ACT\\D} & \shortstack{ACT\\S} & \shortstack{ACT\\V} & \shortstack{DP\\D} & \shortstack{DP\\S} & \shortstack{DP\\V} & \shortstack{VQ-BeT\\D} & \shortstack{VQ-BeT\\S} & \shortstack{VQ-BeT\\V} & \shortstack{Robo\\Demo}\\
\midrule
FoldTowel & L0 & \best{100} & \best{100} & \best{100} & 70.0 & \best{100} & 0.0 & 60.0 & 70.0 & 80.0 & \best{100}\\
 & L1 & \best{100} & \second{90.0} & \best{100} & 80.0 & 70.0 & 0.0 & \best{100} & \best{100} & 0.0 & \best{100}\\
 & L2 & 70.0 & 50.0 & \second{90.0} & 80.0 & 30.0 & 0.0 & 30.0 & 30.0 & 0.0 & \second{90.0}\\
 & L3 & \second{90.0} & 70.0 & \best{100} & 60.0 & 60.0 & 0.0 & 50.0 & 40.0 & 20.0 & \second{90.0}\\
\midrule
PlaceFileFolder & L0 & 80.0 & \best{100} & 30.0 & 70.0 & 60.0 & 20.0 & 70.0 & 80.0 & 60.0 & 80.0\\
 & L1 & 60.0 & 60.0 & 0.0 & \second{90.0} & \best{100} & 0.0 & 80.0 & \second{90.0} & 80.0 & 20.0\\
 & L2 & 50.0 & \second{90.0} & 10.0 & 80.0 & 40.0 & 30.0 & 60.0 & 70.0 & 60.0 & \best{100}\\
 & L3 & \second{90.0} & 80.0 & 80.0 & 60.0 & 70.0 & 0.0 & 60.0 & \second{90.0} & \best{100} & 70.0\\
\midrule
PlaceMugRack & L0 & \second{90.0} & \best{100} & \best{100} & \best{100} & 80.0 & 0.0 & \best{100} & 80.0 & 0.0 & 80.0\\
 & L1 & \second{90.0} & \second{90.0} & \best{100} & 70.0 & \second{90.0} & 0.0 & \best{100} & 70.0 & 0.0 & \second{90.0}\\
 & L2 & \best{100} & \best{100} & \second{90.0} & \second{90.0} & \best{100} & 0.0 & 0.0 & 60.0 & 0.0 & \best{100}\\
 & L3 & \best{100} & \best{100} & \best{100} & \second{90.0} & \second{90.0} & 0.0 & \best{100} & 60.0 & 0.0 & \best{100}\\
\midrule
PlacePlateRack & L0 & \best{100} & \second{90.0} & 80.0 & 60.0 & \best{100} & 0.0 & 30.0 & 70.0 & 0.0 & \best{100}\\
 & L1 & \second{90.0} & \second{90.0} & 80.0 & 40.0 & 60.0 & 0.0 & 60.0 & 50.0 & 10.0 & \second{90.0}\\
 & L2 & 70.0 & 70.0 & 80.0 & 50.0 & 40.0 & 0.0 & 30.0 & 10.0 & 0.0 & \best{100}\\
 & L3 & 80.0 & 80.0 & 70.0 & \second{90.0} & \second{90.0} & 0.0 & 60.0 & 50.0 & 0.0 & \best{100}\\
\midrule
StirSpoon & L0 & \best{100} & 70.0 & 50.0 & 40.0 & 10.0 & 0.0 & 70.0 & 40.0 & 0.0 & \best{100}\\
 & L1 & 40.0 & 50.0 & 40.0 & 40.0 & 10.0 & 0.0 & 0.0 & 0.0 & 0.0 & \best{100}\\
 & L2 & 60.0 & 0.0 & 20.0 & 50.0 & 40.0 & 0.0 & \second{80.0} & 40.0 & 0.0 & 70.0\\
 & L3 & 50.0 & 70.0 & 20.0 & 60.0 & 20.0 & 0.0 & 30.0 & 60.0 & 0.0 & \best{90.0}\\
\midrule
\textbf{Overall} &  & 80.5 & 77.5 & 67.0 & 68.5 & 63.0 & 2.5 & 58.5 & 58.0 & 20.5 & \best{88.5}\\
\bottomrule
\end{tabular}
}
\end{appendixresulttable}
\endgroup
\begingroup
\input{tables/experiment_style}
\begin{appendixresulttable}
\caption{\textbf{Complete C45 Unseen results.} SR (\%) for every task and scene level. D/S/V denote DINOv2, SigLIP2, and VideoMAE. RoboDemo is repeated in both panels for comparison. Blue/light blue mark the best/second-best among all 17 methods for the same row, across both panels.}
\label{tab:baseline-zs-45}
\nopagebreak
\centering
\fontsize{7pt}{8pt}\selectfont
\resizebox{\linewidth}{!}{%
\tabcolsep=1pt
\renewcommand\arraystretch{1.0}
\begin{tabular}{@{}lc|*{8}{c}@{}}
\toprule
\multicolumn{10}{c}{(a) Language-conditioned, video-skill, and in-context policies}\\
\midrule
Task & Level & \shortstack{GR00T-\\N1.6} & \shortstack{Open\\VLA} & $\pi_{0.5}$ & \shortstack{RDT-\\1B} & \shortstack{Uni\\Skill} & XSkill & \shortstack{Zero-\\WAM} & \shortstack{Robo\\Demo}\\
\midrule
FoldBox & L0 & 0.0 & 20.0 & 0.0 & 0.0 & 50.0 & 10.0 & \best{100} & \second{90.0}\\
 & L1 & 0.0 & 0.0 & \best{100} & 0.0 & 0.0 & 30.0 & 30.0 & 0.0\\
 & L2 & 80.0 & \best{100} & 0.0 & \second{90.0} & 0.0 & 0.0 & \second{90.0} & 0.0\\
 & L3 & 0.0 & 0.0 & \second{90.0} & 0.0 & 0.0 & 0.0 & \best{100} & \best{100}\\
\midrule
PickFood & L0 & \best{0.0} & \best{0.0} & \best{0.0} & \best{0.0} & \best{0.0} & \best{0.0} & \best{0.0} & \best{0.0}\\
 & L1 & \second{0.0} & \second{0.0} & \second{0.0} & \second{0.0} & \second{0.0} & \second{0.0} & \best{30.0} & \second{0.0}\\
 & L2 & \best{0.0} & \best{0.0} & \best{0.0} & \best{0.0} & \best{0.0} & \best{0.0} & \best{0.0} & \best{0.0}\\
 & L3 & \best{0.0} & \best{0.0} & \best{0.0} & \best{0.0} & \best{0.0} & \best{0.0} & \best{0.0} & \best{0.0}\\
\midrule
PickRemoteControl & L0 & \best{0.0} & \best{0.0} & \best{0.0} & \best{0.0} & \best{0.0} & \best{0.0} & \best{0.0} & \best{0.0}\\
 & L1 & \best{0.0} & \best{0.0} & \best{0.0} & \best{0.0} & \best{0.0} & \best{0.0} & \best{0.0} & \best{0.0}\\
 & L2 & \best{0.0} & \best{0.0} & \best{0.0} & \best{0.0} & \best{0.0} & \best{0.0} & \best{0.0} & \best{0.0}\\
 & L3 & 0.0 & 0.0 & 0.0 & 0.0 & 0.0 & \second{10.0} & 0.0 & 0.0\\
\midrule
PourKettle & L0 & 0.0 & 0.0 & 10.0 & 0.0 & 0.0 & \second{40.0} & \best{50.0} & 10.0\\
 & L1 & \second{0.0} & \second{0.0} & \second{0.0} & \second{0.0} & \second{0.0} & \second{0.0} & \best{10.0} & \best{10.0}\\
 & L2 & \best{0.0} & \best{0.0} & \best{0.0} & \best{0.0} & \best{0.0} & \best{0.0} & \best{0.0} & \best{0.0}\\
 & L3 & \second{0.0} & \second{0.0} & \second{0.0} & \second{0.0} & \second{0.0} & \second{0.0} & \best{30.0} & \second{0.0}\\
\midrule
ScanMilkBox & L0 & 10.0 & 0.0 & 0.0 & 0.0 & \best{40.0} & 0.0 & 0.0 & 10.0\\
 & L1 & \best{0.0} & \best{0.0} & \best{0.0} & \best{0.0} & \best{0.0} & \best{0.0} & \best{0.0} & \best{0.0}\\
 & L2 & \best{20.0} & 0.0 & 0.0 & 0.0 & 0.0 & \second{10.0} & 0.0 & \second{10.0}\\
 & L3 & \best{0.0} & \best{0.0} & \best{0.0} & \best{0.0} & \best{0.0} & \best{0.0} & \best{0.0} & \best{0.0}\\
\midrule
\textbf{Overall} &  & 5.5 & 6.0 & 10.0 & 4.5 & 4.5 & 5.0 & \best{22.0} & 11.5\\
\bottomrule
\end{tabular}
}
\par\vspace{7pt}
\fontsize{7pt}{8pt}\selectfont
\resizebox{\linewidth}{!}{%
\tabcolsep=1pt
\renewcommand\arraystretch{1.0}
\begin{tabular}{@{}lc|*{10}{c}@{}}
\toprule
\multicolumn{12}{c}{(b) Video-conditioned policies}\\
\midrule
Task & Level & \shortstack{ACT\\D} & \shortstack{ACT\\S} & \shortstack{ACT\\V} & \shortstack{DP\\D} & \shortstack{DP\\S} & \shortstack{DP\\V} & \shortstack{VQ-BeT\\D} & \shortstack{VQ-BeT\\S} & \shortstack{VQ-BeT\\V} & \shortstack{Robo\\Demo}\\
\midrule
FoldBox & L0 & 0.0 & 40.0 & 10.0 & 0.0 & 0.0 & 0.0 & \best{100} & \second{90.0} & 0.0 & \second{90.0}\\
 & L1 & 0.0 & 0.0 & 0.0 & 50.0 & 30.0 & 0.0 & \second{90.0} & \second{90.0} & \best{100} & 0.0\\
 & L2 & 10.0 & 0.0 & 0.0 & 40.0 & 30.0 & 40.0 & 0.0 & 0.0 & 40.0 & 0.0\\
 & L3 & 0.0 & 0.0 & 30.0 & 0.0 & 0.0 & 0.0 & \best{100} & 30.0 & 0.0 & \best{100}\\
\midrule
PickFood & L0 & \best{0.0} & \best{0.0} & \best{0.0} & \best{0.0} & \best{0.0} & \best{0.0} & \best{0.0} & \best{0.0} & \best{0.0} & \best{0.0}\\
 & L1 & \second{0.0} & \second{0.0} & \second{0.0} & \second{0.0} & \second{0.0} & \second{0.0} & \second{0.0} & \second{0.0} & \second{0.0} & \second{0.0}\\
 & L2 & \best{0.0} & \best{0.0} & \best{0.0} & \best{0.0} & \best{0.0} & \best{0.0} & \best{0.0} & \best{0.0} & \best{0.0} & \best{0.0}\\
 & L3 & \best{0.0} & \best{0.0} & \best{0.0} & \best{0.0} & \best{0.0} & \best{0.0} & \best{0.0} & \best{0.0} & \best{0.0} & \best{0.0}\\
\midrule
PickRemoteControl & L0 & \best{0.0} & \best{0.0} & \best{0.0} & \best{0.0} & \best{0.0} & \best{0.0} & \best{0.0} & \best{0.0} & \best{0.0} & \best{0.0}\\
 & L1 & \best{0.0} & \best{0.0} & \best{0.0} & \best{0.0} & \best{0.0} & \best{0.0} & \best{0.0} & \best{0.0} & \best{0.0} & \best{0.0}\\
 & L2 & \best{0.0} & \best{0.0} & \best{0.0} & \best{0.0} & \best{0.0} & \best{0.0} & \best{0.0} & \best{0.0} & \best{0.0} & \best{0.0}\\
 & L3 & 0.0 & 0.0 & 0.0 & 0.0 & 0.0 & 0.0 & \best{60.0} & 0.0 & \second{10.0} & 0.0\\
\midrule
PourKettle & L0 & 0.0 & 0.0 & 0.0 & 0.0 & 0.0 & 10.0 & 0.0 & 0.0 & 0.0 & 10.0\\
 & L1 & \second{0.0} & \second{0.0} & \second{0.0} & \second{0.0} & \second{0.0} & \second{0.0} & \second{0.0} & \second{0.0} & \second{0.0} & \best{10.0}\\
 & L2 & \best{0.0} & \best{0.0} & \best{0.0} & \best{0.0} & \best{0.0} & \best{0.0} & \best{0.0} & \best{0.0} & \best{0.0} & \best{0.0}\\
 & L3 & \second{0.0} & \second{0.0} & \second{0.0} & \second{0.0} & \second{0.0} & \second{0.0} & \second{0.0} & \second{0.0} & \second{0.0} & \second{0.0}\\
\midrule
ScanMilkBox & L0 & 0.0 & 0.0 & 10.0 & \second{20.0} & 0.0 & 0.0 & 0.0 & 0.0 & 0.0 & 10.0\\
 & L1 & \best{0.0} & \best{0.0} & \best{0.0} & \best{0.0} & \best{0.0} & \best{0.0} & \best{0.0} & \best{0.0} & \best{0.0} & \best{0.0}\\
 & L2 & \second{10.0} & 0.0 & 0.0 & 0.0 & 0.0 & \second{10.0} & 0.0 & 0.0 & 0.0 & \second{10.0}\\
 & L3 & \best{0.0} & \best{0.0} & \best{0.0} & \best{0.0} & \best{0.0} & \best{0.0} & \best{0.0} & \best{0.0} & \best{0.0} & \best{0.0}\\
\midrule
\textbf{Overall} &  & 1.0 & 2.0 & 2.5 & 5.5 & 3.0 & 3.0 & \second{17.5} & 10.5 & 7.5 & 11.5\\
\bottomrule
\end{tabular}
}
\end{appendixresulttable}
\endgroup

%% file: tables/experiment_timing_protocol.tex
\begingroup
\input{tables/experiment_style}
\begin{table}[H]
\centering
\scriptsize
\tabcolsep=2pt
\renewcommand\arraystretch{1.0}
\begin{tabularx}{\linewidth}{@{}>{\raggedright\arraybackslash}X|rrrr@{}}
\toprule
Stage & \shortstack{Mean\\(ms)} & \shortstack{Median\\(ms)} & \shortstack{p95\\(ms)} & \shortstack{SD\\(ms)}\\
\midrule
VLM backbone & 64.7 & 63.8 & 67.6 & 2.4\\
Original action expert (five rounds) & 144.1 & 143.6 & 147.9 & 2.8\\
TTT write (once) & 55.1 & 54.5 & 58.4 & 1.7\\
TTT reads (five rounds) & 295.1 & 294.4 & 305.3 & 6.4\\
Memory adapters & 1.8 & 1.8 & 1.9 & 0.1\\
\midrule
Full policy query (uninstrumented) & 563.6 & 559.9 & 588.1 & 12.9\\
Full policy query (instrumented) & 578.7 & 577.1 & 597.5 & 11.0\\
\bottomrule
\end{tabularx}
\caption{\textbf{Measured policy-query latency distributions.} Each row contains 90 measurements from three recorded inputs, following ten warm-up calls per input. TTT writes occur once per query. TTT reads and the original action expert aggregate five denoising rounds. TTT times cover all 36 layers. Policy-component rows use CUDA-event intervals. Whole-query rows use synchronized wall time.}
\label{tab:timing-protocol}
\end{table}
\endgroup

%% file: segments/appendix_refreward.tex
\section{Refreward: Query-Conditioned Demonstration Selection}
\label{app:refreward}

This appendix describes the construction of the retrieval supervision corpus and the training of Refreward for query-conditioned demonstration selection.

\subsection{Problem, inputs, and decision ownership}
\label{app:rr_problem}

Let $\mathcal B=\{e_i\}_{i=1}^{N_{\mathcal B}}$ be a split-legal reference bank, with $e_i=(\ell_i,o_i^0,V_i)$, and let $q=(\ell_q,o_q^0)$ be the deployment query. A reference may successfully complete its own task yet be unsuitable for the query because of object identity, attributes, initial layout, constraints, ordering, or missing procedure coverage. Consequently $r_\psi(q,e_i)$ is a directed, query-dependent score, not an intrinsic episode-quality label or a calibrated control-success probability.

The information flow is
\begin{equation}
\begin{aligned}
&\underbrace{\text{three initial representations}}_{\text{query and candidate}}
\\
&\quad+\underbrace{\text{candidate-process representation}}_{\text{candidate video only}}\\
&\quad\longrightarrow \operatorname{Refreward}_{\psi}\\
&\quad\longrightarrow \text{highest-scoring candidate reference}.
\end{aligned}
\end{equation}
The pretrained encoders are feature extractors, not competing final decision makers. Initial image--text similarity helps construct annotation candidate sets and supplies a learned score prior. The reward model evaluates the task-compatible deployment candidates using their cached evidence. No task ID, reference identity embedding, query future image, executed query action, or query outcome is a model input. Episode IDs serve only bookkeeping and exclusion. The scorer operates on query and candidate features. Task metadata defines candidate eligibility but is not a neural input.

\subsection{Frozen backbones and exact token extraction}
\label{app:rr_tokens}

\paragraph{Weights and preprocessing.}
We use frozen \nolinkurl{Qwen/Qwen3-VL-Embedding-2B} and \nolinkurl{facebook/dinov3-vith16plus-pretrain-lvd1689m} in deterministic evaluation mode. They produce 2048-dimensional Qwen representations and 1280-dimensional DINO patch features~\citep{li2026qwen3vlembedding,simeoni2025dinov3}. The two Qwen input calls below share one loaded checkpoint. DINOv3 supplies dense visual features.

The retriever uses one fixed workspace camera for query and reference. Qwen images are aspect-preserving letterboxed to $448\times448$ and DINO images to $256\times256$, using each encoder's pretrained normalization. Padding masks and inverse pixel transforms preserve the correspondence between features and original image regions. A 448-square Qwen image yields 196 merged visual positions, excluding text and special tokens.

\end{multicols}
\begin{table}[H]
\centering\small
\caption{Raw per-episode feature contract. Padded rows are masked, never interpreted as evidence. $L_t=32$ is the retained instruction-token budget, $L_v=256$ the DINO patch count, and $L_u=32$ the candidate-frame budget.}
\label{tab:rr_features}
\begin{tabularx}{\linewidth}{@{}l l X@{}}
\toprule
Field & Shape & Construction and purpose \\
\midrule
$g_i$ & $2048$ & Normalized text-only Qwen global embedding for instruction recall. \\
$z_i$ & $2048$ & Normalized joint instruction--initial-image Qwen global embedding for scene-aware recall. \\
$T_i$ & $32\times2048$ & Contextual instruction hidden states from the text-only forward, with ordered pooling if needed. \\
$F_i$ & $256\times1280$ & DINO final normalized patch tokens, excluding class and register tokens. \\
$C_i$ & $256\times24$ & Three independent 8-bin Lab channel histograms on each patch's valid pixels. \\
$P_i$ & $256\times4$ & Patch center $(x,y)$ and extent $(w,h)$ in normalized unpadded image coordinates. \\
$U_i$ & $32\times2048$ & Joint Qwen embedding of instruction and each timestamped candidate frame. \\
\bottomrule
\end{tabularx}
\end{table}
\begin{multicols}{2}
\raggedcolumns

\paragraph{Text-only call.}
We use the official embedding chat template with the fixed system instruction ``Represent the user's input.'' and user text $\ell_i$. Let $H_i\in\mathbb R^{S_i\times2048}$ be its final hidden states and $a_i$ the attention mask. As in the official embedding implementation, pool the last \emph{non-padding} position:
\begin{equation}
\begin{aligned}
g_i&=\operatorname{L2Norm}(H_i[j_i]),\\
j_i&=\max\{j:a_i[j]=1\}.
\end{aligned}
\end{equation}
Tokenizer offsets against the rendered chat-template span of $\ell_i$ identify instruction-only rows, excluding template and special tokens. Extraction checks this span before retaining its rows. For $n_i\leq32$ content tokens, we retain their ordered hidden states and mask padding. For $n_i>32$, we partition their indices into 32 nonempty consecutive bins with boundaries $\lfloor bn_i/32\rfloor$ and average within each bin. A fixed sinusoidal encoding of each bin's mean normalized instruction position is added after projection. The global call sees the complete instruction. The default accepts up to 512 instruction tokens and rejects, rather than silently truncates, longer inputs.

The retained rows are pretrained causal contextual hidden states. Contiguous pooling carries coarse instruction order into the scorer, where instruction-position encodings and learned cross-attention connect these rows to visual and process features.

\paragraph{Joint call.}
The joint call uses the same system instruction and template, with $o_i^0$ followed by $\ell_i$. Last-valid-position pooling and L2 normalization produce $z_i$, which represents the initial scene together with its instruction. Dense visual, color, and position tokens complement this global representation.

\paragraph{Dense visual, color, and position tokens.}
We encode the 256-square normalized image with DINOv3 and retain its final patch-token output (the official implementation's \texttt{x\_norm\_patchtokens}). The $16\times16$ grid gives $L_v=256$ rows. Each row $f_a$ has its original patch index. Class and register rows are discarded, not averaged into local tokens. Each $f_a$ is L2-normalized before its learned projection.

For the same patch's valid RGB pixels \emph{before encoder normalization}, we convert sRGB under the standard D65 white point to CIE Lab and scale $L^\ast$ from $[0,100]$ and $a^\ast,b^\ast$ from $[-128,128]$ to $[0,1]$, clipping endpoints. For channel $h$ and bin $b=0,\ldots,7$, the histogram is
\begin{equation}
c_{a,h,b}=\frac{\#\{x\in\Omega_a:\lfloor\min(8v_h(x),7)\rfloor=b\}}
{\max(1,|\Omega_a|)}.
\end{equation}
Concatenating the three eight-bin histograms gives 24 color features per patch. The position $p_a=(x_a,y_a,w_a,h_a)$ is obtained by applying the saved inverse letterbox transform to the patch rectangle and normalizing by original width and height. Entirely padded patches are masked. Partial patches use only valid pixels and the clipped valid rectangle.

Reference-bank entries share the workspace camera and calibration group. Image coordinates describe relative layout in this aligned image plane, while the Lab histograms retain patch-level color distributions.

\paragraph{Candidate-process tokens.}
For an episode with $J_i$ valid frames, we choose up to 32 uniformly spaced source indices including the first and last, remove rounded duplicates, and mask missing slots. For each selected frame $v_{i,u}$, the frozen \emph{joint} call on $(\ell_i,v_{i,u})$ produces the cached normalized 2048-dimensional embedding $U_i[u]$. Frame indices, timestamps, and normalized episode times $\tau_{i,u}$ accompany these features. A fixed sinusoidal time encoding is added after projection. The scorer learns temporal relations from the ordered sequence of per-frame semantic embeddings.

The teacher and student use the same frame manifest and camera. Teacher labels use the query initial observation and these same candidate frames. DINO features are extracted only from the query and candidate initial images. Candidate process frames are represented by per-frame joint Qwen embeddings, with the initial-frame embedding reused as the first process token. No numerical robot state or action is supplied to Refreward. The full chosen video remains available to RoboDemo's subsequent H-GEC and Qwen/V-JEPA2 reference path. Those policy encoders and Refreward's retrieval encoders have different roles and caches.

\subsection{Candidate bank and final selection}
\label{app:rr_recall}

The IG robot bank contains 200 demonstrations per task. Training-reference assignment scores the other 199 same-task episodes and fixes the highest-scoring reference. C15-wo applies this assignment to its 3,000 policy-training episodes. For evaluation, the bank contains the query task's 200 robot training demonstrations. Seen-task candidates belong to C15, and policy-held-out tasks receive a task-matched reference from the IG bank. Candidate features are precomputed, and the frozen scorer selects the reference from the query instruction and initial observation without accessing query future frames. The chosen reference remains fixed throughout the rollout.

At deployment, one Qwen text forward computes $(g_q,T_q)$, one Qwen joint forward computes $z_q$, and one DINO forward plus pixel statistics computes $(F_q,C_q,P_q)$. The query is encoded once and reused across candidate scoring. Training-reference assignment excludes the query episode itself. The final selection is
\begin{equation}
i^\star=\arg\max_{i\in\mathcal C_q}r_\psi(q,e_i).
\end{equation}
The compatibility and sufficiency heads provide auxiliary supervision. Their outputs are not averaged into the ranking score or used as an additional acceptance threshold. After selection, the policy encodes the chosen reference through its own semantic--motion path to construct Retrieval-TTT memory. This policy preprocessing is distinct from Refreward's candidate-process embeddings.

\subsection{Trainable Refreward architecture}
\label{app:rr_architecture}

\paragraph{Projection and spatial encoding.}
The hidden width is $d=256$, with six bias-free learned maps
$A_g,A_z,A_T:\mathbb R^{2048}\rightarrow\mathbb R^d$,
$A_F:\mathbb R^{1280}\rightarrow\mathbb R^d$,
$A_C:\mathbb R^{24}\rightarrow\mathbb R^d$, and
$A_P:\mathbb R^4\rightarrow\mathbb R^d$.
$A_z$ is shared between initial joint and process embeddings, and all projections are shared between query and candidate. A spatial input row is
\begin{equation}
\begin{aligned}
x_a&=N(A_F\hat f_a)+0.25N(A_Cc_a)\\
   &\quad+0.25N(A_Pp_a),\\
N(x)&=\sqrt d\,x/\|x\|_2.
\end{aligned}
\end{equation}
Here $\hat f_a$ is the L2-normalized DINO patch feature. Zero vectors remain zero under normalization. Projected content features are normalized to a common scale. Color and position contributions, as well as fixed positional encodings, use a gain of 0.25. Learned modality and role embeddings are scaled by $\sqrt d$. Two pre-LayerNorm Transformer blocks process these rows with eight heads and a $4d$ GELU feed-forward layer. Each attention head receives a learned relative-position bias
\begin{equation}
\begin{aligned}
b_{ab}=\operatorname{MLP}_{4\to32\to8}
\bigl[&x_a^{\mathrm{pos}}-x_b^{\mathrm{pos}},\\
      &y_a^{\mathrm{pos}}-y_b^{\mathrm{pos}},\\
      &|x_a^{\mathrm{pos}}-x_b^{\mathrm{pos}}|,\\
      &|y_a^{\mathrm{pos}}-y_b^{\mathrm{pos}}|\bigr].
\end{aligned}
\end{equation}
The positional symbols here are patch centers, not hidden channels. The bias MLP is shared across the two blocks and its grid output is computed once. Softmax respects padding masks. This encoder preserves a grid of 256 contextual visual rows. Appearance, color, and geometry become jointly available to learned comparisons.

\paragraph{Cross-evidence scoring.}
Concatenating projected text-global, joint-global, instruction, and spatial rows forms $X_q\in\mathbb R^{290\times256}$. Appending 32 ordered process rows gives $X_i\in\mathbb R^{322\times256}$. Five learned modality embeddings and two role embeddings identify the feature type and query/candidate role. Fixed instruction-position and process-time encodings carry order. All padded rows are masked.

The scorer initializes 32 learned reward-latent rows $R^0$. Each of four blocks performs pre-normalized latent-to-evidence cross-attention, latent self-attention, and a $4d$ GELU feed-forward update:
\begin{align}
\widetilde R^b &=R^{b-1}\nonumber\\
 &\quad+\operatorname{CrossAttn}_b\bigl(
 \operatorname{LN}(R^{b-1}),K_{qi},V_{qi}\bigr),\\
\bar R^b &=\widetilde R^b+
 \operatorname{SelfAttn}_b(\operatorname{LN}(\widetilde R^b)),\\
R^b &=\bar R^b+\operatorname{FFN}_b(\operatorname{LN}(\bar R^b)).
\end{align}
Crucially, the evidence is immutable within these blocks:
\begin{equation}
\begin{aligned}
K_{qi}&=[\operatorname{LN}(X_q)W_K\Vert\operatorname{LN}(X_i)W_K],\\
V_{qi}&=[\operatorname{LN}(X_q)W_V\Vert\operatorname{LN}(X_i)W_V].
\end{aligned}
\end{equation}
Here $\Vert$ concatenates token rows. The two $W_K,W_V$ maps are \emph{shared across all four blocks}. Query/output projections and latent self-attention/FFNs are block-specific. This is the architectural condition permitting one cached candidate K/V pair instead of four. Candidate encoding and K/V are query-independent. Query--candidate interactions occur in the reward latents, so final scores are still query-dependent. During training these trainable candidate transforms are recomputed with gradients. Detached deployed K/V caches are valid only after the selector checkpoint is frozen.

Mean-pooling the final latent rows and applying a bias-free $256\to256\to9$ GELU head gives nine outputs. Its outputs are a scalar reward residual $\Delta_\psi(q,e_i)$, two gate logits, and six factor logits: object/goal fit, attribute fit, layout transferability, procedural order/coverage, demonstration quality, and visibility/evidence adequacy. The final ranking score adds a learned positive-scale initial image--text similarity prior:
\begin{equation}
\begin{aligned}
r_\psi(q,e_i)&=\Delta_\psi(q,e_i)\\
&\quad+\operatorname{softplus}(b)[\cos(z_q,z_i)-0.9].
\end{aligned}
\label{eq:rr-final-score}
\end{equation}
The auxiliary gate and factor outputs are not averaged into this score. The positive prior scale is initialized to 20. Evidence localization uses the last cross-attention layer's attention to candidate-process slots, averaged over heads/latents and normalized within those slots. The resulting distribution supplies the auxiliary frame-attribution prediction.

All learned linear maps are bias-free. LayerNorm has no affine parameters. There are 32 learned latent rows, seven role/modality rows, and four learned ordinal thresholds. This configuration has \textbf{7,359,877 trainable parameters}, excluding the two frozen backbones. Dropout is zero.

\subsection{Teacher annotation and dataset construction}
\label{app:rr_teacher}

\paragraph{Teacher-supervised reference suitability.}
Inspired by reward modeling in reinforcement learning from human feedback, we jointly train Refreward with GPT-provided pointwise suitability grades and query-conditioned pairwise preferences. A pointwise annotation evaluates a query--candidate pair and provides an overall suitability grade from 1 to 5, together with six factor-level ratings. A pairwise annotation compares two candidates under the same query and indicates whether candidate A is better, candidate B is better, or neither is clearly preferred. These complementary signals supervise both graded suitability and relative preference.

\paragraph{Corpus construction.}
We construct the corpus from the 10,000 robot demonstrations in the Imitator Game training split. Across its 50 tasks, we select 350 fixed queries by stratifying over the four scene levels, with seven queries per task. For each query, we construct a candidate set using initial image--text similarity together with level-stratified sampling. The GPT teacher receives the query instruction and initial observation, together with each candidate's instruction and 32 timestamped process frames, including its initial frame. Query future observations and human demonstration videos are not provided to the teacher. The resulting corpus contains approximately 13,000 pointwise conditions and 2,100 pairwise conditions.

Robot-side instructions are established from task specifications and initial robot observations, without using human demonstration media. The teacher is GPT-6. Its service-returned model identifier is retained in annotation provenance. The teacher receives images and text, not the student's encoded embeddings, and is not called during deployment.

\paragraph{Structured labels.}
The JSON schema records schema/model/prompt/evidence versions, compatibility and sufficient-evidence labels with an explicit \texttt{unknown} value, an overall ordinal reference-suitability rating in $\{1,\ldots,5\}$, and six factor ratings on the same scale. It also records supported time intervals and frame IDs, visible object/role correspondences with optional normalized boxes, missing/contradictory evidence codes, and short evidence-grounded rationales. Grades 1 and 2 denote incompatible or misleading references and weak or incomplete references, respectively. Grade 3 denotes partial transferability with material mismatch, while grade 4 denotes suitability with minor adaptation. Grade 5 denotes a strong, complete visible reference for the query. Unknown evidence is not a negative label. The ordinal score measures judged reference usefulness, not probability of successful execution.

A preference request shows one query and two candidate evidence packets. Its JSON records \texttt{A}, \texttt{B}, \texttt{tie}, or \texttt{insufficient\_evidence}, with decisive factors and supporting frame IDs. A/B presentation is randomized and retrieval rank is concealed. Each pointwise and pairwise condition receives three judgments with shuffled candidate presentation where applicable. Raw structured responses and disagreements are retained alongside the consensus.

\paragraph{Aggregation and retained counts.}
Each annotation condition receives three teacher judgments. Known ordinal ratings are aggregated by median and categorical judgments by majority. Gate disagreement or a rating spread above one point triggers review. Unknown fields are masked during training. The retained corpus contains 12,994 pointwise conditions and 2,099 pairwise conditions, yielding 15,093 training conditions and 45,279 ordinary teacher judgments.

\end{multicols}
\input{tables/refreward_corpus}
\begin{multicols}{2}
\raggedcolumns

The overall suitability grade is a separate teacher judgment rather than the arithmetic mean of the six factor ratings. Compatibility and evidence sufficiency are additionally annotated as categorical judgments. Supporting frames, temporal intervals, object correspondences, and short rationales are retained with the labels. Table~\ref{tab:rr-factors} defines the six factors.

\end{multicols}
\input{tables/refreward_factors}
\begin{multicols}{2}
\raggedcolumns

\subsection{Supervised objectives and training}
\label{app:rr_training}

Let $y\in\{1,\ldots,5\}$ be a known ordinal target. Parameterize four increasing thresholds $\theta_k$ using one free base and three positive softplus increments, and use
\begin{equation}
\mathcal L_{\mathrm{ord}}=\frac14\sum_{k=1}^{4}
\operatorname{BCE}\bigl(\sigma(r-\theta_k),\mathbf1[y>k]\bigr).
\end{equation}
For a judged pair $(q,i,j)$ use the Bradley--Terry loss
\begin{equation}
\begin{aligned}
\mathcal L_{\mathrm{pair}}
&=\operatorname{BCE}\bigl(\sigma(r_\psi(q,e_i)-r_\psi(q,e_j)),y_{ij}\bigr),\\
&\qquad y_{ij}\in\{1,0,\tfrac12\}.
\end{aligned}
\end{equation}
The half target denotes a teacher tie. Insufficient-evidence pairs are masked, not ties. Known compatibility and sufficiency labels use BCE. Known factor ratings use BCE with target $(y_f-1)/4$. Annotated supporting frames use cross-entropy on the normalized process-frame evidence distribution. Unknown factors, missing evidence annotations, and all-invalid supports contribute no corresponding loss. Each loss is averaged over its own known-label denominator:
\begin{equation}
\begin{aligned}
\mathcal L_{\mathrm{RR}}
&=\mathcal L_{\mathrm{pair}}+\mathcal L_{\mathrm{ord}}\\
&\quad+0.5\mathcal L_{\mathrm{gates}}+0.25\mathcal L_{\mathrm{factors}}\\
&\quad+0.1\mathcal L_{\mathrm{evidence}}+0.5\mathcal L_{\mathrm{contrast}}.
\end{aligned}
\end{equation}
Pairwise and pointwise examples share the scalar reward. Known-incompatible examples remain in training. No ground-truth gate masks their ordinal/ranking gradients.

We additionally derive pairwise ordering constraints from candidates with different known pointwise grades under the same query. This ordinal-contrast term complements, rather than replaces, the directly annotated pairwise preferences:
\begin{align}
\mathcal L_{\mathrm{contrast}}
&=\frac{\sum_{(i,j)\in\mathcal P_q}w_{ij}
\log[1+\exp(-d_{ij}(r_i-r_j))]}
{\sum_{(i,j)\in\mathcal P_q}w_{ij}},\\
d_{ij}&=\operatorname{sign}(y_i-y_j),\\
w_{ij}&=|y_i-y_j|
\exp\!\left(\frac{s_i+s_j-2s_{\max}}{0.1}\right).
\end{align}
Here $\mathcal P_q$ contains $i<j$ with distinct known pointwise grades under query $q$, $s_i=\cos(g_q,g_i)$ is frozen instruction similarity, and $s_{\max}$ is its maximum over the query group. An empty valid pair set contributes zero. The denominator is clamped below by $10^{-12}$ for numerical stability. All losses are computed over their respective valid-label masks. The scalar score is trained jointly by ordinal grades and relative preferences. The six factor predictions and two categorical predictions provide auxiliary supervision.

We train Refreward with query-grouped batches using all 15,093 retained annotation conditions. Each update scores the candidates associated with one query and combines the available pointwise, pairwise, and auxiliary targets. The frozen encoder features are cached, while trainable candidate transformations are recomputed with gradients. We use AdamW with weight decay $10^{-2}$, gradient clipping at 1, and seed 0. The initial 4,000 updates use a peak learning rate of $10^{-4}$, 5\% warm-up, and cosine decay. The final 2,000 updates use a learning rate of $10^{-5}$. The deployed scorer is the checkpoint at 6,000 updates, selected using a fixed development set.

%% file: tables/refreward_corpus.tex
\begin{table}[H]
\centering
\footnotesize
\captionsetup{width=\linewidth}
\caption{Retained Imitator Game retrieval-supervision corpus. Repeated teacher judgments are votes for one condition, not additional independent query--candidate samples.}
\label{tab:rr-corpus}
\begin{tabularx}{\linewidth}{@{}X r@{}}
\toprule
Item & Count \\
\midrule
Robot demonstrations in the bank & 10,000 \\
Tasks & 50 \\
Fixed queries & 350 \\
Queries per task & 7 \\
Retained pointwise conditions & 12,994 \\
Retained pairwise conditions & 2,099 \\
Total retained conditions & 15,093 \\
Ordinary judgments per condition & 3 \\
Ordinary teacher judgments & 45,279 \\
\bottomrule
\end{tabularx}
\end{table}

%% file: tables/refreward_factors.tex
\begin{table}[H]
\centering
\small
\caption{Six teacher-annotated factors of reference suitability. Each known rating lies on the 1--5 scale.}
\label{tab:rr-factors}
\begin{tabularx}{\linewidth}{@{}>{\raggedright\arraybackslash}p{0.27\linewidth}X@{}}
\toprule
Factor & Definition \\
\midrule
Object/goal fit & Whether the demonstrated objects and intended outcome match the query. \\
Attribute fit & Whether task-relevant object attributes, such as color, category, and configuration, are compatible with the query. \\
Layout transferability & Whether the demonstrated spatial arrangement and object relations can support the query with appropriate spatial adaptation. \\
Procedural order/coverage & Whether the visible procedure covers the required operations in a compatible order. \\
Demonstration quality & Whether the candidate visibly provides coherent and effective execution of its procedure. \\
Visibility & Whether relevant objects, interactions, and outcomes are sufficiently visible to support the judgment. \\
\bottomrule
\end{tabularx}
\end{table}